\documentclass[letterpaper]{article} % DO NOT CHANGE THIS
\usepackage{aaai2026}  % DO NOT CHANGE THIS
\usepackage{times}  % DO NOT CHANGE THIS
\usepackage{helvet}  % DO NOT CHANGE THIS
\usepackage{courier}  % DO NOT CHANGE THIS
\usepackage[hyphens]{url}  % DO NOT CHANGE THIS
\usepackage{graphicx} % DO NOT CHANGE THIS
\usepackage{natbib}  % DO NOT CHANGE THIS AND DO NOT ADD ANY OPTIONS TO IT
\usepackage{caption} % DO NOT CHANGE THIS AND DO NOT ADD ANY OPTIONS TO IT
\usepackage{algorithm}
\usepackage{algorithmic}

\usepackage{enumitem}
\usepackage{amsmath}
\usepackage{amssymb}
\usepackage{multirow}
\usepackage{tabularx}
\usepackage{booktabs}
\usepackage{makecell}
\usepackage{arydshln}
\usepackage{pdfpages}

\usepackage{pifont}
\usepackage{hhline}

\usepackage{xcolor}

\newcounter{checksubsection}
\newcounter{checkitem}[checksubsection]

\usepackage{newfloat}
\usepackage{listings}
\DeclareCaptionStyle{ruled}{labelfont=normalfont,labelsep=colon,strut=off} % DO NOT CHANGE THIS
\floatstyle{ruled}
\newfloat{listing}{tb}{lst}{}
\floatname{listing}{Listing}
\nocopyright %-- Your paper will not be published if you use this command
\title{STF: Shallow-Level Temporal Feedback to Enhance Spiking Transformers}

\author{
\normalfont Zeqi Zheng\textsuperscript{1,2*}, 
\normalfont Zizheng Zhu\textsuperscript{2,6*},
\normalfont Yingchao Yu\textsuperscript{4,2},
\normalfont Yanchen Huang\textsuperscript{2,3},
\normalfont Changze Lv\textsuperscript{5}, \\
\normalfont Junfeng Tang\textsuperscript{1,2},
\normalfont Zhaofei Yu\textsuperscript{7},
\normalfont Yaochu Jin\textsuperscript{2}$^\dagger$ \\
\textsuperscript{1}Zhejiang University \quad 
\textsuperscript{2}Westlake University \quad
\textsuperscript{3}Nanjing University \quad 
\textsuperscript{4}Donghua University\\
\textsuperscript{5}Fudan University \quad
\textsuperscript{6}University of Electronic Science and Technology of China \quad
\textsuperscript{7}Peking University\\
\tt\small \{zhengzeqi, zhuzizheng, jinyaochu\}@westlake.edu.cn
}

\usepackage{bibentry}
\begin{document}

\maketitle

{\renewcommand{\thefootnote}{}
\footnotetext{$^*$ Equal contribution. $^\dagger$ Corresponding author.}}

\begin{abstract}

Transformer-based Spiking Neural Networks (SNNs) suffer from a great performance gap compared to floating-point \mbox{Artificial} Neural Networks (ANNs) due to the binary nature of spike trains. 
Recent efforts have introduced deep-level feedback loops to transmit high-level semantic information to narrow this gap. 
However, these designs often span \mbox{multiple} deep layers, resulting in costly feature transformations, higher parameter overhead, increased energy consumption, and longer inference latency. 
To address this issue, we propose Shallow-level Temporal Feedback (STF), a lightweight plug-and-play module for the encoding layer, which consists of Temporal-Spatial Position Embedding (TSPE) and Temporal Feedback (TF).
Extensive experiments show that STF consistently improves performance across various Transformer-based SNN backbones on static datasets, including CIFAR-10, CIFAR-100, and ImageNet-1K, under different spike timestep settings. 
Further analysis reveals that STF enhances the diversity of spike patterns, which is key to performance gain. 
Moreover, evaluations on adversarial robustness and temporal sensitivity confirm that STF outperforms direct coding and its variants, highlighting its potential as a new spike encoding scheme for static scenarios. Our code will be released upon acceptance.

\end{abstract}

\section{Introduction}

Spiking Neural Networks (SNNs), as the third generation of neural networks, are bio-inspired models that simulate information communication between biological neurons through discrete binary spike trains \cite{maass1997networks}. Their event-driven nature enables greater energy efficiency and biological plausibility, making them a promising alternative to traditional Artificial Neural Networks (ANNs) \cite{rathi2023exploring}. However, the binary nature of spike trains limits the network’s representational capacity, leading to a performance gap remaining relative to floating-point ANNs \cite{eshraghian2023training, wang2025adaptive}. Recent advances in training methods such as Spatio-Temporal Backpropagation (STBP) \cite{wu2018spatio}, Backpropagation Through Time (BPTT) \cite{werbos1990backpropagation}, and the emergence of Transformer-based SNNs have significantly improved performance on various downstream tasks, including image classification \cite{zhou2022spikformer, zhou2024qkformer}, semantic segmentation \cite{yao2023spike, yao2025scaling}, and object detection \cite{luo2024integer, yao2024spikedrivenv2}. However, a noticeable performance gap still persists.

To further narrow this gap, researchers have incorporated biologically inspired mechanisms into SNN design. Inspired by theta rhythms and phase precession in the hippocampus, Zhang et al. improved spatio-temporal integration in SNNs \cite{zhang2025toward}. MSD simulated bypass circuits in the optic nerve nucleus for multi-scale spike fusion \cite{li2025brain}, while STNet leveraged joint spatio-temporal processing mechanisms observed in the brain to enhance high-dimensional data modeling \cite{duan2025brain}. Despite their effectiveness, these approaches, based on convolutional neural networks (CNNs), often rely on the fine-grained design of specific neural mechanisms, which limits their scalability in complex tasks. By contrast, models such as SpiLiFormer \cite{zheng2025spiliformer}, TDFormer \cite{zhu2025tdformer}, and SpikingVTG \cite{bal2024spikingvtg}, inspired by top-down regulation in the brain, introduced feedback loops into Transformer-based SNNs to transmit high-level semantic information without modifying the core architecture. Their strong empirical performance and architectural flexibility have drawn increasing research attention \cite{wang2023complex, li2024spiking}.

\begin{figure*}[!th]
    \centering
    \setlength{\belowcaptionskip}{-0.2cm} 
    \includegraphics[width=0.95\textwidth]{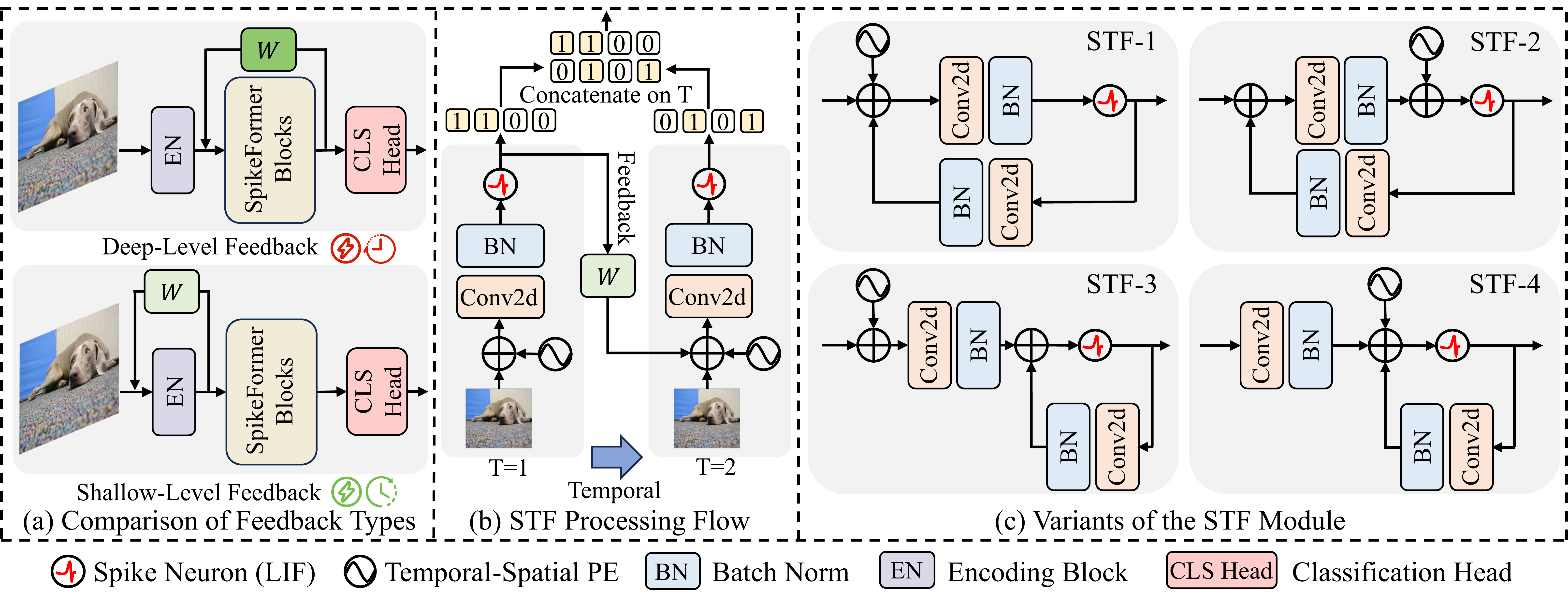}
    \caption{Illustration of the proposed Shallow-level Temporal Feedback (STF) module. (a) Comparison between deep-level and shallow-level feedback. The former incurs higher energy consumption, longer inference latency, and more complex feature transformations. $W$ denotes transformation weights, and darker colors represent higher feature dimensionality. (b) STF processing flow across timesteps. (c) Four architectural variants of the STF module.}
    \label{fig:figure_1}
\end{figure*}

However, as shown in Figure~\ref{fig:figure_1}(a), existing deep-level feedback loops are explicitly constructed in the network’s spatial domain, often being placed in deeper layers or spanning multiple hierarchical levels \cite{bal2024spikingvtg, zheng2025spiliformer}. Although this design effectively improves model performance, it inevitably introduces two forward passes, leading to increased energy consumption and inference latency. Moreover, as the network depth increases, feature dimensionality rises substantially, and processing feedback on high-dimensional features further increases the model’s parameter burden. This raises a natural question: \textit{Why not construct temporal-wise feedback loops directly in the shallow layers of the network?} Such a design could retain the benefits of feedback mechanisms while mitigating the computational overhead caused by high-dimensional feature transformations and minimizing the increase in energy consumption and inference latency caused by repeated forward passes.

\begin{figure*}[!th]
    \centering
    \setlength{\belowcaptionskip}{-0.3cm} 
    \includegraphics[width=0.95\textwidth]{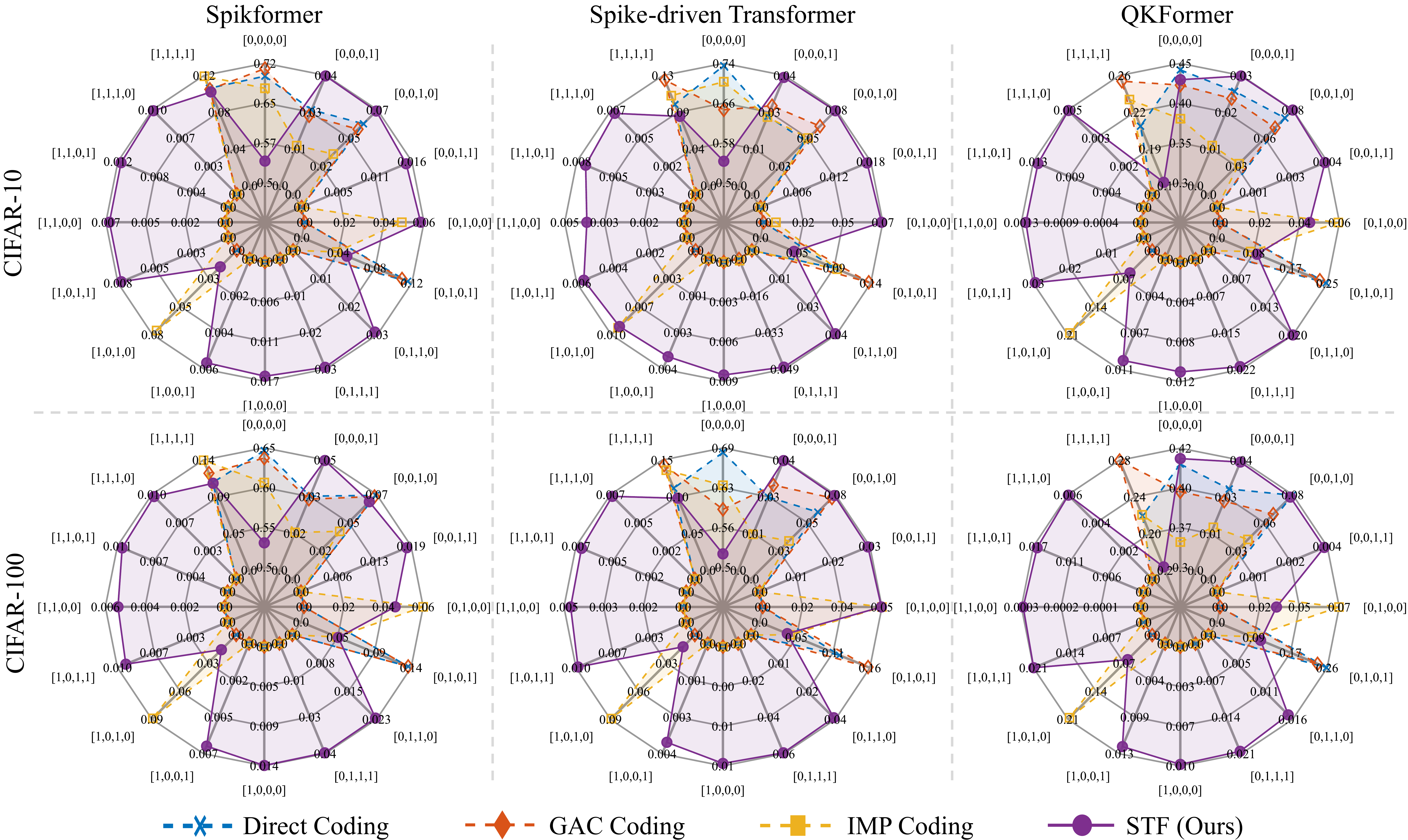}
    \caption{Distribution of spike patterns under different encoding schemes for various Transformer-based SNN backbones at $T=4$. Both direct coding and GAC coding tend to favor specific spike patterns, resulting in limited diversity. Although IMP coding mitigates this bias to some extent, it still fails to activate the full range of patterns compared to STF.}
    \label{fig:figure_2}
\end{figure*}

Building on this insight, we propose a Shallow-level Temporal Feedback (STF) module that can be seamlessly integrated into the encoding layer of existing models. As illustrated in Figure~\ref{fig:figure_1}(b), STF comprises Temporal-Spatial Position Embedding (TSPE) and Temporal Feedback (TF), which together introduce temporal feedback efficiently without compromising the spike-driven nature of SNNs.
We evaluate the impact of STF on several mainstream Transformer-based SNN architectures, including Spikformer \cite{zhou2022spikformer}, Spike-driven Transformer (SDT) \cite{yao2023spike}, and QKFormer \cite{zhou2024qkformer}. Experimental results demonstrate that STF consistently improves accuracy across different spike timesteps on static datasets such as CIFAR-10, CIFAR-100, and ImageNet-1K. Notably, on the large-scale ImageNet-1K dataset, STF improves accuracy by 0.75\%, with only a 0.32\% increase in parameters, 7.2\% more energy consumption, and 6.33\% higher inference latency. Further analysis, including visualizations (Figure~\ref{fig:figure_2}), quantitative comparisons based on spike entropy (Table~\ref{tab:cifar10_100_stf}), and theoretical analysis (Section~\ref{sec:reason_stf_gains}), reveals that the key factor behind STF’s performance gains lies in its ability to enhance the diversity of spike patterns. This enables richer spike-based representations, which are critical for static visual tasks. 
In addition, we conduct a detailed evaluation of STF against direct coding and its variants, focusing on model performance, robustness, and temporal information preservation. The comparative experiments suggest that STF represents a promising new spike encoding scheme, with the potential to replace the long-standing dominance of direct coding in static data scenarios. Our main contributions can be summarized as follows:

\begin{itemize}[label=\textbullet] 
    \item  We propose the STF module, a plug-and-play temporal feedback mechanism applied at the encoding layer. It preserves the spike-driven nature of SNNs by introducing feedback at low-dimensional feature layers and directly modulating spiking neurons. STF consistently improves model performance under varying spike timesteps across CIFAR-10, CIFAR-100, and ImageNet-1K, while incurring only modest increases in model parameters, energy consumption, and inference latency.
    \item Through detailed visualizations, quantitative comparisons, and theoretical analysis, we identify that the key to STF’s performance improvement lies in its ability to enhance the diversity of spike patterns, thereby improving spike-based representation. Besides, we systematically evaluate four architectural variants of STF and determine the optimal configuration.
    \item We perform a comprehensive evaluation of STF in comparison with direct coding and its variants, with respect to model performance, robustness, and the richness of temporal information. The detailed experimental results indicate that STF has strong potential as a novel spike encoding scheme for static data scenarios, with the capacity to replace the long-standing direct coding.
\end{itemize}

\section{Related Work}

\subsection{Transformer-based SNNs}

Due to their strong performance across a wide range of tasks, Transformer-based SNNs have garnered increasing attention from researchers, leading to continuous architectural innovations.
Spikformer \cite{zhou2022spikformer} was the first to integrate the Transformer architecture with SNNs, replacing the softmax operation with binary spike-based representations to effectively reduce energy consumption. Spike-driven Transformer \cite{yao2023spike, yao2024spikedrivenv2} further advanced the paradigm by substituting the dot-product attention with Hadamard product operations and redesigning residual connections based on membrane potentials to establish a fully spike-driven framework. 
QKFormer \cite{zhou2024qkformer} and SpikeGPT \cite{zhu2025spikegpt} introduced linear attention mechanisms to further reduce computational complexity, while E-SpikeFormer \cite{yao2025scaling} adopted integer-based training approaches to enhance both training efficiency and model performance. Despite these advances, most existing Transformer-based SNNs still rely on ANN-inspired designs, neglecting bio-inspired principles in architectural design that may unlock further performance gains.

\subsection{Feedback Mechanisms in SNNs}

Originally rooted in cybernetics \cite{wiener2019cybernetics} and later adopted in neuroscience \cite{lamme2000distinct}, feedback mechanisms provide strong self-adaptive regulation and have been widely integrated into SNNs. BackEISNN \cite{zhao2022backeisnn} and RSNN \cite{xu2024rsnn} applied recurrent feedback to all neurons and introduced additional hidden states. BRP \cite{zhang2021tuning} used cortical reward signal transmission to feed label information back to all hidden layers, enhancing accuracy through local pseudo-gradients. STSF \cite{he2025stsf} and STNet \cite{duan2025brain} adopted spatio-temporal feedback across cortical hierarchies to mitigate feature degradation and temporal loss. Besides, SpikingVTG \cite{bal2024spikingvtg} was the first to utilize high-layer firing rates to modulate attention in intermediate layers, improving performance in video-text temporal grounding. SpiLiFormer \cite{zheng2025spiliformer} and TDFormer \cite{zhu2025tdformer} incorporated lateral inhibition and top-down modulation, respectively, to address attention dispersion. 
However, these approaches primarily rely on structurally deep-level feedback loops, where high-level semantic information is propagated back to multiple or even all hidden layers through high-dimensional feature transformations. In contrast, the shallow-level and temporal feedback mechanism has not yet been thoroughly investigated.

\section{Methods}
This section introduces the spiking neuron model employed in our study and elaborates on the proposed Shallow-level Temporal Feedback (STF) module.

\subsection{Spiking Neuron}

Following prior related research \cite{zhou2022spikformer, yao2023spike, zhou2024qkformer, yao2025scaling}, we adopt the leaky integrate-and-fire (LIF) neuron \cite{maass1997networks} as the minimal processing unit. The dynamics of the LIF neuron are described by the following equations:

\begin{equation}
    H[t] = U[t-1] + \frac{1}{\tau_{\text{m}}}(I[t]-(U[t-1]-U_{\text{reset}})),
    \label{eq:lif_1}
\end{equation}

\begin{equation}
    S[t] = \Theta(H[t]-U_{\text{th}}),
    \label{eq:lif_2}
\end{equation}

\begin{equation}
    U[t] = H[t](1-S[t])+U_{\text{reset}}S[t],
    \label{eq:lif_3}
\end{equation}
where $H[t]$ denotes the intermediate membrane state at time $t$, which is influenced by the input current $I[t]$, the previous membrane potential $U[t-1]$, and the reset voltage $U_{\text{reset}}$. The output spike $S[t]$ is generated when $H[t]$ exceeds the threshold $U_{\text{th}}$, determined by the Heaviside function $\Theta(\cdot)$, after which the membrane potential is reset to $U_{\text{reset}}$.

{
\renewcommand{\arraystretch}{1.0}
\begin{table*}[t]
	\centering
        \setlength{\belowcaptionskip}{-0.1cm} 
	\begin{tabular}{lccccccccc}
		\toprule
		\multirow{3}{*}{Methods}          & \multirow{3}{*}{Architecture}     & \multicolumn{8}{c}{Acc (\%) / \textit{Spike Entropy}}                                                           \\ \cmidrule{3-10}
		                                  &                                   & \multicolumn{4}{c}{CIFAR-10}          & \multicolumn{4}{c}{CIFAR-100}             \\
		                                  &                                   & T=2     & T=4     & T=6     & T=8     & T=2      & T=4      & T=6      & T=8      \\ \midrule
		\multirow{2}{*}{Spikformer} & \multirow{2}{*}{Spikformer-4-384} & 93.20   & 95.19   & 95.14   & 95.24   & 75.17    & \underline{77.86}     & \underline{78.17}    & 78.22    \\
		                                  &     & \textit{0.964} & \textit{1.442} & \textit{1.530} & \textit{1.664} & \textit{1.203} & \textit{1.590} & \textit{1.774} & \textit{1.860} \\
                                          \\[-2.5ex]
                                         \cdashline{1-10} \\[-2ex]
                                          
        \multirow{2}{*}{Spikformer w/ GAC} & \multirow{2}{*}{Spikformer-4-384} & 93.1  & 94.63 & 95.03 & 95.17 & 75.17 & 77.16 & 77.90 & 78.54            \\
                                            &    & \textit{0.966} & \textit{1.396} & \textit{1.515} & \textit{1.627} & \textit{1.244} & \textit{1.609} & \textit{1.760} & \textit{1.821} \\
                                            \\[-2.5ex]
                                            \cdashline{1-10} \\[-2ex]

        \multirow{2}{*}{Spikformer w/ IMP} & \multirow{2}{*}{Spikformer-4-384} & \underline{94.08} & \underline{94.92} & \underline{95.42} & \underline{95.39} & \underline{75.52} & 77.34 & 78.16 & \underline{78.73} \\
                                            &  & \underline{\textit{1.406}} & \underline{\textit{1.670}} & \underline{\textit{1.713}} & \underline{\textit{1.739}} & \textbf{\textit{1.492}} & \underline{\textit{1.883}}& \underline{\textit{2.019}} & \underline{\textit{2.114}} \\
                                            \\[-2.5ex]
                                            \cdashline{1-10} \\[-2ex]

		\multirow{2}{*}{Spikformer w/ STF} & \multirow{2}{*}{Spikformer-4-384} & \textbf{94.95} & \textbf{95.61} & \textbf{95.85} & \textbf{95.99} & \textbf{76.94} & \textbf{78.00}  & \textbf{78.87} & \textbf{79.14} \\
		                                  &     & \textit{\textbf{1.463}} & \textit{\textbf{2.573}} & \textit{\textbf{4.369}} & \textit{\textbf{4.932}} & \underline{\textit{1.479}} & \textit{\textbf{2.597}} & \textit{\textbf{4.417}} & \textit{\textbf{5.240}} \\ \midrule

		\multirow{2}{*}{SDT}        & \multirow{2}{*}{SDT-2-512}        & 95.01 & 95.60 & 96.11 & 96.24 & 77.47 & 78.40 & 79.43 & 79.87 \\
		                                  &    & \textit{0.877} & \textit{1.318} & \textit{1.527} & \textit{1.626} & \textit{1.104} & \textit{1.482} & \textit{1.648} & \textit{1.733} \\
                                          \\[-2.5ex]
                                         \cdashline{1-10} \\[-2ex]

        \multirow{2}{*}{SDT w/ GAC}   & \multirow{2}{*}{SDT-2-512}  & \underline{95.18} & 95.69 & \underline{96.17} & \underline{95.97} & 77.65 & 79.10 & \underline{79.76} & \underline{80.04} \\
                                    &   &\underline{\textit{1.084}} & \underline{\textit{1.571}} & \textit{1.630} & \textit{1.698} &\textit{1.200} &\textit{1.710} & \textit{1.773} & \textit{1.928} \\
                                    \\[-2.5ex]
                                    \cdashline{1-10} \\[-2ex]

        \multirow{2}{*}{SDT w/ IMP}   & \multirow{2}{*}{SDT-2-512} & 95.15 & \underline{95.71} & 96.09 & 96.21 & \underline{77.8}  & \underline{79.22} & 79.65 & 79.98 \\
                                    &   & \textit{1.073} & \textit{1.501} & \underline{\textit{1.839}} & \underline{\textit{1.903}} & \textbf{\textit{1.396}} & \underline{\textit{1.797}} & \underline{\textit{1.952}} & \underline{\textit{2.019}} \\
                                    \\[-2.5ex]
                                    \cdashline{1-10} \\[-2ex]
            
		\multirow{2}{*}{SDT w/ STF}        & \multirow{2}{*}{SDT-2-512}        & \textbf{95.24} & \textbf{95.86} & \textbf{96.23} & \textbf{96.41} & \textbf{78.21} & \textbf{79.44} & \textbf{80.34} & \textbf{80.61} \\
		                                  &     & \textbf{\textit{1.318}} & \textbf{\textit{2.514}} & \textbf{\textit{3.600}} & \textbf{\textit{4.484}} & \underline{\textit{1.381}} & \textbf{\textit{2.609}} & \textbf{\textit{3.720}} & \textbf{\textit{4.439}} \\ \midrule
                                         
		\multirow{2}{*}{QKFormer}   & \multirow{2}{*}{HST-4-384}        & \underline{95.79} & \underline{96.18} & \underline{96.37} & 96.35 & 79.79 & \underline{81.15} & 81.35 & 81.64 \\
		                                  &    & \textit{1.483} & \textit{1.911} & \textit{2.080} & \textit{2.116} & \textit{1.490} & \textit{1.956} & \textit{2.114} & \textit{2.134} \\
                                          \\[-2.5ex]
                                          \cdashline{1-10} \\[-2ex]

        \multirow{2}{*}{QKFormer w/ GAC}   & \multirow{2}{*}{HST-4-384}  & 95.76 & 96.16 & 96.32 & \underline{96.37} & 79.92 & 80.88 & \underline{81.40} & 81.59 \\
                                        &  & \textit{1.469} & \textit{1.906} & \textit{2.066} & \textit{2.120} & \textit{1.490} & \textit{1.920} & \textit{2.056} & \textit{2.139} \\
                                        \\[-2.5ex]
                                        \cdashline{1-10} \\[-2ex]

        \multirow{2}{*}{QKFormer w/ IMP}   & \multirow{2}{*}{HST-4-384} & 95.28 & 96.03 & 96.35 & 96.31 & \underline{80.01} & 81.10 & 81.30 & \underline{81.69} \\
                                        &  & \textbf{\textit{1.810}} & \underline{\textit{2.245}} & \underline{\textit{2.412}} & \underline{\textit{2.398}} & \textbf{\textit{1.833}} & \underline{\textit{2.325}} & \underline{\textit{2.389}} & \underline{\textit{2.488}} \\
                                        \\[-2.5ex]
                                        \cdashline{1-10} \\[-2ex]

        \multirow{2}{*}{QKFormer w/ STF}   & \multirow{2}{*}{HST-4-384}   & \textbf{96.02} & \textbf{96.33} & \textbf{96.51} & \textbf{96.61} & \textbf{80.07} & \textbf{81.26} & \textbf{81.51} & \textbf{81.89} \\
		                                  &        & \underline{\textit{1.676}} & \textit{\textbf{2.809}} & \textit{\textbf{4.028}} & \textit{\textbf{4.862}} & \underline{\textit{1.722}} & \textit{\textbf{2.819}} & \textit{\textbf{4.054}} & \textit{\textbf{5.076}}  \\ \bottomrule
	\end{tabular}
	\caption{Performance comparison of our method on $3$ backbones: Spikformer, Spike-driven Transformer (SDT), and QKFormer, on the CIFAR datasets across various time steps. Bold indicates the best result, and underlined denotes the second-best.}
	\label{tab:cifar10_100_stf}
\end{table*}
}

\subsection{Shallow-Level Temporal Feedback Module}
\label{sec:STF}

As illustrated in Figure~\ref{fig:figure_1}(b), the STF module consists of two key components, namely Temporal-Spatial Position Embedding (TSPE) and Temporal Feedback (TF).

\subsubsection{Temporal-Spatial Position Embedding} To effectively capture temporal and spatial dynamics in spike trains, position embedding plays a crucial role in SNNs \cite{lv2024advancing}. In our formulation, we represent the input image sequence as $I \in \mathbb{R}^{T\times C \times H \times W}$, where $T$ is the number of spike timesteps, and $C$, $H$, and $W$ refer to the number of channels, image height, and image width, respectively. In STF-1 and STF-3 configurations, as shown in Figure~\ref{fig:figure_1}(c), the TSPE mechanism   is formulated as follows:

\begin{equation}
    I_{\text{embed}}[t] = W_{\text{ConvBN}} \cdot (X_{\text{TPE}}[t] + I[t]),
    \label{eq:TSPE_1}
\end{equation}
where $X_{\text{TPE}} \in \mathbb{R}^{T\times C \times H \times W}$ represents a spatio-temporal positional embedding, and $W_{\text{ConvBN}}$ denotes the weights of a 2D convolution layer followed by batch normalization. 
Motivated by the inherent similarity between spike trains and sequential video signal processing, a 3D trigonometric positional strategy is adopted to initialize $X_{\text{TPE}}$, following common practice in video transformers \cite{liu2022video, liang2024vrt}. To explore the optimal location for incorporating TSPE, two additional variants, STF-2 and STF-4, are introduced. In these configurations, the embedding formulation is modified as follows:

\begin{equation}
    I_{\text{embed}}[t] = X_{\text{TPE}}[t] + W_{\text{ConvBN}}	\cdot I[t].
    \label{eq:TSPE_2}
\end{equation}

\subsubsection{Temporal Feedback} To maintain the spike-driven nature of the network, TF directly feeds the spike output from the previous time step $S[t-1]$ into the input at the current time step $S[t]$. This forms a feedback loop that captures temporal dependencies. 
In contrast to conventional recurrent SNNs \cite{zhao2022backeisnn, xu2024rsnn}, TF avoids the use of auxiliary hidden states or non-spiking activations. The feedback operation in STF-1 and STF-2, as illustrated in Figure~\ref{fig:figure_1}(c), is formulated as:

\begin{equation}
    I'[t] = I[t] + W_{\text{TF}} \cdot S[t-1],
    \label{eq:twf_1}
\end{equation}
where $W_{\text{TF}}$ denotes a learnable transformation comprising convolution and batch normalization, and $I'[t]$ is the input at time $t$ after integrating the feedback signal. 
To further explore alternative feedback strategies, STF-3 and STF-4 adopt a more direct approach, in which the previous output is used to update the membrane potential of the LIF neuron. This process is described by the following equation:

\begin{equation}
    H[t] = (1-\frac{1}{\tau_m})U[t-1] + I[t] + W_{\text{TF}} \cdot  S[t-1],
    \label{eq:twf_2}
\end{equation}
where all variables strictly adhere to the definitions provided in Equations~\ref{eq:lif_1}–\ref{eq:lif_3}.

\section{Experiments}

In this section, we conduct a series of experiments to investigate the following research questions: \textbf{RQ1:} Can the proposed STF module effectively enhance the performance of Transformer-based SNNs on image classification tasks across varying spike timesteps, while maintaining modest increases in model parameters, energy consumption, and inference latency? \textbf{RQ2:} What are the underlying factors that contribute to the performance gains achieved by STF? \textbf{RQ3:} Can STF serve as a novel spiking encoding scheme that replaces the existing direct coding and its variants?

{
\renewcommand{\arraystretch}{0.9}
\begin{table*}[th]
        \setlength{\tabcolsep}{1mm}
        \setlength{\belowcaptionskip}{-0.2cm} 
	\centering
	\begin{tabular}{lccccccc}
		\toprule
		Methods                         & SNN & Architecture         & Input Size & Param(M) & Power(mJ) & Time Step & Top-1 Acc(\%) \\ \midrule
		ViT                             & \ding{55}   & ViT-B/16             & 224        & 86.59    & 254.84    & 1         & 77.90         \\
		\multirow{2}{*}{Swin}           & \ding{55}   & Swin Transformer-B   & 224        & 87.77    & 70.84     & 1         & 83.50         \\
		                                & \ding{55}   & Swin Transformer-B     & 384        & 87.77    & 216.20    & 1         & 84.50         \\\noalign{\vskip 0.1ex}\hdashline \noalign{\vskip 0.5ex}
		SEW ResNet                      & \ding{51}   & SEW-ResNet-152       & 224        & 60.19    & 12.89     & 4         & 69.26         \\
		Spikformer                      & \ding{51}   & Spikformer-8-768     & 224        & 66.34    & 21.48     & 4         & 74.81         \\
		Spikingformer                   & \ding{51}   & Spikingformer-8-768  & 224        & 66.34    & 13.68     & 4         & 75.85         \\
		\multirow{3}{*}{E-SpikeFormer}  & \multirow{3}{*}{\ding{51}}   & E-SpikeFormer-12-768 & 224        & 173.0    & 35.6      & \ \ 4\textsuperscript{$\star$}   & 84.70          \\
		                                &    & E-SpikeFormer-12-768 & 224        & 173.0    & 54.7      & \ \ 8\textsuperscript{$\star$}  & 85.10          \\
		                                &   & \ \ \ E-SpikeFormer-12-768\textsuperscript{**} & 384        & 173.0    & -         & \ \ 8\textsuperscript{$\star$}   & 86.20          \\ \midrule
		SDT                             & \ding{51}   & SDT-6-512            & 224        & 23.37    & 3.56      & 4         & \;72.41\textsuperscript{$\dagger$}        \\\noalign{\vskip 0.1ex} \hdashline \noalign{\vskip 0.5ex}
		SDT w/ STF                       & \ding{51}   & SDT-6-512            & 224        & 23.41    & 3.99      & 4         & \textbf{73.01}         \\ \midrule
		\multirow{3}{*}{\raisebox{-0.8ex}{QKFormer}}       & \multirow{3}{*}{\raisebox{-0.3ex}{\ding{51}}}   & HST-10-768           & 224        & 64.96    & 38.91     & 4         & 84.22         \\
		                                &    & \ \ HST-10-768\textsuperscript{*}   & 288        & 64.96    & 64.27     & 4         & 85.20         \\
		                                &    & \;\;\;HST-10-768\textsuperscript{**}   & 384        & 64.96    & 113.64    & 4         & 85.65         \\\noalign{\vskip 0.1ex}\hdashline \noalign{\vskip 0.5ex}
		\multirow{3}{*}{\raisebox{-0.8ex}{QKFormer w/ STF}} & \multirow{3}{*}{\raisebox{-0.3ex}{\ding{51}}}   & HST-10-768           & 224        & 65.17    & 41.75     & 4         & \textbf{84.97}         \\
		                                &    & \ \ HST-10-768\textsuperscript{*}            & 288        & 65.17    & 68.96     & 4         & \textbf{85.89}         \\
		                                &    & \;\;\;HST-10-768\textsuperscript{**}           & 384        & 65.17    & 121.77    & 4         & \textbf{86.26}         \\ \bottomrule
	\end{tabular}
	\caption{Performance comparison on ImageNet-1K. Numbers marked with $\dagger$ indicate our implementation based on publicly available code. $*$ and $**$ indicate inference resolutions of $288^2$ and $384^2$, respectively. $\star$ represents a quantized training method \cite{yao2025scaling} using integer-only activations, which are later mapped to spike trains for inference.}
	\label{tab:imagenet_1k_stf}
\end{table*}
}

\subsection{Datasets and Baseline Models}

To answer the above research questions, we select representative Transformer-based SNNs, including Spikformer \cite{zhou2022spikformer}, Spike-driven Transformer (SDT) \cite{yao2023spike}, and QKFormer \cite{zhou2024qkformer}, as baseline models. Following \cite{yao2024spikedrivenv2, zhou2024qkformer}, we use three standard image classification datasets, CIFAR-10 \cite{krizhevsky2009learning}, CIFAR-100 \cite{krizhevsky2009learning}, and ImageNet-1K \cite{deng2009imagenet}, to evaluate the performance gains introduced by STF. Details of these datasets are provided in Appendix A. In addition, to further investigate \textbf{RQ3}, we compare STF with existing spiking encoding schemes, including direct coding \cite{wu2019direct} and its variants such as GAC coding \cite{qiu2024gated} and IMP coding \cite{shen2024rethinking}. The training configuration and hyperparameter settings are detailed in Appendix B.

{
\renewcommand{\arraystretch}{0.95}
\begin{table*}[h]
\setlength{\tabcolsep}{1mm}
\centering
\setlength{\belowcaptionskip}{-0.2cm} 
\begin{tabular}{lccccccccc}
\toprule
\multicolumn{1}{l}{\multirow{3}{*}{Methods}} & \multicolumn{8}{c}{Inference Time per Sample (ms)}            \\ \cmidrule{2-10}
\multicolumn{1}{c}{}                         & \multicolumn{4}{c}{CIFAR-10}  & \multicolumn{4}{c}{CIFAR-100} \\
\multicolumn{1}{c}{}                         & T=2   & T=4   & T=6   & T=8   & T=2   & T=4   & T=6   & T=8   \\ \midrule
Spikformer                                   & 0.384 & 0.499 & 0.718 & 0.891 & 0.389 & 0.525 & 0.725 & 0.932 \\
\multirow{2}{*}{Spikformer w/ STF}            & 0.404 & 0.541 & 0.777 & 0.962 & 0.395 & 0.532 & 0.749 & 0.964 \\
 &
  \textbf{(+5.21\%)} &
  \textbf{(+8.42\%)} &
  \textbf{(+8.22\%)} &
  \textbf{(+7.97\%)} &
  \textbf{(+1.54\%)} &
  \textbf{(+1.33\%)} &
  \textbf{(+3.31\%)} &
  \textbf{(+3.34\%)} \\ [-0.1ex] \cdashline{1-10} \\[-1.8ex]
SDT                                          & 0.319 & 0.521 & 0.711 & 0.886 & 0.304 & 0.503 & 0.692 & 0.884 \\
\multirow{2}{*}{SDT  + STF}                  & 0.337 & 0.550 & 0.745 & 0.928 & 0.317 & 0.531 & 0.728 & 0.919 \\ 
 &
  \textbf{(+5.64\%)} &
  \textbf{(+5.57\%)} &
  \textbf{(+4.78\%)} &
  \textbf{(4.74\%)} &
  \textbf{(+4.28\%)} &
  \textbf{(+5.57\%)} &
  \textbf{(+5.2\%)} &
  \textbf{(+3.96\%)} \\ [-0.1ex] \cdashline{1-10} \\[-1.8ex]
QKFormer                                     & 0.715 & 0.726 & 1.012 & 1.324 & 0.547 & 0.763 & 0.984 & 1.270 \\
\multirow{2}{*}{QKFormer w/ STF}              & 0.769 & 0.771 & 1.083 & 1.414 & 0.592 & 0.788 & 1.035 & 1.387 \\
 &
  \textbf{(+7.55\%)} &
  \textbf{(+6.20\%)} &
  \textbf{(+7.02\%)} &
  \textbf{(+6.8\%)} &
  \textbf{(+8.23\%)} &
  \textbf{(+3.28\%)} &
  \textbf{(+5.18\%)} &
  \textbf{(+9.21\%)} \\ [-0.1ex] \cdashline{1-10} \\[-1.8ex]
SpiLiFormer w/o feedback                           & 0.726 & 0.766 & 0.995 & 1.313 & 0.556 & 0.697 & 0.999 & 1.303 \\
\multirow{2}{*}{SpiLiFormer}           & 0.919 & 0.932 & 1.179 & 1.521 & 0.698 & 0.858 & 1.228 & 1.505 \\
 &
  \textbf{(+26.58\%)} &
  \textbf{(+21.67\%)} &
  \textbf{(+18.49\%)} &
  \textbf{(+15.84\%)} &
  \textbf{(+25.54\%)} &
  \textbf{(+23.1\%)} &
  \textbf{(+22.92\%)} &
  \textbf{(+15.50\%)} \\ \bottomrule
\end{tabular}
\caption{Comparison of inference latency overhead between the shallow-level STF module and traditional deep-level feedback on CIFAR datasets. Corresponding results on ImageNet-1K are provided in Appendix F.}
\label{tab:inference_latency}
\end{table*}
}

\subsection{Effectiveness of STF on Image Classification Tasks}

\begin{figure}[!h]
  \centering
  \setlength{\belowcaptionskip}{-0.2cm} 
  \includegraphics[width=0.45\textwidth]{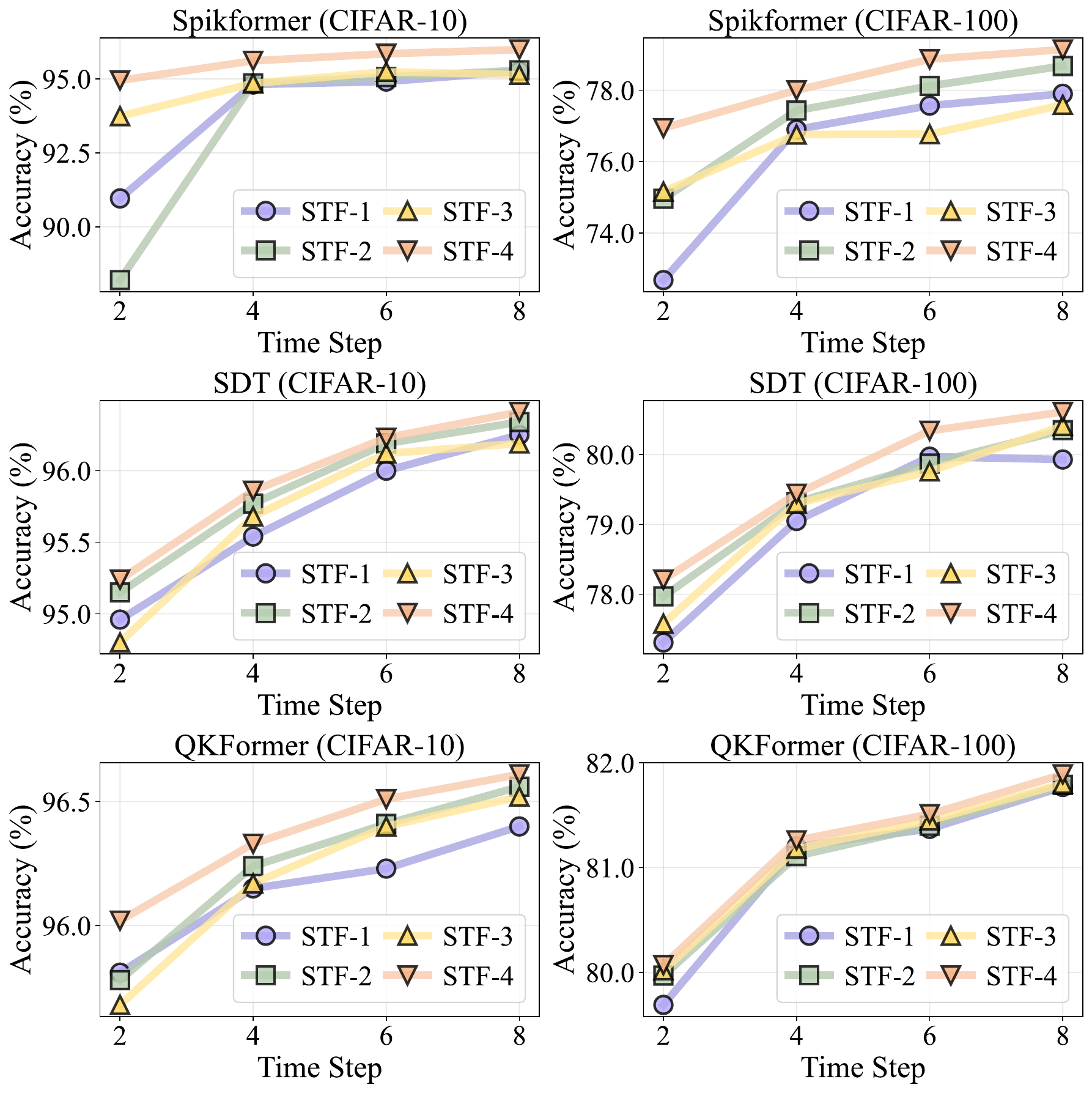}
  \caption{Performance comparison of four STF variants (Figure~\ref{fig:figure_1}(c)) on CIFAR datasets across various timesteps.}
  \label{fig:stf_varients}

\end{figure}

Before this, we conduct a systematic evaluation of the four STF variants defined in Section~\ref{sec:STF} and illustrated in Figure~\ref{fig:figure_1}(c) to explore the optimal STF architecture. As shown in Figure~\ref{fig:stf_varients}, STF-4 consistently achieves the best performance across different datasets and spike timesteps. Therefore, unless otherwise specified, we adopt STF-4 as the default configuration throughout this paper.

To answer \textbf{RQ1}, we train various models from scratch and evaluate the performance improvements introduced by STF across three datasets and three Transformer-based SNN backbones. 
As shown in Table~\ref{tab:cifar10_100_stf}, STF consistently outperforms the corresponding baselines on both CIFAR-10 and CIFAR-100 across four different timesteps. Specifically, on CIFAR-100, STF improves performance by 1.77\% with Spikformer at $T=2$ and by 1.04\% with SDT at $T=4$.

We further evaluate STF on the more challenging ImageNet-1K dataset to assess its performance improvement over SDT and QKFormer, along with the additional parameter and energy consumption overhead. 
As shown in Table~\ref{tab:imagenet_1k_stf}, STF improves accuracy by 0.6\% to 0.75\%. Specifically, for SDT with the timestep of $T=4$, STF achieves a 0.6\% accuracy gain while introducing only a 0.17\% increase in parameters and a 12.08\% increase in energy consumption. Similarly, for QKFormer with $T=4$ and a standard input resolution of $224$, STF improves accuracy by 0.75\% with just a 0.32\% increase in parameters and approximately 7.2\% more energy consumption. Furthermore, STF allows QKFormer, which uses only 37.5\% of the parameters of the best-performing E-SpikeFormer, to reach comparable accuracy (86.2\%) with only half the number of timesteps.

It is worth noting that the temporal feedback loop in STF introduces challenges for parallel computation. To quantify the impact of the feedback structure on the inference efficiency, we analyze the inference latency introduced by STF. Although feedback mechanisms inherently add some computational overhead, STF employs a shallow-level feedback design that exhibits significant advantages over deep-level feedback structures such as SpiLiFormer \cite{zheng2025spiliformer}. As shown in Table~\ref{tab:inference_latency}, the latency overhead introduced by STF remains below 10\% and is often reduced by half or more compared to SpiLiFormer.

\subsection{Analyzing the Performance Gains of STF}
\label{sec:reason_stf_gains}

To answer \textbf{RQ2}, we analyze the distribution of spike trains from the same LIF neuron in the encoding layer, both with and without STF, on the CIFAR datasets. Specifically, we visualize the output patterns under STF and the direct coding scheme for comparison. 
As shown in Figure~\ref{fig:figure_2}, taking $T=4$ as an example, STF alleviates the imbalance in the distribution of spike patterns to varying degrees between different main SNN models. It significantly increases the number of activated spike patterns and successfully covers all 16 possible binary combinations. 
A more diverse spike pattern indicates greater information capacity, facilitating more effective information propagation through the network and ultimately enhancing model representation and performance. Therefore, we consider that the improved diversity of spike patterns is a key contributor to STF performance gains.

To quantitatively measure this property, we adopt spike entropy as a diversity metric, inspired by previous work in neuroscience and information theory \cite{luczak2024entropy}. The formal definition is provided in Appendix H. As reported in Table~\ref{tab:cifar10_100_stf}, STF consistently improves the diversity of spike patterns across all models and datasets, with greater gains at higher spike timesteps.

We further provide a theoretical analysis to explain why STF can improve the diversity of spike patterns. For static datasets, the input image $I$ is duplicated across $T$ spike timesteps to form a sequence $I = [I[1], I[2], ..., I[T]]$, where $I[t] = I$ for all $t \in {1, 2, ..., T}$. The spike generation time $SG_t$ of a neuron in the encoding layer without STF can be formulated as follows (the detailed derivation is provided in Appendix G):
\begin{equation}
SG_t = \left\lceil \log_{\tau} (1-\frac{U_{\text{th}}(1 - \tau)}{I}) \right\rceil.
\label{equ:generate_spike_time}
\end{equation}
This expression indicates that under fixed values of leakage decay factor $\tau$, $U_{\text{th}}$, and a repeated static input $I$, baseline models tend to activate only a limited set of specific spike patterns. In contrast, STF incorporates both the current input $I[t]$ and the previous spike output $S[t{-}1]$ into Equation~\ref{eq:twf_2}, allowing $SG_t$ to be triggered over a broader temporal range, rather than being constrained by the constant input $I$. This effectively expands the expressive space of spike patterns, leading to enhanced diversity in spike patterns. In addition, we conduct detailed ablation studies on TSPE and TF, as presented in Appendix J. The results demonstrate that both components significantly contribute to improving the diversity of spike patterns and overall model performance, with TF exhibiting a more pronounced effect.

\subsection{Comparison with Existing Spiking Encoding Schemes}

To answer \textbf{RQ3}, we evaluate two representative spike encoding variants, GAC coding \cite{qiu2024gated} and IMP coding \cite{shen2024rethinking}, in terms of model performance and the diversity of spike patterns on the CIFAR datasets. As illustrated by the spike pattern visualizations in Figure~\ref{fig:figure_2} and the results in Table~\ref{tab:cifar10_100_stf}, GAC provides performance improvements for certain models. However, it exhibits a strong preference for specific spike patterns, similar to the direct coding scheme, resulting in a concentrated activation distribution. 
In contrast, IMP demonstrates a more pronounced effect in increasing the diversity of spike patterns, which contributes to performance gains. Nevertheless, it still fails to activate the full range of possible spike combinations, indicating limited diversity. 
STF, on the other hand, consistently outperforms both GAC and IMP in terms of the diversity of the spike patterns, leading to further improvements in model accuracy. Although STF yields slightly lower spike entropy values than IMP under certain timestep \mbox{configurations}, its advantage becomes increasingly evident as the number of timesteps grows, demonstrating better \mbox{scalability} and \mbox{stability}.

\begin{figure}[h]
  \centering
  \setlength{\belowcaptionskip}{-0.2cm} 
  \includegraphics[width=0.45\textwidth]{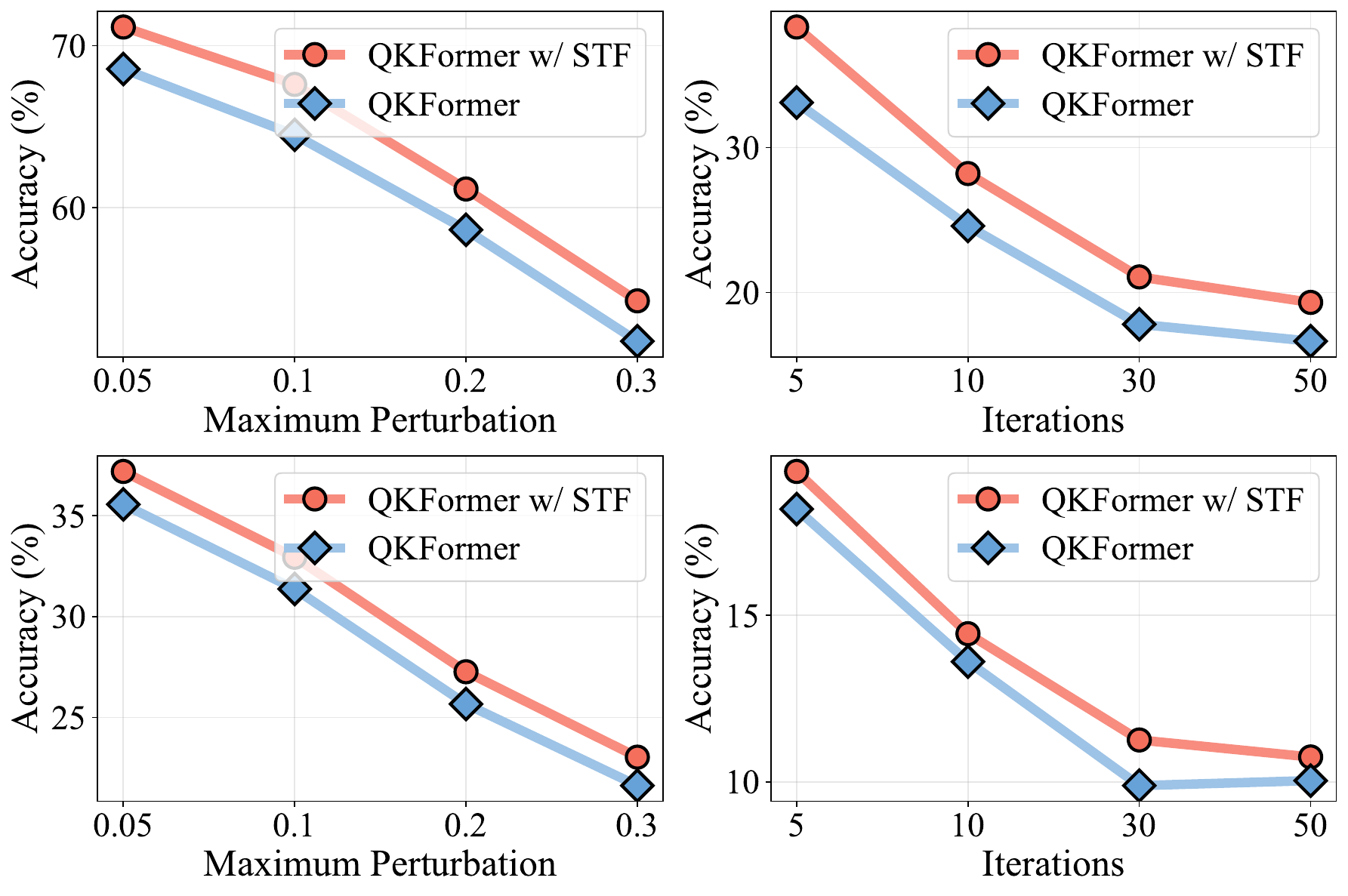}
  \caption{Adversarial robustness of QKFormer (4 timesteps) with and without STF on CIFAR-10 (top) and CIFAR-100 (bottom) under FGSM (left) and PGD (right) attacks.}
  \label{fig:adv}
\end{figure}

An effective encoding scheme should not only enhance model performance but also improve robustness against adversarial perturbations and capture richer temporal information \cite{kim2022rate, eshraghian2023training}. To evaluate whether STF meets these criteria, we first employ white-box adversarial attacks, including FGSM \cite{goodfellow2014explaining} and PGD \cite{madry2017towards}, to comprehensively assess the impact of STF on model robustness under various settings, compared to direct coding. As shown in Figure~\ref{fig:adv}, STF significantly improves robustness without compromising clean accuracy. Additional results of the robustness evaluation are provided in Appendix I. Then, following the previous study \cite{bu2023rate}, we employ a random number generator to shuffle the spike trains of neuron outputs, in order to examine whether STF captures richer temporal information. As shown in Table~\ref{tab:temporal_information}, STF-based models are more sensitive to temporal disruptions compared to those using direct coding. Combined with the findings of Bu et al. \cite{bu2023rate}, these results confirm that STF effectively encodes richer temporal information.

{
\renewcommand{\arraystretch}{0.4}
\begin{table}[h]
\centering
\setlength{\belowcaptionskip}{-0.2cm} 
\begin{tabular}{lccc}
\toprule
\multirow{2}{*}{Methods}        & \multirow{2}{*}{$T$} & Clean   & Shuffled     \\
                                &                    & Acc(\%) & Acc(\%)      \\  \midrule
\multirow{4}{*}{\raisebox{-1.9ex}{QKFormer}}       & 2                  & 79.79   & 79.01(-0.78) \\
                                & 4                  & 81.15   & 80.38(-0.77) \\
                                & 6                  & 81.35   & 80.76(-0.59) \\
                                & 8                  & 81.64   & 81.12(-0.52) \\ \midrule
\multirow{4}{*}{\raisebox{-1.9ex}{QKFormer w/ STF}} & 2                  & 80.07   & 79.10(-0.97) \\
                                & 4                  & 81.26   & 80.06(-1.2)  \\
                                & 6                  & 81.51   & 80.76(-0.75) \\
                                & 8                  & 81.89   & 81.05(-0.84) \\  \bottomrule
\end{tabular}
\caption{Model performance comparison before and after applying spike shuffling.}
\label{tab:temporal_information}
\end{table}
}

Besides, we further assess the applicability of STF to neuromorphic datasets. As shown in Table~5 in the appendix, STF leads to performance degradation on CIFAR10-DVS \cite{li2017cifar10} across different timestep settings, which aligns with our expectations. As analyzed in Section~\ref{sec:reason_stf_gains}, the effectiveness of STF in static tasks stems from its ability to break the limitation of Equation~\ref{equ:generate_spike_time}, thereby enhancing the diversity of the spike patterns. By contrast, event-based data inherently contain temporal structure, and the use of STF may interfere with this built-in temporal information, potentially degrading performance. 
Therefore, while STF shows promise as a new spike encoding scheme that can replace direct coding and its variants, its advantage appears to be limited to static scenarios.

\section{Conclusion}

In this paper, we propose Shallow-level Temporal Feedback (STF), a lightweight module that can be seamlessly integrated into the encoding layer of SNNs. Unlike deep-level feedback loops, STF avoids high-dimensional transformations and redundant forward passes. Experimental results show that STF consistently improves the performance of various Transformer-based SNN backbones under different timestep settings, with only a modest increase in the model size, energy consumption, and inference latency. Further analysis reveals that the performance gain can be primarily attributed to the enhanced diversity of spike patterns. Moreover, we compare STF with direct coding and its variants in terms of adversarial robustness and temporal information, showing that STF is promising as a novel spike encoding scheme for static scenarios. However, additional experiments reveal that STF exhibits suboptimal performance on the neuromorphic dataset. In future work, we will aim to improve its adaptability to event-based scenarios.

\bibliography{aaai2026}

\clearpage

\appendix

\section*{Appendix}

\section{Dataset Description}
\label{sec:dataset_description}

We conduct our experiments on three widely adopted datasets: ImageNet-1K \cite{deng2009imagenet}, CIFAR-10 \cite{krizhevsky2009learning}, and CIFAR-100 \cite{krizhevsky2009learning}.

\textbf{ImageNet-1K}: The ImageNet-1K dataset, also known as the ILSVRC benchmark, is a large-scale dataset that has become a cornerstone in visual recognition tasks. It includes approximately 1.28 million training samples drawn from 1,000 object categories, with an additional 50,000 images for validation and 100,000 for testing.

\textbf{CIFAR-10}: As a classic benchmark in image classification, CIFAR-10 comprises 60,000 colored images at a resolution of $32 \times 32$ pixels. The dataset is evenly split into 10 distinct categories, with 6,000 images per class.

\textbf{CIFAR-100}: Sharing the same structure as CIFAR-10, CIFAR-100 presents a more complex challenge by increasing the number of classes to 100. It also consists of 60,000 color images of $32 \times 32$ pixels, with 600 samples per class, making it suitable for evaluating fine-grained recognition capabilities.

\section{Training Configuration and Hyperparameter Settings}

To ensure a fair experimental comparison, we follow the experimental settings of previous works \cite{zhou2022spikformer, yao2023spike, zhou2024qkformer} for the three Transformer-based SNN backbones: Spikformer, Spike-driven Transformer (SDT), and QKFormer, evaluated on the datasets described in Section~\ref{sec:dataset_description}.

\subsection{Spikformer}

We follow the protocol of Zhou et al. \cite{zhou2022spikformer} and train all models from scratch on the CIFAR-10 and CIFAR-100 datasets for 300 epochs using the AdamW optimizer, with an initial learning rate of $5e-4$, a weight decay of $6e-2$, and a batch size of 128. A cosine annealing learning rate scheduler is employed, with 20 warm-up epochs and 10 cool-down epochs. To ensure reproducibility, the random seed is set to 42.

\subsection{SDT}

Following the work of Yao et al. \cite{yao2023spike}, we adopt SDT-2-512 and SDT-6-512 to train models from scratch on the CIFAR datasets and ImageNet-1K, respectively. Especially on the large-scale ImageNet-1K dataset, we adopt a data augmentation strategy during training, including RandAugment (with a magnitude of 7 and a standard deviation of 0.5), random resized cropping, horizontal flipping, and color jittering. In addition, Mixup and CutMix are applied with a probability of 0.6 to improve generalization. Furthermore, each image is augmented three times per epoch to enhance data diversity. Detailed hyperparameters and training settings are provided in Table~\ref{tab:SDT_hyper}.

{
\setcounter{table}{0}
\begin{table}[!h]
\centering
\begin{tabular}{ccc}
\toprule
Hyperparameter  & CIFAR Datasets         & ImageNet-1K          \\ \midrule
Head             & \multicolumn{2}{c}{8}                   \\
Layer            & 2                    & 6                    \\
Embed dim        & 512                  & 512                  \\
Min lr           & $1e-5$               & $2e-5$                  \\
Warm-up lr       & $1e-6$               & $1e-5$                     \\
Image size       & 32                   & 224 \\
Lr               &  $3e-4$              & $1e-3$              \\
Timestep         &  2/4/6/8                    &    4                  \\
Batch size       &    64                  &      48                \\
Epochs           &    \multicolumn{2}{c}{300}             \\
Weight decay     &    $6e-2$                  &  $1e-2$                    \\
Optimizer        &    AdamW                  &   LAMB                   \\
Warm-up epochs   &     \multicolumn{2}{c}{20}                 \\
Cool-down epochs &     \multicolumn{2}{c}{10}               \\
Random seed      &      \multicolumn{2}{c}{42}      \\ 
Lr scheduler     &  \multicolumn{2}{c}{Cosine} \\ \bottomrule
\end{tabular}
\caption{Training hyperparameters for SDT on the CIFAR datasets and ImageNet-1K}
\label{tab:SDT_hyper}
\end{table}
}

\subsection{QKFormer}

Following Zhou et al. \cite{zhou2024qkformer}, we train the network models from scratch. Detailed hyperparameters and training settings are provided in Table~\ref{tab:QKFormer_hyper}. Additionally, for ImageNet-1K, we follow the data augmentation and training strategy of Zheng et al. \cite{zheng2025spiliformer}.
{
\setcounter{table}{1}
\begin{table}[!h]
\centering
\begin{tabular}{ccc}
\toprule
Hyperparameter   & CIFAR Datasets         & ImageNet-1K          \\ \midrule
Head             &   8     &  12              \\
Layer            & 4                    & 10                    \\
Embed dim        & 384                  & 768                  \\
Min lr           & $1e-5$               & $1e-6$                  \\
Warm-up lr       & $1e-5$               & $1e-6$                     \\
Image size       & 32                   & 224 \\
Lr               &  $1e-3$              & $6e-4$              \\
Timestep         &  2/4/6/8             &    4                  \\
Batch size       &    64                &  \ \;512\textsuperscript{$\ddagger$}  \\
Epochs           &   400                & 200     \\
Weight decay     &    $6e-2$            &  $5e-2$                    \\
Optimizer        &     \multicolumn{2}{c}{AdamW}                   \\
Warm-up epochs   &     20               &       5                 \\
Cool-down epochs &   10                 &   5             \\
Random seed      &   \ 42               &  0               \\
Lr scheduler     &  \multicolumn{2}{c}{Cosine} \\ \bottomrule
\end{tabular}
\caption{Training hyperparameters for QKFormer on the CIFAR datasets and ImageNet-1K. $\ddagger$ indicates the use of accumulated gradient iterations \cite{he2022masked}.}
\label{tab:QKFormer_hyper}
\end{table}
}

{
\setcounter{table}{2}
\begin{table*}[!h]
\centering
\begin{tabular}{lccccccccc}
\toprule
\multirow{3}{*}{Methods} & \multirow{3}{*}{Architecture}     & \multicolumn{8}{c}{Acc (\%)}                                 \\ \cmidrule{3-10}
                         &                                   & \multicolumn{4}{c}{CIFAR-10} & \multicolumn{4}{c}{CIFAR-100} \\
                         &                                   & T=2   & T=4   & T=6   & T=8  & T=2   & T=4   & T=6   & T=8   \\ \midrule
Spikformer w/ STF-1       & \multirow{4}{*}{Spikformer-4-384} &90.96 & 94.81 & 94.91 & 95.26 & 72.68 & 76.90 & 77.57 & 77.90      \\
Spikformer w/ STF-2       &      & 88.20 & \underline{94.85} & 95.06 & \underline{95.29} & 74.96 & \underline{77.43} & \underline{78.12} & \underline{78.68}      \\
Spikformer w/ STF-3       &      & \underline{93.74} & \underline{94.85} & \underline{95.24} & 95.14 & \underline{75.16}& 76.76 & 76.77 & 77.59      \\
Spikformer w/ STF-4 &  & \textbf{94.95} & \textbf{95.61} & \textbf{95.85} & \textbf{95.99} & \textbf{76.94} & \textbf{78.00} & \textbf{78.87} & \textbf{79.14} \\ \midrule
SDT w/ STF-1              & \multirow{4}{*}{SDT-2-512}        & 94.96 & 95.54 & 96.00 & 96.25 & 77.31 & 79.05 & \underline{79.97} & 79.93       \\
SDT w/ STF-2              &    & \underline{95.15} & \underline{95.77} & \underline{96.19} & \underline{96.34} & \underline{77.97} & \underline{79.32} & 79.87 & 80.35       \\
SDT w/ STF-3              &    & 94.80 & 95.68 & 96.12 & 96.19 & 77.58 & 79.30 & 79.76 & \underline{80.41}      \\
SDT w/ STF-4        &  & \textbf{95.24} & \textbf{95.86} & \textbf{96.23} & \textbf{96.41} & \textbf{78.21} & \textbf{79.44} & \textbf{80.34} & \textbf{80.61} \\ \midrule
QKFormer w/ STF-1         & \multirow{4}{*}{HST-4-384}        & \underline{95.81} & 96.15 & 96.23 & 96.40 & 79.69 & \underline{81.20} & 81.37 & 81.77       \\
QKFormer w/ STF-2         &    &  95.78 & \underline{96.24} & \underline{96.41} & \underline{96.56} & 79.97 & 81.11 & 81.40 & 81.79      \\
QKFormer w/ STF-3         &    & 95.68 & 96.17 & 96.40 & 96.52 & \underline{80.02} & 81.18 & \underline{81.45} & \underline{81.80}       \\
QKFormer w/ STF-4   &  & \textbf{96.02} & \textbf{96.33} & \textbf{96.51} & \textbf{96.61} & \textbf{80.07} & \textbf{81.26} & \textbf{81.51} & \textbf{81.89} \\
\bottomrule
\end{tabular}
\caption{Performance comparison of different STF variants across various Transformer-based SNN backbones and spike timestep settings. Bold indicates the best performance, and underlined denotes the second-best.}
\label{tab:cifar_datasets_stf_varient_detail_results}
\end{table*}
}

\section{Computational Environment}

\subsection{Software Setup} 
We employ PyTorch version 2.0.1, supported by CUDA 11.8, together with SpikingJelly v0.0.0.0.12 to implement and train our models.

\subsection{Hardware Setup}

To accommodate varying computational demands, we employ different types of GPUs depending on the dataset scale.

For experiments conducted on the smaller datasets, we use a single NVIDIA L40 GPU per run, each equipped with 42 GB of memory. All experiments are performed in a single-GPU setting to ensure consistency across configurations.

For the large-scale ImageNet experiments, we utilize eight NVIDIA H20 GPUs per run, with each GPU offering 96 GB of memory. This multi-GPU setup is adopted to accommodate the computational and memory demands of training on high-resolution data.

\section{Ablation Study on Variants of the STF Module}

Table~\ref{tab:cifar_datasets_stf_varient_detail_results} presents a detailed performance comparison of different STF variants on the CIFAR datasets, corresponding to the visualizations in Figure 3 of the main text. Experiments demonstrate that STF-4 consistently outperforms other STF variants under different timestep settings across various Transformer-based SNN backbones. Unless explicitly specified, STF-4 serves as the default configuration throughout this paper.

\section{Theoretical Energy Consumption of SNNs and ANNs}

As illustrated in Figure 1 of the main text, the temporal feedback in STF follows the LIF neuron and subsequently passes through a convolution and batch normalization (BN) layer, thereby enhancing spike-driven characteristics. Prior studies~\cite{deng2022temporal, hu2021spiking} have shown that, due to the regularity of convolution operations, BN and linear scaling can be fused into the convolutional layer as an effective bias term during deployment. Consequently, the influence of BN can be neglected in theoretical energy estimations.

Based on this observation, the theoretical energy consumption of STF integrated into various Transformer-based SNN backbones is computed as:

\begin{multline}
    E_{\text{total}} =  E_{\text{MAC}} \times \text{FLOPs}_{\text{Conv}}^{1} + \\ 
    E_{\text{AC}} \times \Bigg( \sum_{t=1}^{T} \text{SOP}_{\text{Conv}_{\text{STF}}}^{\prime \ t} 
  + \sum_{j=1}^{M} \text{SOP}_{\text{block}}^{j} \Bigg),
  \label{equ:energy_appendix}
\end{multline}

\noindent
where $E_{\mathrm{MAC}}$ and $E_{\mathrm{AC}}$ denote the energy consumption of multiply-and-accumulate (MAC) operations and accumulate (AC) operations, respectively. We assume these operations are executed on 45\,nm hardware, with $E_{\mathrm{MAC}} = 4.6\,\mathrm{pJ}$ and $E_{\mathrm{AC}} = 0.9\,\mathrm{pJ}$, as reported in previous studies \cite{horowitz20141, yao2023spike, zhou2022spikformer}. $\text{FLOPs}_{\text{Conv}}^{1}$ denotes the number of floating-point MAC operations in the first convolutional layer. $\text{SOP}_{\text{Conv}_{\text{STF}}}^{\prime \ t}$ represents the number of synaptic operations (SOPs) at time step $t$ under STF feedback, and $\text{SOP}_{\text{block}}^{j}$ indicates the total SOPs over all time steps $T$ in the $j$-th block of a given Transformer-based SNN backbone. The corresponding formulations for computing $\text{SOP}_{\text{Conv}_{\text{STF}}}^{\prime \ t}$ and $\text{SOP}_{\text{block}}^{j}$ are given as:

\begin{equation}
    \text{SOP}_{\text{Conv}_{\text{STF}}}^{\prime \ t} = f_r \times \text{FLOPs}^{\text{STF}},
\end{equation}

\begin{equation}
    \text{SOP}_{\text{block}}^{j} = f_r \times T \times \text{FLOPs}^{j},  
\end{equation}

\noindent
where $f_r$ denotes the firing rate of spike trains, and $\text{FLOPs}$ represents the number of floating-point operations.

\section{Inference Latency Statistics on ImageNet-1K}

We further analyze the impact of STF on inference latency over the ImageNet-1K dataset and compare it with SpiLiFormer \cite{zheng2025spiliformer} that adopts deep-level feedback. As shown in Table~\ref{tab:imagetnet_1k_latency_appendix}, the relative increase in inference latency caused by STF accounts for roughly half of the proportion introduced by deep-level feedback, which aligns with the observations discussed in Section 4.2 of the main text.

{
\setcounter{table}{3} 
\begin{table}[h]
\setlength{\tabcolsep}{1mm}
\centering
\begin{tabular}{lc}
\toprule
Methods                  & Inference Time per Sample (ms) \\ \midrule
SDT                          & 26.987                         \\
SDT w/ STF               & 28.375 (\textbf{+5.14\%})               \\ \midrule
QKFormer                  & 57.316                         \\
QKFormer w/ STF            & 60.942 (\textbf{+6.33\%})               \\ \midrule
SpiLiFormer w/o feedback   & 58.990                         \\
SpiLiFormer                 & 66.302 (\textbf{+12.40\%})          \\ \bottomrule
\end{tabular}
\caption{Comparison of inference latency overhead between the shallow-level STF module and traditional deep-level feedback on ImageNet-1K.}
\label{tab:imagetnet_1k_latency_appendix}
\end{table}
}

\section{$SG_t$ Derivation Under Static Datasets}

Building on prior work \cite{qiu2024gated}, we provide a step-by-step derivation of the spike generation time ($SG_t$) formulation under static image input without using STF. To begin with, the membrane dynamics of a standard LIF neuron \cite{maass1997networks} can be described by the following differential equation:

\begin{equation}
    \tau_m \frac{\mathrm{d} U[t]}{\mathrm{d} t} = -(U[t] - U_{\text{reset}}) + I[t],
\end{equation}
where $\tau_m$ denotes the membrane time constant, $U[t]$ represents the membrane potential at time step $t$, and $I[t]$ denotes the input current at time step $t$. Subsequently, by discretizing the differential equation using the Euler method and assuming $U_{\text{reset}} = 0$ for simplicity in practical implementations, we obtain:

\begin{equation}
    \tau_m \frac{U[t] - U[t-1]}{\Delta t} = -U[t-1] + I[t].
\end{equation}
After further rearrangement, we obtain the following:

\begin{equation}
    U[t] = U[t-1] (1 - \frac{\Delta t}{\tau_m}) +\frac{\Delta t}{\tau_m} I[t].
\end{equation}
We define $\tau = 1 - \frac{\Delta t}{\tau_m}$ and $\beta = \frac{\Delta t}{\tau_m}$, where $\tau$ is the leak decay factor. Following prior work \cite{zhou2022spikformer, yao2023spike, zhou2024qkformer}, we adopt the default setting $\tau_m = 2.0$, which leads to $\tau = \frac{1}{2}$. Moreover, in practice, $\beta$ can be further absorbed into the data preprocessing. Thus, we obtain:

\begin{equation}
    U[t] = \tau U[t-1] + I[t].
    \label{eq:SG_t_appendix_1}
\end{equation}
Suppose that a spike is emitted at time step $SG_t$, meaning no spikes are generated during the interval from $t = 1$ to $t < SG_t$. Based on Equation~\ref{eq:SG_t_appendix_1}, we can obtain:

\begin{equation}
    U[SG_t - 1] = \tau^{SG_t-1} U[0] + \sum_{t=1}^{SG_t-1}{\tau^{SG_t-1-t} \ I[t]}.
\end{equation}
Since $I[t]$ is replicated across $T$ time steps, we have $I[t] = I$ for $t = 1, 2, \dots, T$. The initial membrane potential is set to $U[0] = 0$ by default. Thus, we obtain:

\begin{equation}
    U[SG_t-1] = \sum_{t=1}^{SG_t-1}{\tau^{SG_t-1-t}I} < U_{\text{th}}.
    \label{eq:sg_t_appendix}
\end{equation}
Since a spike is triggered at time step $SG_t$, we can derive the following based on Equation \ref{eq:sg_t_appendix}:

\begin{equation}
    \sum_{t=1}^{SG_t-1}{\tau^{SG_t-1-t} \ I} < U_{\text{th}} \leq \sum_{t=1}^{SG_t}{\tau^{SG_t-t} \ I}.
\end{equation}
From Equation 10 and the geometric series properties, we can obtain:

\begin{equation}
    \frac{1 - \tau^{SG_t-1}}{1 - \tau} \ I < U_{\text{th}} \leq  \frac{1-\tau^{SG_t}}{1 - \tau} \ I,
\end{equation}

\begin{equation}
    \tau^{SG_t}  \leq 1 - \frac{U_{\text{th}}(1 - \tau)}{I} <  \tau^{SG_t-1},
\end{equation}

\begin{equation}
    SG_t - 1 < \log_{\tau} (1 - \frac{U_{\text{th}}(1 - \tau)}{I}) \leq SG_t,
\end{equation}

\begin{equation}
    SG_t = \left\lceil \log_{\tau} (1-\frac{U_{\text{th}}(1 - \tau)}{I})\right\rceil.
\end{equation}

{
\setcounter{table}{5}
\renewcommand{\arraystretch}{1.0}
\begin{table*}[!h]
	\centering
	\begin{tabular}{lccccccccc}
		\toprule
		\multirow{3}{*}{Methods}          & \multirow{3}{*}{Architecture}     & \multicolumn{8}{c}{Acc (\%) / \textit{Spike Entropy}}                                                           \\ \cmidrule{3-10}
		                                  &                                   & \multicolumn{4}{c}{CIFAR-10}          & \multicolumn{4}{c}{CIFAR-100}             \\
		                                  &                                   & T=2     & T=4     & T=6     & T=8     & T=2      & T=4      & T=6      & T=8      \\ \midrule
                                          
        \multirow{2}{*}{Spikformer} & \multirow{2}{*}{Spikformer-4-384} & 93.20   & 95.19   & 95.14   & 95.24   & 75.17    & 77.86     & 78.17    & 78.22    \\
		                                  &     & \textit{0.964} & \textit{1.442} & \textit{1.530} & \textit{1.664} & \textit{1.203} & \textit{1.590} & \textit{1.774} & \textit{1.860} \\ \midrule
        
        \multirow{2}{*}{Spikformer w/ TSPE} & \multirow{2}{*}{Spikformer-4-384} & 94.72 & 95.26 & 95.44 & 95.69 & 76.21 & 77.88 & 78.43 & 78.54 \\
                                            &  & \textit{1.286} & \textit{1.866} & \textit{2.277} & \textit{2.445} & \textit{1.263} & \textit{1.964} & \textit{2.335} & \textit{2.593} \\ \midrule

        \multirow{2}{*}{Spikformer w/ TF} & \multirow{2}{*}{Spikformer-4-384} & 94.85  & 95.52 & 95.62 & 95.89 & 76.40 & 77.96 & 78.75 & 78.72            \\
                                            &    & \textit{1.364} & \textit{2.427} & \textit{4.245} & \textit{4.784} & \textit{1.386} & \textit{2.566} & \textit{4.383} & \textit{5.021} \\ \midrule
                                            
		\multirow{2}{*}{Spikformer w/ STF} & \multirow{2}{*}{Spikformer-4-384} & \textbf{94.95} & \textbf{95.61} & \textbf{95.85} & \textbf{95.99} & \textbf{76.94} & \textbf{78.00} & \textbf{78.87} & \textbf{79.14} \\
		                                  &     & \textit{\textbf{1.463}} & \textit{\textbf{2.573}} & \textit{\textbf{4.369}} & \textit{\textbf{4.932}} & \textit{\textbf{1.479}} & \textit{\textbf{2.597}} & \textit{\textbf{4.417}} & \textit{\textbf{5.240}} \\ \bottomrule
	\end{tabular}
	\caption{Comparison of ablation studies on the TSPE and TF components.}
	\label{tab:TSPE_TF_ablation_appendix}
\end{table*}
}

\section{Definition of Spike Entropy}
\label{sec:spike_entropy_appendix}

Based on the previous work \cite{luczak2024entropy}, the definition of spike entropy is as follows:

\begin{equation}
    \text{Spike Entropy} = - \sum_{i=1}^{2^T} p_i \log_2p_i,
\end{equation}
where $p_i$ denotes the proportion of each distinct spike pattern. Assuming the spike timestep is $T$, the index $i$ ranges from $1$ to $2^T$, covering all possible patterns. Higher spike entropy reflects greater diversity in neural spike patterns, implying a higher potential information capacity.

\section{Adversarial Evaluation and Statistical Results}
In our adversarial evaluation, we adopt two commonly used methods for generating adversarial examples: the Fast Gradient Sign Method (FGSM) \cite{goodfellow2014explaining} and Projected Gradient Descent (PGD) \cite{madry2017towards}.

\textbf{FGSM} is a one-step perturbation method that modifies input samples based on the direction of the loss gradient. The adversarial input $\tilde{\mathbf{x}}$ is computed as:
\begin{equation}
\tilde{\mathbf{x}} = \mathbf{x} + \beta \cdot \mathrm{sign} \left( \nabla_{\mathbf{x}} J(\mathbf{x}, \mathbf{y}) \right).
\label{eq:fgsm_alt_appendix}
\end{equation}
Here, $\mathbf{x}$ denotes the clean input, $\mathbf{y}$ is the target label, $J$ represents the loss function, and $\beta$ controls the magnitude of the perturbation.

\textbf{PGD} generalizes FGSM by introducing iterative refinements. Starting from an initial perturbed sample $\tilde{\mathbf{x}}^{(0)}$, PGD updates the adversarial input over multiple steps as follows:
{\small
\begin{equation}
\tilde{\mathbf{x}}^{(t)} = \Pi_{\mathcal{B}\infty(\mathbf{x}, \epsilon)} \left( \tilde{\mathbf{x}}^{(t-1)} + \eta \cdot \mathrm{sign} \left( \nabla{\mathbf{x}} J(\tilde{\mathbf{x}}^{(t-1)}, \mathbf{y}) \right) \right),
\label{eq:pgd_alt_appendix}
\end{equation}
}
where $\eta$ is the step size, $\Pi_{\mathcal{B}\infty(\mathbf{x}, \epsilon)}(\cdot)$ denotes projection onto the $L\infty$ ball of radius $\epsilon$ centered at $\mathbf{x}$, ensuring the perturbation remains bounded.

We evaluate three Transformer-based SNN backbones under four different spike timestep settings ($T=2,4,6,8$) to demonstrate the better robustness of STF over direct coding (baseline models without STF).

As shown in Tables~7, 8, and 9, the detailed and comprehensive experimental results demonstrate that STF achieves better robustness without sacrificing clean performance, compared to direct coding.

\section{Ablation Study of TSPE and TF Components}

We conduct detailed ablation experiments on the TSPE and TF components using Spikformer as the backbone. The experiments are performed under four timestep settings on the CIFAR datasets to evaluate their impact on model performance and the diversity of spike patterns. Following the methodology described in Section~\ref{sec:spike_entropy_appendix}, we use spike entropy as the metric to quantify the diversity of spike patterns. As shown in Table~\ref{tab:TSPE_TF_ablation_appendix}, TF leads to more substantial improvements in both the diversity of spike patterns and model performance compared to TSPE, while their combination yields further gains.

\section{Performance Evaluation on CIFAR10-DVS}

We select two Transformer-based SNN backbones, including Spikformer and SDT, to evaluate the effect of STF on model performance under varying spike timestep settings using the neuromorphic dataset CIFAR10-DVS \cite{li2017cifar10}. As presented in Table~\ref{tab:cifar_10_dvs_appendix}, STF results in performance degradation, which is consistent with our expectations. A detailed analysis can be found in Section 4.4 of the main text.

{
\setcounter{table}{4}
\renewcommand{\arraystretch}{1.0}
\begin{table}[!h]
\centering
\begin{tabular}{cccc}
\toprule
Architecture                      & $T$  & Baseline & w/ STF   \\ \midrule
\multirow{4}{*}{Spikformer-4-384} & 10 & 77.3\%    & 71.4\%  \\
                                  & 16 & 79.3\%    & 75.1\%  \\
                                  & 24 & 80.4\%    & 80.2\%  \\
                                  & 32 & 81.9\%    & 79.6\%  \\ \midrule
\multirow{4}{*}{SDT-2-512}        & 10 & 74.3\%    & 64.7\%  \\
                                  & 16 & 76.1\%    & 56.3\%  \\
                                  & 24 & 77.6\%    & 66.4\%  \\
                                  & 32 & 79.0\%    & 73.0\%  \\ \bottomrule
\end{tabular}
\caption{Performance comparison on CIFAR10-DVS}
\label{tab:cifar_10_dvs_appendix}
\end{table}
}

\section{Detailed Statistics of Spike Pattern Diversity}

As shown in Tables~10-19, we compare the spike pattern statistics at the output of LIF neurons in the encoding layer of Spikformer on the CIFAR-10 dataset. The comparison covers direct coding, its variants (GAC coding and IMP coding), and our proposed STF under different time steps ($T=2,4,6,8$). Additional detailed statistics for other backbone architectures can be found in the submitted supplementary material.

\clearpage



\section*{Supplementary material}

| Dataset | Methods | Time Step | Clean | FGSM Maximum Perturbation | | | | PGD Iterations | | | |
|---|---|---|---|---|---|---|---|---|---|---|---|
| | | | | 0.05 | 0.1 | 0.2 | 0.3 | 5 | 10 | 30 | 50 |
| CIFAR-10 | Spikformer | 2 | 93.20 | 46.24 | 40.33 | 29.44 | 23.00 | 31.44 | 25.73 | 23.30 | 22.38 |
| | Spikformer w/ STF | 2 | **94.95** (+1.75) | **46.69** (+0.45) | **42.64** (+2.31) | **34.23** (+4.79) | **26.22** (+3.22) | **33.72** (+2.28) | **29.32** (+3.59) | **27.14** (+3.84) | **26.96** (+4.58) |
| | Spikformer | 4 | 95.19 | 54.92 | 51.72 | 41.20 | 30.98 | 27.56 | 21.08 | 18.33 | 17.68 |
| | Spikformer w/ STF | 4 | **95.61** (+0.42) | **58.37** (+3.45) | **55.75** (+4.03) | **45.40** (+4.2) | **35.89** (+4.91) | **30.95** (+3.39) | **23.81** (+2.73) | **20.54** (+2.21) | **20.05** (+2.37) |
| | Spikformer | 6 | 95.14 | 56.13 | 50.48 | 35.31 | 24.93 | 28.15 | 22.50 | 19.88 | 19.11 |
| | Spikformer w/ STF | 6 | **95.85** (+0.71) | **61.33** (+5.2) | **59.72** (+9.24) | **50.70** (+15.39) | **40.56** (+15.63) | **31.37** (+3.22) | **25.37** (+2.87) | **22.47** (+2.59) | **21.97** (+2.86) |
| | Spikformer | 8 | 95.24 | 53.13 | 47.21 | 35.53 | 26.24 | 28.61 | 24.53 | 22.29 | 22.04 |
| | Spikformer w/ STF | 8 | **95.99** (+0.75) | **60.59** (+7.46) | **58.88** (+11.67) | **50.19** (+14.66) | **39.58** (+13.34) | **32.81** (+4.20) | **26.49** (+1.96) | **23.44** (+1.15) | **22.56** (+0.52) |
| CIFAR-100 | Spikformer | 2 | 75.17 | 21.96 | 15.82 | 9.79 | 6.91 | 14.03 | 10.32 | 9.01 | 9.06 |
| | Spikformer w/ STF | 2 | **76.94** (+1.77) | **22.10** (+0.14) | **16.11** (+0.29) | **10.10** (+0.31) | **7.07** (+0.16) | **14.06** (+0.03) | **10.51** (+0.19) | **9.11** (+0.10) | **9.34** (+0.28) |
| | Spikformer | 4 | 77.86 | 26.05 | 21.86 | 15.19 | 10.58 | 15.03 | 11.52 | 9.79 | 9.58 |
| | Spikformer w/ STF | 4 | **78.00** (+0.14) | **26.41** (+0.36) | **22.00** (+0.14) | **15.75** (+0.56) | **11.73** (+1.15) | **15.12** (+0.09) | **12.36** (+0.84) | **10.48** (+0.69) | **10.50** (+0.92) |
| | Spikformer | 6 | 78.17 | 29.94 | 25.49 | 16.96 | 11.86 | 15.39 | 12.32 | 10.55 | **10.11** |
| | Spikformer w/ STF | 6 | **78.87** (+0.70) | **30.68** (+0.74) | **27.18** (+1.69) | **20.15** (+3.19) | **14.59** (+2.73) | **15.93** (+0.54) | **12.37** (+0.05) | **10.64** (+0.09) | 10.03 (-0.08) |
| | Spikformer | 8 | 78.22 | 31.18 | 26.34 | 18.22 | 12.77 | 15.74 | 12.55 | 10.97 | 10.67 |
| | Spikformer w/ STF | 8 | **79.14** (+0.92) | **31.47** (+0.29) | **28.14** (+1.80) | **21.17** (+2.95) | **15.90** (+3.13) | **17.22** (+1.48) | **13.54** (+0.99) | **11.79** (+0.82) | **11.58** (+0.91) |

Table 7: Adversarial robustness comparison between Spikformer with and without STF on the CIFAR datasets.

| Dataset | Methods | Time Step | Clean | FGSM Maximum Perturbation | | | | PGD Iterations | | | |
|---|---|---|---|---|---|---|---|---|---|---|---|
| | | | | 0.05 | 0.1 | 0.2 | 0.3 | 5 | 10 | 30 | 50 |
| CIFAR-10 | SDT | 2 | 95.01 | 52.65 | 42.74 | 33.97 | 28.23 | 35.77 | 27.6 | 23.17 | 23.99 |
| | SDT w/ STF | 2 | **95.24** (+0.23) | **52.89** (+0.24) | **43.29** (+0.55) | **34.22** (+0.25) | **28.95** (+0.72) | **36.12** (+0.35) | **28.32** (+0.72) | **24.74** (+1.57) | **24.92** (+0.93) |
| | SDT | 4 | 95.60 | 62.11 | 56.18 | 50.54 | 46.46 | 38.59 | 30.81 | 26.1 | **25.62** |
| | SDT w/ STF | 4 | **95.86** (+0.26) | **63.91** (+1.80) | **59.57** (+3.39) | **53.78** (+3.24) | **50.00** (+3.54) | **39.01** (+0.42) | **30.85** (+0.04) | **26.48** (+0.38) | 25.53 (-0.09) |
| | SDT | 6 | 96.11 | 69.31 | 66.37 | 61.13 | 57.01 | 42.29 | 34.81 | 30.45 | 28.77 |
| | SDT w/ STF | 6 | **96.23** (+0.12) | **70.22** (+0.91) | **67.74** (+1.37) | **63.51** (+2.38) | **59.74** (+2.73) | **43.03** (+0.74) | **35.1** (+0.29) | **30.57** (+0.12) | **29.6** (+0.83) |
| | SDT | 8 | 96.24 | 70.26 | 67.94 | 63.04 | 58.28 | 42.85 | 35.42 | 30.63 | 30.16 |
| | SDT w/ STF | 8 | **96.41** (+0.17) | **72.65** (+2.39) | **69.06** (+1.12) | **63.89** (+0.85) | **59.22** (+0.94) | **43.99** (+1.14) | **36.37** (+0.95) | **31.59** (+0.96) | **30.70** (+0.54) |
| CIFAR-100 | SDT | 2 | 77.47 | 32.29 | 26.66 | 21.63 | 18.30 | 20.73 | 15.88 | 14.82 | 14.4 |
| | SDT w/ STF | 2 | **78.21** (+0.74) | **33.23** (+0.94) | **27.86** (+1.20) | **22.91** (+1.28) | **20.02** (+1.72) | **21.02** (+0.29) | **16.81** (+0.93) | **15.17** (+0.35) | **14.83** (+0.43) |
| | SDT | 4 | 78.40 | 37.23 | **32.97** | 27.69 | 25.02 | 23.16 | 17.91 | 16.84 | 16.25 |
| | SDT w/ STF | 4 | **79.44** (+1.04) | **37.46** (+0.23) | 32.60 (-0.37) | **28.52** (+0.83) | **25.08** (+0.06) | **23.43** (+0.27) | **18.76** (+0.85) | **17.30** (+0.46) | **16.48** (+0.23) |
| | SDT | 6 | 79.43 | 38.01 | 34.75 | 30.13 | 26.88 | 22.15 | 17.71 | 15.64 | 15.49 |
| | SDT w/ STF | 6 | **80.34** (+0.91) | **39.39** (+1.38) | **35.53** (+0.78) | **31.67** (+1.54) | **28.24** (+1.36) | **22.80** (+0.65) | **18.18** (+0.47) | **16.35** (+0.71) | **16.10** (+0.61) |
| | SDT | 8 | 79.87 | 40.22 | 36.49 | 31.79 | 28.49 | 22.31 | 18.07 | 14.81 | 14.71 |
| | SDT w/ STF | 8 | **80.61** (+0.74) | **40.33** (+0.11) | **36.53** (+0.04) | **32.42** (+0.63) | **29.38** (+0.89) | **22.54** (+0.23) | **18.41** (+0.34) | **15.90** (+1.09) | **15.57** (+0.86) |

Table 8: Adversarial robustness comparison between SDT with and without STF on the CIFAR datasets.

| Dataset | Methods | Time Step | Clean | FGSM Maximum Perturbation | | | | PGD Iterations | | | |
|---|---|---|---|---|---|---|---|---|---|---|---|
| | | | | 0.05 | 0.1 | 0.2 | 0.3 | 5 | 10 | 30 | 50 |
| CIFAR-10 | QKFormer | 2 | 95.79 | 59.08 | 51.57 | 43.19 | 37.19 | 31.37 | 22.11 | 16.50 | 15.45 |
| | QKFormer w/ STF | 2 | **96.02** (+0.23) | **59.52** (+0.44) | **51.85** (+0.28) | **43.36** (+0.17) | **37.31** (+0.12) | **31.81** (+0.44) | **22.23** (+0.12) | **17.09** (+0.59) | **15.86** (+0.41) |
| | QKFormer | 4 | 96.18 | 68.56 | 64.50 | 58.64 | 51.77 | 33.11 | 24.59 | 17.80 | 16.63 |
| | QKFormer w/ STF | 4 | **96.33** (+0.15) | **71.15** (+2.59) | **67.61** (+3.11) | **61.16** (+2.52) | **54.26** (+2.49) | **38.33** (+5.22) | **28.21** (+3.62) | **21.07** (+3.27) | **19.31** (+2.68) |
| | QKFormer | 6 | 96.37 | 66.07 | 61.30 | 53.19 | 45.80 | 34.21 | 24.71 | 18.28 | 17.09 |
| | QKFormer w/ STF | 6 | **96.51** (+0.14) | **66.88** (+0.81) | **64.06** (+2.76) | **58.61** (+5.42) | **52.83** (+7.03) | **34.43** (+0.22) | **25.08** (+0.37) | **18.47** (+0.19) | **17.13** (+0.04) |
| | QKFormer | 8 | 96.35 | 68.51 | 65.68 | 60.12 | 54.37 | 35.43 | 26.20 | 20.30 | **18.93** |
| | QKFormer w/ STF | 8 | **96.61** (+0.26) | **68.79** (+0.28) | **66.97** (+1.29) | **61.61** (+1.49) | **54.69** (+0.32) | **36.59** (+1.16) | **27.49** (+1.29) | **20.32** (+0.02) | 18.86 (-0.07) |
| CIFAR-100 | QKFormer | 2 | 79.79 | 32.63 | 28.05 | 22.98 | 18.95 | 17.25 | 12.52 | 10.15 | 9.86 |
| | QKFormer w/ STF | 2 | **80.07** (+0.28) | **36.21** (+3.58) | **31.44** (+3.39) | **25.42** (+2.44) | **21.04** (+2.09) | **20.44** (+3.19) | **15.29** (+2.77) | **12.62** (+2.47) | **12.09** (+2.23) |
| | QKFormer | 4 | 81.15 | 35.56 | 31.37 | 25.67 | 21.63 | 18.19 | 13.61 | 9.89 | 10.04 |
| | QKFormer w/ STF | 4 | **81.26** (+0.11) | **37.19** (+1.63) | **32.94** (+1.57) | **27.27** (+1.60) | **23.04** (+1.41) | **19.32** (+1.13) | **14.45** (+0.84) | **11.26** (+1.37) | **10.75** (+0.71) |
| | QKFormer | 6 | 81.35 | 36.90 | 33.39 | 28.43 | 23.97 | 18.40 | 13.32 | 9.89 | 9.10 |
| | QKFormer w/ STF | 6 | **81.51** (+0.16) | **38.33** (+1.43) | **35.27** (+1.88) | **30.01** (+1.58) | **25.42** (+1.45) | **20.49** (+2.09) | **15.41** (+2.09) | **11.84** (+1.95) | **11.13** (+2.03) |
| | QKFormer | 8 | 81.64 | 37.97 | 34.69 | 28.44 | 23.70 | 35.44 | 26.20 | 20.29 | **18.93** |
| | QKFormer w/ STF | 8 | **81.89** (+0.25) | **39.96** (+1.99) | **36.84** (+2.15) | **30.52** (+2.08) | **25.95** (+2.25) | **36.59** (+1.15) | **27.50** (+1.30) | **20.30** (+0.01) | 18.91 (-0.02) |

Table 9: Adversarial robustness comparison between QKFormer with and without STF on the CIFAR datasets.

| Spike Pattern | Count | Ratio |
|---|---|---|
| [0, 0] | 386,181,630 | 0.78569 |
| [0, 1] | 54,595,182 | 0.11107 |
| [1, 0] | 0 | 0 |
| [1, 1] | 50,743,188 | 0.10324 |
| Spike Entropy | 0.9637 | |
| Firing Rate | 0.1588 | |

(a) Direct Coding

| Spike Pattern | Count | Ratio |
|---|---|---|
| [0, 0] | 385,766,667 | 0.78484 |
| [0, 1] | 54,736,027 | 0.11136 |
| [1, 0] | 0 | 0 |
| [1, 1] | 51,017,306 | 0.10379 |
| Spike Entropy | 0.9662 | |
| Firing Rate | 0.1595 | |

(b) GAC Coding

| Spike Pattern | Count | Ratio |
|---|---|---|
| [0, 0] | 332,652,925 | 0.67678 |
| [0, 1] | 44,688,373 | 0.09092 |
| [1, 0] | 42,131,763 | 0.08572 |
| [1, 1] | 72,046,939 | 0.14658 |
| Spike Entropy | 1.4056 | |
| Firing Rate | 0.2349 | |

(c) IMP Coding

| Spike Pattern | Count | Ratio |
|---|---|---|
| [0, 0] | 318,063,858 | 0.64710 |
| [0, 1] | 70,499,779 | 0.14343 |
| [1, 0] | 29,936,350 | 0.06091 |
| [1, 1] | 73,020,013 | 0.14856 |
| Spike Entropy | 1.4627 | |
| Firing Rate | 0.2507 | |

(d) STF (Ours)

Table 10: Distribution of the spike patterns in the encoding layer of Spikformer with different schemes on CIFAR-10 at $T = 2$.

| Spike Pattern | Count | Ratio |
|---|---|---|
| [0, 0, 0, 0] | 342,242,368 | 0.69629 |
| [0, 0, 0, 1] | 13,510,753 | 0.02749 |
| [0, 0, 1, 0] | 28,708,328 | 0.05841 |
| [0, 0, 1, 1] | 0 | 0 |
| [0, 1, 0, 0] | 0 | 0 |
| [0, 1, 0, 1] | 56,695,620 | 0.11535 |
| [0, 1, 1, 0] | 0 | 0 |
| [0, 1, 1, 1] | 0 | 0 |
| [1, 0, 0, 0] | 0 | 0 |
| [1, 0, 0, 1] | 0 | 0 |
| [1, 0, 1, 0] | 0 | 0 |
| [1, 0, 1, 1] | 0 | 0 |
| [1, 1, 0, 0] | 0 | 0 |
| [1, 1, 0, 1] | 0 | 0 |
| [1, 1, 1, 0] | 0 | 0 |
| [1, 1, 1, 1] | 50,362,931 | 0.10246 |
| Spike Entropy | 1.4417 | |
| Firing Rate | 0.1816 | |

(a) Direct Coding

| Spike Pattern | Count | Ratio |
|---|---|---|
| [0, 0, 0, 0] | 349,113,731 | 0.71027 |
| [0, 0, 0, 1] | 12,453,653 | 0.02534 |
| [0, 0, 1, 0] | 26,487,893 | 0.05389 |
| [0, 0, 1, 1] | 0 | 0 |
| [0, 1, 0, 0] | 0 | 0 |
| [0, 1, 0, 1] | 53,701,975 | 0.10926 |
| [0, 1, 1, 0] | 0 | 0 |
| [0, 1, 1, 1] | 0 | 0 |
| [1, 0, 0, 0] | 0 | 0 |
| [1, 0, 0, 1] | 0 | 0 |
| [1, 0, 1, 0] | 0 | 0 |
| [1, 0, 1, 1] | 0 | 0 |
| [1, 1, 0, 0] | 0 | 0 |
| [1, 1, 0, 1] | 0 | 0 |
| [1, 1, 1, 0] | 0 | 0 |
| [1, 1, 1, 1] | 49,762,748 | 0.10124 |
| Spike Entropy | 1.3955 | |
| Firing Rate | 0.1757 | |

(b) GAC Coding

| Spike Pattern | Count | Ratio |
|---|---|---|
| [0, 0, 0, 0] | 331,272,166 | 0.67397 |
| [0, 0, 0, 1] | 7,127,049 | 0.01450 |
| [0, 0, 1, 0] | 16,387,942 | 0.03334 |
| [0, 0, 1, 1] | 0 | 0 |
| [0, 1, 0, 0] | 24,011,229 | 0.04885 |
| [0, 1, 0, 1] | 18,759,892 | 0.03817 |
| [0, 1, 1, 0] | 0 | 0 |
| [0, 1, 1, 1] | 0 | 0 |
| [1, 0, 0, 0] | 0 | 0 |
| [1, 0, 0, 1] | 0 | 0 |
| [1, 0, 1, 0] | 37,373,113 | 0.07604 |
| [1, 0, 1, 1] | 0 | 0 |
| [1, 1, 0, 0] | 0 | 0 |
| [1, 1, 0, 1] | 0 | 0 |
| [1, 1, 1, 0] | 0 | 0 |
| [1, 1, 1, 1] | 56,588,609 | 0.11513 |
| Spike Entropy | 1.6701 | |
| Firing Rate | 0.1964 | |

(c) IMP Coding

| Spike Pattern | Count | Ratio |
|---|---|---|
| [0, 0, 0, 0] | 265,198,402 | 0.53955 |
| [0, 0, 0, 1] | 19,567,140 | 0.03981 |
| [0, 0, 1, 0] | 33,923,411 | 0.06902 |
| [0, 0, 1, 1] | 7,457,890 | 0.01517 |
| [0, 1, 0, 0] | 28,667,675 | 0.05833 |
| [0, 1, 0, 1] | 24,183,612 | 0.04920 |
| [0, 1, 1, 0] | 11,907,156 | 0.02423 |
| [0, 1, 1, 1] | 15,956,240 | 0.03246 |
| [1, 0, 0, 0] | 8,005,800 | 0.01629 |
| [1, 0, 0, 1] | 2,780,431 | 0.00566 |
| [1, 0, 1, 0] | 7,693,307 | 0.01565 |
| [1, 0, 1, 1] | 3,815,101 | 0.00776 |
| [1, 1, 0, 0] | 3,338,439 | 0.00679 |
| [1, 1, 0, 1] | 5,733,335 | 0.01166 |
| [1, 1, 1, 0] | 4,863,576 | 0.00989 |
| [1, 1, 1, 1] | 48,428,485 | 0.09853 |
| Spike Entropy | 2.5732 | |
| Firing Rate | 0.2491 | |

(d) STF (Ours)

Table 11: Distribution of the spike patterns in the encoding layer of Spikformer with different schemes on CIFAR-10 at $T = 4$.

| Spike Pattern | Count | Ratio | Spike Pattern | Count | Ratio |
| --- | --- | --- | --- | --- | --- |
| [0, 0, 0, 0, 0, 0] | 342,373,731 | 0.69656 | [1, 0, 0, 0, 0, 0] | 0 | 0 |
| [0, 0, 0, 0, 0, 1] | 3,111,285 | 0.00633 | [1, 0, 0, 0, 0, 1] | 0 | 0 |
| [0, 0, 0, 0, 1, 0] | 6,282,613 | 0.01278 | [1, 0, 0, 0, 1, 0] | 0 | 0 |
| [0, 0, 0, 1, 0, 0] | 13,225,515 | 0.02691 | [1, 0, 0, 1, 0, 0] | 0 | 0 |
| [0, 0, 0, 1, 0, 1] | 0 | 0 | [1, 0, 0, 1, 0, 1] | 0 | 0 |
| [0, 0, 0, 0, 1, 1] | 0 | 0 | [1, 0, 0, 0, 1, 1] | 0 | 0 |
| [0, 0, 1, 1, 0] | 0 | 0 | [1, 0, 0, 1, 1, 0] | 0 | 0 |
| [0, 0, 0, 1, 1, 1] | 0 | 0 | [1, 0, 0, 1, 1, 1] | 0 | 0 |
| [0, 0, 1, 0, 0, 0] | 0 | 0 | [1, 0, 1, 0, 0, 0] | 0 | 0 |
| [0, 0, 1, 0, 0, 1] | 27,506,090 | 0.05596 | [1, 0, 1, 0, 0, 1] | 0 | 0 |
| [0, 0, 1, 0, 1, 0] | 0 | 0 | [1, 0, 1, 0, 1, 0] | 0 | 0 |
| [0, 0, 1, 0, 1, 1] | 0 | 0 | [1, 0, 1, 0, 1, 1] | 0 | 0 |
| [0, 0, 1, 1, 0, 0] | 0 | 0 | [1, 0, 1, 1, 0, 0] | 0 | 0 |
| [0, 0, 1, 1, 0, 1] | 0 | 0 | [1, 0, 1, 1, 0, 1] | 0 | 0 |
| [0, 0, 1, 1, 1, 0] | 0 | 0 | [1, 0, 1, 1, 1, 0] | 0 | 0 |
| [0, 0, 1, 1, 1, 1] | 0 | 0 | [1, 0, 1, 1, 1, 1] | 0 | 0 |
| [0, 1, 0, 0, 0, 0] | 0 | 0 | [1, 1, 0, 0, 0, 0] | 0 | 0 |
| [0, 1, 0, 0, 0, 1] | 0 | 0 | [1, 1, 0, 0, 0, 1] | 0 | 0 |
| [0, 1, 0, 0, 1, 0] | 0 | 0 | [1, 1, 0, 0, 1, 0] | 0 | 0 |
| [0, 1, 0, 0, 1, 1] | 0 | 0 | [1, 1, 0, 0, 1, 1] | 0 | 0 |
| [0, 1, 0, 1, 0, 0] | 0 | 0 | [1, 1, 0, 1, 0, 0] | 0 | 0 |
| [0, 1, 0, 1, 0, 1] | 52,721,780 | 0.10726 | [1, 1, 0, 1, 0, 1] | 0 | 0 |
| [0, 1, 0, 1, 1, 0] | 0 | 0 | [1, 1, 0, 1, 1, 0] | 0 | 0 |
| [0, 1, 0, 1, 1, 1] | 0 | 0 | [1, 1, 0, 1, 1, 1] | 0 | 0 |
| [0, 1, 1, 0, 0, 0] | 0 | 0 | [1, 1, 1, 0, 0, 0] | 0 | 0 |
| [0, 1, 1, 0, 0, 1] | 0 | 0 | [1, 1, 1, 0, 0, 1] | 0 | 0 |
| [0, 1, 1, 0, 1, 0] | 0 | 0 | [1, 1, 1, 0, 1, 0] | 0 | 0 |
| [0, 1, 1, 0, 1, 1] | 0 | 0 | [1, 1, 1, 0, 1, 1] | 0 | 0 |
| [0, 1, 1, 1, 0, 0] | 0 | 0 | [1, 1, 1, 1, 0, 0] | 0 | 0 |
| [0, 1, 1, 1, 0, 1] | 0 | 0 | [1, 1, 1, 1, 0, 1] | 0 | 0 |
| [0, 1, 1, 1, 1, 0] | 0 | 0 | [1, 1, 1, 1, 1, 0] | 0 | 0 |
| [0, 1, 1, 1, 1, 1] | 0 | 0 | [1, 1, 1, 1, 1, 1] | 46,298,986 | 0.0942 |
| Spike Entropy | | | 1.5296 | | |
| Firing Rate | | | 0.1742 | | |

Table 12: Distribution of the spike patterns in the encoding layer of Spikformer with direct coding on CIFAR-10 at $T = 6$.

| Spike Pattern | Count | Ratio | Spike Pattern | Count | Ratio |
|---|---|---|---|---|---|
| $[0, 0, 0, 0, 0, 0]$ | 343,464,329 | 0.69878 | $[1, 0, 0, 0, 0, 0]$ | 0 | 0 |
| $[0, 0, 0, 0, 0, 1]$ | 2,946,366 | 0.00599 | $[1, 0, 0, 0, 0, 1]$ | 0 | 0 |
| $[0, 0, 0, 0, 1, 0]$ | 6,005,346 | 0.01222 | $[1, 0, 0, 0, 1, 0]$ | 0 | 0 |
| $[0, 0, 0, 0, 1, 1]$ | 0 | 0 | $[1, 0, 0, 0, 1, 1]$ | 0 | 0 |
| $[0, 0, 0, 1, 0, 0]$ | 12,401,887 | 0.02523 | $[1, 0, 0, 1, 0, 0]$ | 0 | 0 |
| $[0, 0, 0, 1, 0, 1]$ | 0 | 0 | $[1, 0, 0, 1, 0, 1]$ | 0 | 0 |
| $[0, 0, 0, 1, 1, 0]$ | 0 | 0 | $[1, 0, 0, 1, 1, 0]$ | 0 | 0 |
| $[0, 0, 0, 1, 1, 1]$ | 0 | 0 | $[1, 0, 0, 1, 1, 1]$ | 0 | 0 |
| $[0, 0, 1, 0, 0, 0]$ | 0 | 0 | $[1, 0, 1, 0, 0, 0]$ | 0 | 0 |
| $[0, 0, 1, 0, 0, 1]$ | 26,284,530 | 0.05348 | $[1, 0, 1, 0, 0, 1]$ | 0 | 0 |
| $[0, 0, 1, 0, 1, 0]$ | 0 | 0 | $[1, 0, 1, 0, 1, 0]$ | 0 | 0 |
| $[0, 0, 1, 0, 1, 1]$ | 0 | 0 | $[1, 0, 1, 0, 1, 1]$ | 0 | 0 |
| $[0, 0, 1, 1, 0, 0]$ | 0 | 0 | $[1, 0, 1, 1, 0, 0]$ | 0 | 0 |
| $[0, 0, 1, 1, 0, 1]$ | 0 | 0 | $[1, 0, 1, 1, 0, 1]$ | 0 | 0 |
| $[0, 0, 1, 1, 1, 0]$ | 0 | 0 | $[1, 0, 1, 1, 1, 0]$ | 0 | 0 |
| $[0, 0, 1, 1, 1, 1]$ | 0 | 0 | $[1, 0, 1, 1, 1, 1]$ | 0 | 0 |
| $[0, 1, 0, 0, 0, 0]$ | 0 | 0 | $[1, 1, 0, 0, 0, 0]$ | 0 | 0 |
| $[0, 1, 0, 0, 0, 1]$ | 0 | 0 | $[1, 1, 0, 0, 0, 1]$ | 0 | 0 |
| $[0, 1, 0, 0, 1, 0]$ | 0 | 0 | $[1, 1, 0, 0, 1, 0]$ | 0 | 0 |
| $[0, 1, 0, 0, 1, 1]$ | 0 | 0 | $[1, 1, 0, 0, 1, 1]$ | 0 | 0 |
| $[0, 1, 0, 1, 0, 0]$ | 0 | 0 | $[1, 1, 0, 1, 0, 0]$ | 0 | 0 |
| $[0, 1, 0, 1, 0, 1]$ | 53,059,139 | 0.10795 | $[1, 1, 0, 1, 0, 1]$ | 0 | 0 |
| $[0, 1, 0, 1, 1, 0]$ | 0 | 0 | $[1, 1, 0, 1, 1, 0]$ | 0 | 0 |
| $[0, 1, 0, 1, 1, 1]$ | 0 | 0 | $[1, 1, 0, 1, 1, 1]$ | 0 | 0 |
| $[0, 1, 1, 0, 0, 0]$ | 0 | 0 | $[1, 1, 1, 0, 0, 0]$ | 0 | 0 |
| $[0, 1, 1, 0, 0, 1]$ | 0 | 0 | $[1, 1, 1, 0, 0, 1]$ | 0 | 0 |
| $[0, 1, 1, 0, 1, 0]$ | 0 | 0 | $[1, 1, 1, 0, 1, 0]$ | 0 | 0 |
| $[0, 1, 1, 0, 1, 1]$ | 0 | 0 | $[1, 1, 1, 0, 1, 1]$ | 0 | 0 |
| $[0, 1, 1, 1, 0, 0]$ | 0 | 0 | $[1, 1, 1, 1, 0, 0]$ | 0 | 0 |
| $[0, 1, 1, 1, 0, 1]$ | 0 | 0 | $[1, 1, 1, 1, 0, 1]$ | 0 | 0 |
| $[0, 1, 1, 1, 1, 0]$ | 0 | 0 | $[1, 1, 1, 1, 1, 0]$ | 0 | 0 |
| $[0, 1, 1, 1, 1, 1]$ | 0 | 0 | $[1, 1, 1, 1, 1, 1]$ | 47,358,403 | 0.09635 |
| Spike Entropy | | | 1.5150 | | |
| Firing Rate | | | 0.1754 | | |

Table 13: Distribution of the spike patterns in the encoding layer of Spikformer with GAC coding on CIFAR-10 at $T = 6$.

| Spike Pattern | Count | Ratio | Spike Pattern | Count | Ratio |
|---|---|---|---|---|---|
| [0, 0, 0, 0, 0, 0] | 338,159,121 | 0.68799 | [1, 0, 0, 0, 0, 0] | 0 | 0 |
| [0, 0, 0, 0, 0, 1] | 1,639,282 | 0.00333 | [1, 0, 0, 0, 0, 1] | 0 | 0 |
| [0, 0, 0, 0, 1, 0] | 3,392,152 | 0.00690 | [1, 0, 0, 0, 1, 0] | 0 | 0 |
| [0, 0, 0, 0, 1, 1] | 0 | 0 | [1, 0, 0, 0, 1, 1] | 0 | 0 |
| [0, 0, 0, 1, 0, 0] | 7,176,057 | 0.01460 | [1, 0, 0, 1, 0, 0] | 0 | 0 |
| [0, 0, 0, 1, 0, 1] | 0 | 0 | [1, 0, 0, 1, 0, 1] | 0 | 0 |
| [0, 0, 0, 1, 1, 0] | 0 | 0 | [1, 0, 0, 1, 1, 0] | 0 | 0 |
| [0, 0, 0, 1, 1, 1] | 0 | 0 | [1, 0, 0, 1, 1, 1] | 0 | 0 |
| [0, 0, 1, 0, 0, 0] | 12,294,224 | 0.02501 | [1, 0, 1, 0, 0, 0] | 0 | 0 |
| [0, 0, 1, 0, 0, 1] | 4,026,400 | 0.00819 | [1, 0, 1, 0, 0, 1] | 0 | 0 |
| [0, 0, 1, 0, 1, 0] | 0 | 0 | [1, 0, 1, 0, 1, 0] | 34,806,496 | 0.07081 |
| [0, 0, 1, 0, 1, 1] | 0 | 0 | [1, 0, 1, 0, 1, 1] | 0 | 0 |
| [0, 0, 1, 1, 0, 0] | 0 | 0 | [1, 0, 1, 1, 0, 0] | 0 | 0 |
| [0, 0, 1, 1, 0, 1] | 0 | 0 | [1, 0, 1, 1, 0, 1] | 0 | 0 |
| [0, 0, 1, 1, 1, 0] | 0 | 0 | [1, 0, 1, 1, 1, 0] | 0 | 0 |
| [0, 0, 1, 1, 1, 1] | 0 | 0 | [1, 0, 1, 1, 1, 1] | 0 | 0 |
| [0, 1, 0, 0, 0, 0] | 0 | 0 | [1, 1, 0, 0, 0, 0] | 0 | 0 |
| [0, 1, 0, 0, 0, 1] | 105,634 | 0.00021 | [1, 1, 0, 0, 0, 1] | 0 | 0 |
| [0, 1, 0, 0, 1, 0] | 23,232,901 | 0.04727 | [1, 1, 0, 0, 1, 0] | 0 | 0 |
| [0, 1, 0, 0, 1, 1] | 0 | 0 | [1, 1, 0, 0, 1, 1] | 0 | 0 |
| [0, 1, 0, 1, 0, 0] | 0 | 0 | [1, 1, 0, 1, 0, 0] | 0 | 0 |
| [0, 1, 0, 1, 0, 1] | 17,930,876 | 0.03648 | [1, 1, 0, 1, 0, 1] | 0 | 0 |
| [0, 1, 0, 1, 1, 0] | 0 | 0 | [1, 1, 0, 1, 1, 0] | 0 | 0 |
| [0, 1, 0, 1, 1, 1] | 0 | 0 | [1, 1, 0, 1, 1, 1] | 0 | 0 |
| [0, 1, 1, 0, 0, 0] | 0 | 0 | [1, 1, 1, 0, 0, 0] | 0 | 0 |
| [0, 1, 1, 0, 0, 1] | 0 | 0 | [1, 1, 1, 0, 0, 1] | 0 | 0 |
| [0, 1, 1, 0, 1, 0] | 0 | 0 | [1, 1, 1, 0, 1, 0] | 0 | 0 |
| [0, 1, 1, 0, 1, 1] | 0 | 0 | [1, 1, 1, 0, 1, 1] | 0 | 0 |
| [0, 1, 1, 1, 0, 0] | 0 | 0 | [1, 1, 1, 1, 0, 0] | 0 | 0 |
| [0, 1, 1, 1, 0, 1] | 0 | 0 | [1, 1, 1, 1, 0, 1] | 0 | 0 |
| [0, 1, 1, 1, 1, 0] | 0 | 0 | [1, 1, 1, 1, 1, 0] | 0 | 0 |
| [0, 1, 1, 1, 1, 1] | 0 | 0 | [1, 1, 1, 1, 1, 1] | 48,756,857 | 0.09920 |
| Spike Entropy | | | 1.7133 | | |
| Firing Rate | | | 0.1797 | | |

Table 14: Distribution of the spike patterns in the encoding layer of Spikformer with IMP coding on CIFAR-10 at $T = 6$.

| Spike Pattern | Count | Ratio | Spike Pattern | Count | Ratio |
| --- | --- | --- | --- | --- | --- |
| [0, 0, 0, 0, 0, 0] | 153,390,794 | 0.31207 | [1, 0, 0, 0, 0, 0] | 13,437,249 | 0.02734 |
| [0, 0, 0, 0, 0, 1] | 17,672,955 | 0.03596 | [1, 0, 0, 0, 0, 1] | 1,539,807 | 0.00313 |
| [0, 0, 0, 0, 1, 0] | 21,481,933 | 0.04371 | [1, 0, 0, 0, 1, 0] | 1,668,172 | 0.00339 |
| [0, 0, 0, 0, 1, 1] | 4,090,165 | 0.00832 | [1, 0, 0, 0, 1, 1] | 367,067 | 0.00075 |
| [0, 0, 0, 1, 0, 0] | 25,066,475 | 0.05100 | [1, 0, 0, 1, 0, 0] | 21,684,054 | 0.04412 |
| [0, 0, 0, 1, 0, 1] | 4,208,713 | 0.00856 | [1, 0, 0, 1, 0, 1] | 3,148,724 | 0.00641 |
| [0, 0, 0, 1, 1, 0] | 5,314,403 | 0.01081 | [1, 0, 0, 1, 1, 0] | 3,829,962 | 0.00779 |
| [0, 0, 0, 1, 1, 1] | 1,395,765 | 0.00284 | [1, 0, 0, 1, 1, 1] | 879,695 | 0.00179 |
| [0, 0, 1, 0, 0, 0] | 24,662,353 | 0.05018 | [1, 0, 1, 0, 0, 0] | 2,519,839 | 0.00513 |
| [0, 0, 1, 0, 0, 1] | 28,702,361 | 0.05840 | [1, 0, 1, 0, 0, 1] | 2,557,085 | 0.00520 |
| [0, 0, 1, 0, 1, 0] | 3,846,168 | 0.00783 | [1, 0, 1, 0, 1, 0] | 388,368 | 0.00079 |
| [0, 0, 1, 0, 1, 1] | 5,584,658 | 0.01136 | [1, 0, 1, 0, 1, 1] | 502,201 | 0.00102 |
| [0, 0, 1, 1, 0, 0] | 4,763,066 | 0.00969 | [1, 0, 1, 1, 0, 0] | 4,483,259 | 0.00912 |
| [0, 0, 1, 1, 0, 1] | 6,549,139 | 0.01332 | [1, 0, 1, 1, 0, 1] | 7,232,276 | 0.01471 |
| [0, 0, 1, 1, 1, 0] | 1,180,092 | 0.00240 | [1, 0, 1, 1, 1, 0] | 919,458 | 0.00187 |
| [0, 0, 1, 1, 1, 1] | 1,815,570 | 0.00369 | [1, 0, 1, 1, 1, 1] | 1,597,980 | 0.00325 |
| [0, 1, 0, 0, 0, 0] | 18,269,651 | 0.03717 | [1, 1, 0, 0, 0, 0] | 2,522,659 | 0.00513 |
| [0, 1, 0, 0, 0, 1] | 2,068,725 | 0.00421 | [1, 1, 0, 0, 0, 1] | 314,479 | 0.00064 |
| [0, 1, 0, 0, 1, 0] | 24,728,089 | 0.05031 | [1, 1, 0, 0, 1, 0] | 2,570,239 | 0.00523 |
| [0, 1, 0, 0, 1, 1] | 3,734,288 | 0.00760 | [1, 1, 0, 0, 1, 1] | 443,169 | 0.00090 |
| [0, 1, 0, 1, 0, 0] | 3,142,395 | 0.00639 | [1, 1, 0, 1, 0, 0] | 3,455,308 | 0.00703 |
| [0, 1, 0, 1, 0, 1] | 560,545 | 0.00114 | [1, 1, 0, 1, 0, 1] | 550,655 | 0.00112 |
| [0, 1, 0, 1, 1, 0] | 5,397,830 | 0.01098 | [1, 1, 0, 1, 1, 0] | 6,574,824 | 0.01338 |
| [0, 1, 0, 1, 1, 1] | 1,139,751 | 0.00232 | [1, 1, 0, 1, 1, 1] | 1,149,593 | 0.00234 |
| [0, 1, 1, 0, 0, 0] | 4,462,547 | 0.00908 | [1, 1, 1, 0, 0, 0] | 589,074 | 0.00120 |
| [0, 1, 1, 0, 0, 1] | 4,297,173 | 0.00874 | [1, 1, 1, 0, 0, 1] | 565,157 | 0.00115 |
| [0, 1, 1, 0, 1, 0] | 5,386,879 | 0.01096 | [1, 1, 1, 0, 1, 0] | 624,925 | 0.00127 |
| [0, 1, 1, 0, 1, 1] | 8,595,064 | 0.01749 | [1, 1, 1, 0, 1, 1] | 884,239 | 0.00180 |
| [0, 1, 1, 1, 0, 0] | 972,876 | 0.00198 | [1, 1, 1, 1, 0, 0] | 1,000,862 | 0.00204 |
| [0, 1, 1, 1, 0, 1] | 1,077,323 | 0.00219 | [1, 1, 1, 1, 0, 1] | 1,295,744 | 0.00264 |
| [0, 1, 1, 1, 1, 0] | 1,391,814 | 0.00283 | [1, 1, 1, 1, 1, 0] | 1,625,378 | 0.00331 |
| [0, 1, 1, 1, 1, 1] | 2,334,636 | 0.00475 | [1, 1, 1, 1, 1, 1] | 3,314,303 | 0.00674 |
| Spike Entropy | | | 4.3685 | | |
| Firing Rate | | | 0.2438 | | |

Table 15: Distribution of the spike patterns in the encoding layer of Spikformer with STF on CIFAR-10 at $T = 6$.

| Spike Pattern | Count | Ratio | Spike Pattern | Count | Ratio |
| --- | --- | --- | --- | --- | --- |
| [0, 0, 0, 0, 0, 0, 0, 0] | 325,647,136 | 0.66253 | [1, 0, 0, 0, 0, 0, 0, 0] | 0 | 0 |
| [0, 0, 0, 0, 0, 0, 0, 1] | 780,771 | 0.00159 | [1, 0, 0, 0, 0, 0, 0, 1] | 0 | 0 |
| [0, 0, 0, 0, 0, 0, 1, 0] | 1,549,957 | 0.00315 | [1, 0, 0, 0, 0, 0, 1, 0] | 0 | 0 |
| [0, 0, 0, 0, 0, 0, 1, 1] | 0 | 0 | [1, 0, 0, 0, 0, 0, 1, 1] | 0 | 0 |
| [0, 0, 0, 0, 0, 1, 0, 0] | 3,343,140 | 0.0068 | [1, 0, 0, 0, 0, 1, 0, 0] | 0 | 0 |
| [0, 0, 0, 0, 0, 1, 0, 1] | 0 | 0 | [1, 0, 0, 0, 0, 1, 0, 1] | 0 | 0 |
| [0, 0, 0, 0, 0, 1, 1, 0] | 0 | 0 | [1, 0, 0, 0, 0, 1, 1, 0] | 0 | 0 |
| [0, 0, 0, 0, 0, 1, 1, 1] | 0 | 0 | [1, 0, 0, 0, 0, 1, 1, 1] | 0 | 0 |
| [0, 0, 0, 0, 1, 0, 0, 0] | 6,669,556 | 0.01357 | [1, 0, 0, 0, 1, 0, 0, 0] | 0 | 0 |
| [0, 0, 0, 0, 1, 0, 0, 1] | 0 | 0 | [1, 0, 0, 0, 1, 0, 0, 1] | 0 | 0 |
| [0, 0, 0, 0, 1, 0, 1, 0] | 0 | 0 | [1, 0, 0, 0, 1, 0, 1, 0] | 0 | 0 |
| [0, 0, 0, 0, 1, 0, 1, 1] | 0 | 0 | [1, 0, 0, 0, 1, 0, 1, 1] | 0 | 0 |
| [0, 0, 0, 0, 1, 1, 0, 0] | 0 | 0 | [1, 0, 0, 0, 1, 1, 0, 0] | 0 | 0 |
| [0, 0, 0, 0, 1, 1, 0, 1] | 0 | 0 | [1, 0, 0, 0, 1, 1, 0, 1] | 0 | 0 |
| [0, 0, 0, 0, 1, 1, 1, 0] | 0 | 0 | [1, 0, 0, 0, 1, 1, 1, 0] | 0 | 0 |
| [0, 0, 0, 0, 1, 1, 1, 1] | 0 | 0 | [1, 0, 0, 0, 1, 1, 1, 1] | 0 | 0 |
| [0, 0, 0, 1, 0, 0, 0, 0] | 0 | 0 | [1, 0, 0, 1, 0, 0, 0, 0] | 0 | 0 |
| [0, 0, 0, 1, 0, 0, 0, 1] | 13,888,765 | 0.02826 | [1, 0, 0, 1, 0, 0, 0, 1] | 0 | 0 |
| [0, 0, 0, 1, 0, 0, 1, 0] | 0 | 0 | [1, 0, 0, 1, 0, 0, 1, 0] | 0 | 0 |
| [0, 0, 0, 1, 0, 0, 1, 1] | 0 | 0 | [1, 0, 0, 1, 0, 0, 1, 1] | 0 | 0 |
| [0, 0, 0, 1, 0, 1, 0, 0] | 0 | 0 | [1, 0, 0, 1, 0, 1, 0, 0] | 0 | 0 |
| [0, 0, 0, 1, 0, 1, 0, 1] | 0 | 0 | [1, 0, 0, 1, 0, 1, 0, 1] | 0 | 0 |
| [0, 0, 0, 1, 0, 1, 1, 0] | 0 | 0 | [1, 0, 0, 1, 0, 1, 1, 0] | 0 | 0 |
| [0, 0, 0, 1, 0, 1, 1, 1] | 0 | 0 | [1, 0, 0, 1, 0, 1, 1, 1] | 0 | 0 |
| [0, 0, 0, 1, 1, 0, 0, 0] | 0 | 0 | [1, 0, 0, 1, 1, 0, 0, 0] | 0 | 0 |
| [0, 0, 0, 1, 1, 0, 0, 1] | 0 | 0 | [1, 0, 0, 1, 1, 0, 0, 1] | 0 | 0 |
| [0, 0, 0, 1, 1, 0, 1, 0] | 0 | 0 | [1, 0, 0, 1, 1, 0, 1, 0] | 0 | 0 |
| [0, 0, 0, 1, 1, 0, 1, 1] | 0 | 0 | [1, 0, 0, 1, 1, 0, 1, 1] | 0 | 0 |
| [0, 0, 0, 1, 1, 1, 0, 0] | 0 | 0 | [1, 0, 0, 1, 1, 1, 0, 0] | 0 | 0 |
| [0, 0, 0, 1, 1, 1, 0, 1] | 0 | 0 | [1, 0, 0, 1, 1, 1, 0, 1] | 0 | 0 |
| [0, 0, 0, 1, 1, 1, 1, 0] | 0 | 0 | [1, 0, 0, 1, 1, 1, 1, 0] | 0 | 0 |
| [0, 0, 0, 1, 1, 1, 1, 1] | 0 | 0 | [1, 0, 0, 1, 1, 1, 1, 1] | 0 | 0 |
| [0, 0, 1, 0, 0, 0, 0, 0] | 0 | 0 | [1, 0, 1, 0, 0, 0, 0, 0] | 0 | 0 |
| [0, 0, 1, 0, 0, 0, 0, 1] | 0 | 0 | [1, 0, 1, 0, 0, 0, 0, 1] | 0 | 0 |
| [0, 0, 1, 0, 0, 0, 1, 0] | 0 | 0 | [1, 0, 1, 0, 0, 0, 1, 0] | 0 | 0 |
| [0, 0, 1, 0, 0, 0, 1, 1] | 0 | 0 | [1, 0, 1, 0, 0, 0, 1, 1] | 0 | 0 |
| [0, 0, 1, 0, 0, 1, 0, 0] | 29,723,360 | 0.06047 | [1, 0, 1, 0, 0, 1, 0, 0] | 0 | 0 |
| [0, 0, 1, 0, 0, 1, 0, 1] | 0 | 0 | [1, 0, 1, 0, 0, 1, 0, 1] | 0 | 0 |
| [0, 0, 1, 0, 0, 1, 1, 0] | 0 | 0 | [1, 0, 1, 0, 0, 1, 1, 0] | 0 | 0 |
| [0, 0, 1, 0, 0, 1, 1, 1] | 0 | 0 | [1, 0, 1, 0, 0, 1, 1, 1] | 0 | 0 |
| [0, 0, 1, 0, 1, 0, 0, 0] | 0 | 0 | [1, 0, 1, 0, 1, 0, 0, 0] | 0 | 0 |
| [0, 0, 1, 0, 1, 0, 0, 1] | 0 | 0 | [1, 0, 1, 0, 1, 0, 0, 1] | 0 | 0 |
| [0, 0, 1, 0, 1, 0, 1, 0] | 0 | 0 | [1, 0, 1, 0, 1, 0, 1, 0] | 0 | 0 |
| [0, 0, 1, 0, 1, 0, 1, 1] | 0 | 0 | [1, 0, 1, 0, 1, 0, 1, 1] | 0 | 0 |
| [0, 0, 1, 0, 1, 1, 0, 0] | 0 | 0 | [1, 0, 1, 0, 1, 1, 0, 0] | 0 | 0 |
| [0, 0, 1, 0, 1, 1, 0, 1] | 0 | 0 | [1, 0, 1, 0, 1, 1, 0, 1] | 0 | 0 |
| [0, 0, 1, 0, 1, 1, 1, 0] | 0 | 0 | [1, 0, 1, 0, 1, 1, 1, 0] | 0 | 0 |
| [0, 0, 1, 0, 1, 1, 1, 1] | 0 | 0 | [1, 0, 1, 0, 1, 1, 1, 1] | 0 | 0 |
| [0, 0, 1, 1, 0, 0, 0, 0] | 0 | 0 | [1, 0, 1, 1, 0, 0, 0, 0] | 0 | 0 |
| [0, 0, 1, 1, 0, 0, 0, 1] | 0 | 0 | [1, 0, 1, 1, 0, 0, 0, 1] | 0 | 0 |
| [0, 0, 1, 1, 0, 0, 1, 0] | 0 | 0 | [1, 0, 1, 1, 0, 0, 1, 0] | 0 | 0 |
| [0, 0, 1, 1, 0, 0, 1, 1] | 0 | 0 | [1, 0, 1, 1, 0, 0, 1, 1] | 0 | 0 |
| [0, 0, 1, 1, 0, 1, 0, 0] | 0 | 0 | [1, 0, 1, 1, 0, 1, 0, 0] | 0 | 0 |
| [0, 0, 1, 1, 0, 1, 0, 1] | 0 | 0 | [1, 0, 1, 1, 0, 1, 0, 1] | 0 | 0 |

| Spike Pattern | Count | Ratio | Spike Pattern | Count | Ratio |
|---|---|---|---|---|---|
| [0, 0, 1, 1, 0, 1, 1, 0] | 0 | 0 | [1, 0, 1, 1, 0, 1, 1, 0] | 0 | 0 |
| [0, 0, 1, 1, 0, 1, 1, 1] | 0 | 0 | [1, 0, 1, 1, 0, 1, 1, 1] | 0 | 0 |
| [0, 0, 1, 1, 1, 0, 0, 0] | 0 | 0 | [1, 0, 1, 1, 1, 0, 0, 0] | 0 | 0 |
| [0, 0, 1, 1, 1, 0, 0, 1] | 0 | 0 | [1, 0, 1, 1, 1, 0, 0, 1] | 0 | 0 |
| [0, 0, 1, 1, 1, 0, 1, 0] | 0 | 0 | [1, 0, 1, 1, 1, 0, 1, 0] | 0 | 0 |
| [0, 0, 1, 1, 1, 0, 1, 1] | 0 | 0 | [1, 0, 1, 1, 1, 0, 1, 1] | 0 | 0 |
| [0, 0, 1, 1, 1, 1, 0, 0] | 0 | 0 | [1, 0, 1, 1, 1, 1, 0, 0] | 0 | 0 |
| [0, 0, 1, 1, 1, 1, 0, 1] | 0 | 0 | [1, 0, 1, 1, 1, 1, 0, 1] | 0 | 0 |
| [0, 0, 1, 1, 1, 1, 1, 0] | 0 | 0 | [1, 0, 1, 1, 1, 1, 1, 0] | 0 | 0 |
| [0, 0, 1, 1, 1, 1, 1, 1] | 0 | 0 | [1, 0, 1, 1, 1, 1, 1, 1] | 0 | 0 |
| [0, 1, 0, 0, 0, 0, 0, 0] | 0 | 0 | [1, 1, 0, 0, 0, 0, 0, 0] | 0 | 0 |
| [0, 1, 0, 0, 0, 0, 0, 1] | 0 | 0 | [1, 1, 0, 0, 0, 0, 0, 1] | 0 | 0 |
| [0, 1, 0, 0, 0, 0, 1, 0] | 0 | 0 | [1, 1, 0, 0, 0, 0, 1, 0] | 0 | 0 |
| [0, 1, 0, 0, 0, 0, 1, 1] | 0 | 0 | [1, 1, 0, 0, 0, 0, 1, 1] | 0 | 0 |
| [0, 1, 0, 0, 0, 1, 0, 0] | 0 | 0 | [1, 1, 0, 0, 0, 1, 0, 0] | 0 | 0 |
| [0, 1, 0, 0, 0, 1, 0, 1] | 0 | 0 | [1, 1, 0, 0, 0, 1, 0, 1] | 0 | 0 |
| [0, 1, 0, 0, 0, 1, 1, 0] | 0 | 0 | [1, 1, 0, 0, 0, 1, 1, 0] | 0 | 0 |
| [0, 1, 0, 0, 0, 1, 1, 1] | 0 | 0 | [1, 1, 0, 0, 0, 1, 1, 1] | 0 | 0 |
| [0, 1, 0, 0, 1, 0, 0, 0] | 0 | 0 | [1, 1, 0, 0, 1, 0, 0, 0] | 0 | 0 |
| [0, 1, 0, 0, 1, 0, 0, 1] | 0 | 0 | [1, 1, 0, 0, 1, 0, 0, 1] | 0 | 0 |
| [0, 1, 0, 0, 1, 0, 1, 0] | 0 | 0 | [1, 1, 0, 0, 1, 0, 1, 0] | 0 | 0 |
| [0, 1, 0, 0, 1, 0, 1, 1] | 0 | 0 | [1, 1, 0, 0, 1, 0, 1, 1] | 0 | 0 |
| [0, 1, 0, 0, 1, 1, 0, 0] | 0 | 0 | [1, 1, 0, 0, 1, 1, 0, 0] | 0 | 0 |
| [0, 1, 0, 0, 1, 1, 0, 1] | 0 | 0 | [1, 1, 0, 0, 1, 1, 0, 1] | 0 | 0 |
| [0, 1, 0, 0, 1, 1, 1, 0] | 0 | 0 | [1, 1, 0, 0, 1, 1, 1, 0] | 0 | 0 |
| [0, 1, 0, 0, 1, 1, 1, 1] | 0 | 0 | [1, 1, 0, 0, 1, 1, 1, 1] | 0 | 0 |
| [0, 1, 0, 1, 0, 0, 0, 0] | 0 | 0 | [1, 1, 0, 1, 0, 0, 0, 0] | 0 | 0 |
| [0, 1, 0, 1, 0, 0, 0, 1] | 0 | 0 | [1, 1, 0, 1, 0, 0, 0, 1] | 0 | 0 |
| [0, 1, 0, 1, 0, 0, 1, 0] | 0 | 0 | [1, 1, 0, 1, 0, 0, 1, 0] | 0 | 0 |
| [0, 1, 0, 1, 0, 0, 1, 1] | 0 | 0 | [1, 1, 0, 1, 0, 0, 1, 1] | 0 | 0 |
| [0, 1, 0, 1, 0, 1, 0, 0] | 0 | 0 | [1, 1, 0, 1, 0, 1, 0, 0] | 0 | 0 |
| [0, 1, 0, 1, 0, 1, 0, 1] | 57,422,345 | 0.11683 | [1, 1, 0, 1, 0, 1, 0, 1] | 0 | 0 |
| [0, 1, 0, 1, 0, 1, 1, 0] | 0 | 0 | [1, 1, 0, 1, 0, 1, 1, 0] | 0 | 0 |
| [0, 1, 0, 1, 0, 1, 1, 1] | 0 | 0 | [1, 1, 0, 1, 0, 1, 1, 1] | 0 | 0 |
| [0, 1, 0, 1, 1, 0, 0, 0] | 0 | 0 | [1, 1, 0, 1, 1, 0, 0, 0] | 0 | 0 |
| [0, 1, 0, 1, 1, 0, 0, 1] | 0 | 0 | [1, 1, 0, 1, 1, 0, 0, 1] | 0 | 0 |
| [0, 1, 0, 1, 1, 0, 1, 0] | 0 | 0 | [1, 1, 0, 1, 1, 0, 1, 0] | 0 | 0 |
| [0, 1, 0, 1, 1, 0, 1, 1] | 0 | 0 | [1, 1, 0, 1, 1, 0, 1, 1] | 0 | 0 |
| [0, 1, 0, 1, 1, 1, 0, 0] | 0 | 0 | [1, 1, 0, 1, 1, 1, 0, 0] | 0 | 0 |
| [0, 1, 0, 1, 1, 1, 0, 1] | 0 | 0 | [1, 1, 0, 1, 1, 1, 0, 1] | 0 | 0 |
| [0, 1, 0, 1, 1, 1, 1, 0] | 0 | 0 | [1, 1, 0, 1, 1, 1, 1, 0] | 0 | 0 |
| [0, 1, 0, 1, 1, 1, 1, 1] | 0 | 0 | [1, 1, 0, 1, 1, 1, 1, 1] | 0 | 0 |
| [0, 1, 1, 0, 0, 0, 0, 0] | 0 | 0 | [1, 1, 1, 0, 0, 0, 0, 0] | 0 | 0 |
| [0, 1, 1, 0, 0, 0, 0, 1] | 0 | 0 | [1, 1, 1, 0, 0, 0, 0, 1] | 0 | 0 |
| [0, 1, 1, 0, 0, 0, 1, 0] | 0 | 0 | [1, 1, 1, 0, 0, 0, 1, 0] | 0 | 0 |
| [0, 1, 1, 0, 0, 0, 1, 1] | 0 | 0 | [1, 1, 1, 0, 0, 0, 1, 1] | 0 | 0 |
| [0, 1, 1, 0, 0, 1, 0, 0] | 0 | 0 | [1, 1, 1, 0, 0, 1, 0, 0] | 0 | 0 |
| [0, 1, 1, 0, 0, 1, 0, 1] | 0 | 0 | [1, 1, 1, 0, 0, 1, 0, 1] | 0 | 0 |
| [0, 1, 1, 0, 0, 1, 1, 0] | 0 | 0 | [1, 1, 1, 0, 0, 1, 1, 0] | 0 | 0 |
| [0, 1, 1, 0, 0, 1, 1, 1] | 0 | 0 | [1, 1, 1, 0, 0, 1, 1, 1] | 0 | 0 |
| [0, 1, 1, 0, 1, 0, 0, 0] | 0 | 0 | [1, 1, 1, 0, 1, 0, 0, 0] | 0 | 0 |
| [0, 1, 1, 0, 1, 0, 0, 1] | 0 | 0 | [1, 1, 1, 0, 1, 0, 0, 1] | 0 | 0 |
| [0, 1, 1, 0, 1, 0, 1, 0] | 0 | 0 | [1, 1, 1, 0, 1, 0, 1, 0] | 0 | 0 |
| [0, 1, 1, 0, 1, 0, 1, 1] | 0 | 0 | [1, 1, 1, 0, 1, 0, 1, 1] | 0 | 0 |
| [0, 1, 1, 0, 1, 1, 0, 0] | 0 | 0 | [1, 1, 1, 0, 1, 1, 0, 0] | 0 | 0 |

| Spike Pattern | Count | Ratio | Spike Pattern | Count | Ratio |
|---|---|---|---|---|---|
| [0, 1, 1, 0, 1, 1, 0, 1] | 0 | 0 | [1, 1, 1, 0, 1, 1, 0, 1] | 0 | 0 |
| [0, 1, 1, 0, 1, 1, 1, 0] | 0 | 0 | [1, 1, 1, 0, 1, 1, 1, 0] | 0 | 0 |
| [0, 1, 1, 0, 1, 1, 1, 1] | 0 | 0 | [1, 1, 1, 0, 1, 1, 1, 1] | 0 | 0 |
| [0, 1, 1, 1, 0, 0, 0, 0] | 0 | 0 | [1, 1, 1, 1, 0, 0, 0, 0] | 0 | 0 |
| [0, 1, 1, 1, 0, 0, 0, 1] | 0 | 0 | [1, 1, 1, 1, 0, 0, 0, 1] | 0 | 0 |
| [0, 1, 1, 1, 0, 0, 1, 0] | 0 | 0 | [1, 1, 1, 1, 0, 0, 1, 0] | 0 | 0 |
| [0, 1, 1, 1, 0, 0, 1, 1] | 0 | 0 | [1, 1, 1, 1, 0, 0, 1, 1] | 0 | 0 |
| [0, 1, 1, 1, 0, 1, 0, 0] | 0 | 0 | [1, 1, 1, 1, 0, 1, 0, 0] | 0 | 0 |
| [0, 1, 1, 1, 0, 1, 0, 1] | 0 | 0 | [1, 1, 1, 1, 0, 1, 0, 1] | 0 | 0 |
| [0, 1, 1, 1, 0, 1, 1, 0] | 0 | 0 | [1, 1, 1, 1, 0, 1, 1, 0] | 0 | 0 |
| [0, 1, 1, 1, 0, 1, 1, 1] | 0 | 0 | [1, 1, 1, 1, 0, 1, 1, 1] | 0 | 0 |
| [0, 1, 1, 1, 1, 0, 0, 0] | 0 | 0 | [1, 1, 1, 1, 1, 0, 0, 0] | 0 | 0 |
| [0, 1, 1, 1, 1, 0, 0, 1] | 0 | 0 | [1, 1, 1, 1, 1, 0, 0, 1] | 0 | 0 |
| [0, 1, 1, 1, 1, 0, 1, 0] | 0 | 0 | [1, 1, 1, 1, 1, 0, 1, 0] | 0 | 0 |
| [0, 1, 1, 1, 1, 0, 1, 1] | 0 | 0 | [1, 1, 1, 1, 1, 0, 1, 1] | 0 | 0 |
| [0, 1, 1, 1, 1, 1, 0, 0] | 0 | 0 | [1, 1, 1, 1, 1, 1, 0, 0] | 0 | 0 |
| [0, 1, 1, 1, 1, 1, 0, 1] | 0 | 0 | [1, 1, 1, 1, 1, 1, 0, 1] | 0 | 0 |
| [0, 1, 1, 1, 1, 1, 1, 0] | 0 | 0 | [1, 1, 1, 1, 1, 1, 1, 0] | 0 | 0 |
| [0, 1, 1, 1, 1, 1, 1, 1] | 0 | 0 | [1, 1, 1, 1, 1, 1, 1, 1] | 52,494,970 | 0.1068 |
| Spike Entropy | | | 1.6643 | | |
| Firing Rate | | | 0.1905 | | |

Table 16: Distribution of the spike patterns in the encoding layer of Spikformer with direct coding on CIFAR-10 at $T = 8$.

| Spike Pattern | Count | Ratio | Spike Pattern | Count | Ratio |
| --- | --- | --- | --- | --- | --- |
| [0, 0, 0, 0, 0, 0, 0, 0] | 330,957,691 | 0.67334 | [1, 0, 0, 0, 0, 0, 0, 0] | 0 | 0 |
| [0, 0, 0, 0, 0, 0, 0, 1] | 771,824 | 0.00157 | [1, 0, 0, 0, 0, 0, 0, 1] | 0 | 0 |
| [0, 0, 0, 0, 0, 0, 1, 0] | 1,528,431 | 0.00311 | [1, 0, 0, 0, 0, 0, 1, 0] | 0 | 0 |
| [0, 0, 0, 0, 0, 0, 1, 1] | 0 | 0 | [1, 0, 0, 0, 0, 0, 1, 1] | 0 | 0 |
| [0, 0, 0, 0, 0, 1, 0, 0] | 3,246,666 | 0.00660 | [1, 0, 0, 0, 0, 1, 0, 0] | 0 | 0 |
| [0, 0, 0, 0, 0, 1, 0, 1] | 0 | 0 | [1, 0, 0, 0, 0, 1, 0, 1] | 0 | 0 |
| [0, 0, 0, 0, 0, 1, 1, 0] | 0 | 0 | [1, 0, 0, 0, 0, 1, 1, 0] | 0 | 0 |
| [0, 0, 0, 0, 0, 1, 1, 1] | 0 | 0 | [1, 0, 0, 0, 0, 1, 1, 1] | 0 | 0 |
| [0, 0, 0, 0, 1, 0, 0, 0] | 6,402,526 | 0.01303 | [1, 0, 0, 0, 1, 0, 0, 0] | 0 | 0 |
| [0, 0, 0, 0, 1, 0, 0, 1] | 0 | 0 | [1, 0, 0, 0, 1, 0, 0, 1] | 0 | 0 |
| [0, 0, 0, 0, 1, 0, 1, 0] | 0 | 0 | [1, 0, 0, 0, 1, 0, 1, 0] | 0 | 0 |
| [0, 0, 0, 0, 1, 0, 1, 1] | 0 | 0 | [1, 0, 0, 0, 1, 0, 1, 1] | 0 | 0 |
| [0, 0, 0, 0, 1, 1, 0, 0] | 0 | 0 | [1, 0, 0, 0, 1, 1, 0, 0] | 0 | 0 |
| [0, 0, 0, 0, 1, 1, 0, 1] | 0 | 0 | [1, 0, 0, 0, 1, 1, 0, 1] | 0 | 0 |
| [0, 0, 0, 0, 1, 1, 1, 0] | 0 | 0 | [1, 0, 0, 0, 1, 1, 1, 0] | 0 | 0 |
| [0, 0, 0, 0, 1, 1, 1, 1] | 0 | 0 | [1, 0, 0, 0, 1, 1, 1, 1] | 0 | 0 |
| [0, 0, 0, 1, 0, 0, 0, 0] | 0 | 0 | [1, 0, 0, 1, 0, 0, 0, 0] | 0 | 0 |
| [0, 0, 0, 1, 0, 0, 0, 1] | 13,208,866 | 0.02687 | [1, 0, 0, 1, 0, 0, 0, 1] | 0 | 0 |
| [0, 0, 0, 1, 0, 0, 1, 0] | 0 | 0 | [1, 0, 0, 1, 0, 0, 1, 0] | 0 | 0 |
| [0, 0, 0, 1, 0, 0, 1, 1] | 0 | 0 | [1, 0, 0, 1, 0, 0, 1, 1] | 0 | 0 |
| [0, 0, 0, 1, 0, 1, 0, 0] | 0 | 0 | [1, 0, 0, 1, 0, 1, 0, 0] | 0 | 0 |
| [0, 0, 0, 1, 0, 1, 0, 1] | 0 | 0 | [1, 0, 0, 1, 0, 1, 0, 1] | 0 | 0 |
| [0, 0, 0, 1, 0, 1, 1, 0] | 0 | 0 | [1, 0, 0, 1, 0, 1, 1, 0] | 0 | 0 |
| [0, 0, 0, 1, 0, 1, 1, 1] | 0 | 0 | [1, 0, 0, 1, 0, 1, 1, 1] | 0 | 0 |
| [0, 0, 0, 1, 1, 0, 0, 0] | 0 | 0 | [1, 0, 0, 1, 1, 0, 0, 0] | 0 | 0 |
| [0, 0, 0, 1, 1, 0, 0, 1] | 0 | 0 | [1, 0, 0, 1, 1, 0, 0, 1] | 0 | 0 |
| [0, 0, 0, 1, 1, 0, 1, 0] | 0 | 0 | [1, 0, 0, 1, 1, 0, 1, 0] | 0 | 0 |
| [0, 0, 0, 1, 1, 0, 1, 1] | 0 | 0 | [1, 0, 0, 1, 1, 0, 1, 1] | 0 | 0 |
| [0, 0, 0, 1, 1, 1, 0, 0] | 0 | 0 | [1, 0, 0, 1, 1, 1, 0, 0] | 0 | 0 |
| [0, 0, 0, 1, 1, 1, 0, 1] | 0 | 0 | [1, 0, 0, 1, 1, 1, 0, 1] | 0 | 0 |
| [0, 0, 0, 1, 1, 1, 1, 0] | 0 | 0 | [1, 0, 0, 1, 1, 1, 1, 0] | 0 | 0 |
| [0, 0, 0, 1, 1, 1, 1, 1] | 0 | 0 | [1, 0, 0, 1, 1, 1, 1, 1] | 0 | 0 |
| [0, 0, 1, 0, 0, 0, 0, 0] | 0 | 0 | [1, 0, 1, 0, 0, 0, 0, 0] | 0 | 0 |
| [0, 0, 1, 0, 0, 0, 0, 1] | 0 | 0 | [1, 0, 1, 0, 0, 0, 0, 1] | 0 | 0 |
| [0, 0, 1, 0, 0, 0, 1, 0] | 0 | 0 | [1, 0, 1, 0, 0, 0, 1, 0] | 0 | 0 |
| [0, 0, 1, 0, 0, 0, 1, 1] | 0 | 0 | [1, 0, 1, 0, 0, 0, 1, 1] | 0 | 0 |
| [0, 0, 1, 0, 0, 1, 0, 0] | 27,747,968 | 0.05645 | [1, 0, 1, 0, 0, 1, 0, 0] | 0 | 0 |
| [0, 0, 1, 0, 0, 1, 0, 1] | 0 | 0 | [1, 0, 1, 0, 0, 1, 0, 1] | 0 | 0 |
| [0, 0, 1, 0, 0, 1, 1, 0] | 0 | 0 | [1, 0, 1, 0, 0, 1, 1, 0] | 0 | 0 |
| [0, 0, 1, 0, 0, 1, 1, 1] | 0 | 0 | [1, 0, 1, 0, 0, 1, 1, 1] | 0 | 0 |
| [0, 0, 1, 0, 1, 0, 0, 0] | 0 | 0 | [1, 0, 1, 0, 1, 0, 0, 0] | 0 | 0 |
| [0, 0, 1, 0, 1, 0, 0, 1] | 0 | 0 | [1, 0, 1, 0, 1, 0, 0, 1] | 0 | 0 |
| [0, 0, 1, 0, 1, 0, 1, 0] | 0 | 0 | [1, 0, 1, 0, 1, 0, 1, 0] | 0 | 0 |
| [0, 0, 1, 0, 1, 0, 1, 1] | 0 | 0 | [1, 0, 1, 0, 1, 0, 1, 1] | 0 | 0 |
| [0, 0, 1, 0, 1, 1, 0, 0] | 0 | 0 | [1, 0, 1, 0, 1, 1, 0, 0] | 0 | 0 |
| [0, 0, 1, 0, 1, 1, 0, 1] | 0 | 0 | [1, 0, 1, 0, 1, 1, 0, 1] | 0 | 0 |
| [0, 0, 1, 0, 1, 1, 1, 0] | 0 | 0 | [1, 0, 1, 0, 1, 1, 1, 0] | 0 | 0 |
| [0, 0, 1, 0, 1, 1, 1, 1] | 0 | 0 | [1, 0, 1, 0, 1, 1, 1, 1] | 0 | 0 |
| [0, 0, 1, 1, 0, 0, 0, 0] | 0 | 0 | [1, 0, 1, 1, 0, 0, 0, 0] | 0 | 0 |
| [0, 0, 1, 1, 0, 0, 0, 1] | 0 | 0 | [1, 0, 1, 1, 0, 0, 0, 1] | 0 | 0 |
| [0, 0, 1, 1, 0, 0, 1, 0] | 0 | 0 | [1, 0, 1, 1, 0, 0, 1, 0] | 0 | 0 |
| [0, 0, 1, 1, 0, 0, 1, 1] | 0 | 0 | [1, 0, 1, 1, 0, 0, 1, 1] | 0 | 0 |
| [0, 0, 1, 1, 0, 1, 0, 0] | 0 | 0 | [1, 0, 1, 1, 0, 1, 0, 0] | 0 | 0 |
| [0, 0, 1, 1, 0, 1, 0, 1] | 0 | 0 | [1, 0, 1, 1, 0, 1, 0, 1] | 0 | 0 |

| Spike Pattern | Count | Ratio | Spike Pattern | Count | Ratio |
| --- | --- | --- | --- | --- | --- |
| [0, 0, 1, 1, 0, 1, 1, 0] | 0 | 0 | [1, 0, 1, 1, 0, 1, 1, 0] | 0 | 0 |
| [0, 0, 1, 1, 0, 1, 1, 1] | 0 | 0 | [1, 0, 1, 1, 0, 1, 1, 1] | 0 | 0 |
| [0, 0, 1, 1, 1, 0, 0, 0] | 0 | 0 | [1, 0, 1, 1, 1, 0, 0, 0] | 0 | 0 |
| [0, 0, 1, 1, 1, 0, 0, 1] | 0 | 0 | [1, 0, 1, 1, 1, 0, 0, 1] | 0 | 0 |
| [0, 0, 1, 1, 1, 0, 1, 0] | 0 | 0 | [1, 0, 1, 1, 1, 0, 1, 0] | 0 | 0 |
| [0, 0, 1, 1, 1, 0, 1, 1] | 0 | 0 | [1, 0, 1, 1, 1, 0, 1, 1] | 0 | 0 |
| [0, 0, 1, 1, 1, 1, 0, 0] | 0 | 0 | [1, 0, 1, 1, 1, 1, 0, 0] | 0 | 0 |
| [0, 0, 1, 1, 1, 1, 0, 1] | 0 | 0 | [1, 0, 1, 1, 1, 1, 0, 1] | 0 | 0 |
| [0, 0, 1, 1, 1, 1, 1, 0] | 0 | 0 | [1, 0, 1, 1, 1, 1, 1, 0] | 0 | 0 |
| [0, 0, 1, 1, 1, 1, 1, 1] | 0 | 0 | [1, 0, 1, 1, 1, 1, 1, 1] | 0 | 0 |
| [0, 1, 0, 0, 0, 0, 0, 0] | 0 | 0 | [1, 1, 0, 0, 0, 0, 0, 0] | 0 | 0 |
| [0, 1, 0, 0, 0, 0, 0, 1] | 0 | 0 | [1, 1, 0, 0, 0, 0, 0, 1] | 0 | 0 |
| [0, 1, 0, 0, 0, 0, 1, 0] | 0 | 0 | [1, 1, 0, 0, 0, 0, 1, 0] | 0 | 0 |
| [0, 1, 0, 0, 0, 0, 1, 1] | 0 | 0 | [1, 1, 0, 0, 0, 0, 1, 1] | 0 | 0 |
| [0, 1, 0, 0, 0, 1, 0, 0] | 0 | 0 | [1, 1, 0, 0, 0, 1, 0, 0] | 0 | 0 |
| [0, 1, 0, 0, 0, 1, 0, 1] | 0 | 0 | [1, 1, 0, 0, 0, 1, 0, 1] | 0 | 0 |
| [0, 1, 0, 0, 0, 1, 1, 0] | 0 | 0 | [1, 1, 0, 0, 0, 1, 1, 0] | 0 | 0 |
| [0, 1, 0, 0, 0, 1, 1, 1] | 0 | 0 | [1, 1, 0, 0, 0, 1, 1, 1] | 0 | 0 |
| [0, 1, 0, 0, 1, 0, 0, 0] | 0 | 0 | [1, 1, 0, 0, 1, 0, 0, 0] | 0 | 0 |
| [0, 1, 0, 0, 1, 0, 0, 1] | 0 | 0 | [1, 1, 0, 0, 1, 0, 0, 1] | 0 | 0 |
| [0, 1, 0, 0, 1, 0, 1, 0] | 0 | 0 | [1, 1, 0, 0, 1, 0, 1, 0] | 0 | 0 |
| [0, 1, 0, 0, 1, 0, 1, 1] | 0 | 0 | [1, 1, 0, 0, 1, 0, 1, 1] | 0 | 0 |
| [0, 1, 0, 0, 1, 1, 0, 0] | 0 | 0 | [1, 1, 0, 0, 1, 1, 0, 0] | 0 | 0 |
| [0, 1, 0, 0, 1, 1, 0, 1] | 0 | 0 | [1, 1, 0, 0, 1, 1, 0, 1] | 0 | 0 |
| [0, 1, 0, 0, 1, 1, 1, 0] | 0 | 0 | [1, 1, 0, 0, 1, 1, 1, 0] | 0 | 0 |
| [0, 1, 0, 0, 1, 1, 1, 1] | 0 | 0 | [1, 1, 0, 0, 1, 1, 1, 1] | 0 | 0 |
| [0, 1, 0, 1, 0, 0, 0, 0] | 0 | 0 | [1, 1, 0, 1, 0, 0, 0, 0] | 0 | 0 |
| [0, 1, 0, 1, 0, 0, 0, 1] | 0 | 0 | [1, 1, 0, 1, 0, 0, 0, 1] | 0 | 0 |
| [0, 1, 0, 1, 0, 0, 1, 0] | 0 | 0 | [1, 1, 0, 1, 0, 0, 1, 0] | 0 | 0 |
| [0, 1, 0, 1, 0, 0, 1, 1] | 0 | 0 | [1, 1, 0, 1, 0, 0, 1, 1] | 0 | 0 |
| [0, 1, 0, 1, 0, 1, 0, 0] | 0 | 0 | [1, 1, 0, 1, 0, 1, 0, 0] | 0 | 0 |
| [0, 1, 0, 1, 0, 1, 0, 1] | 55,169,205 | 0.11224 | [1, 1, 0, 1, 0, 1, 0, 1] | 0 | 0 |
| [0, 1, 0, 1, 0, 1, 1, 0] | 0 | 0 | [1, 1, 0, 1, 0, 1, 1, 0] | 0 | 0 |
| [0, 1, 0, 1, 0, 1, 1, 1] | 0 | 0 | [1, 1, 0, 1, 0, 1, 1, 1] | 0 | 0 |
| [0, 1, 0, 1, 1, 0, 0, 0] | 0 | 0 | [1, 1, 0, 1, 1, 0, 0, 0] | 0 | 0 |
| [0, 1, 0, 1, 1, 0, 0, 1] | 0 | 0 | [1, 1, 0, 1, 1, 0, 0, 1] | 0 | 0 |
| [0, 1, 0, 1, 1, 0, 1, 0] | 0 | 0 | [1, 1, 0, 1, 1, 0, 1, 0] | 0 | 0 |
| [0, 1, 0, 1, 1, 0, 1, 1] | 0 | 0 | [1, 1, 0, 1, 1, 0, 1, 1] | 0 | 0 |
| [0, 1, 0, 1, 1, 1, 0, 0] | 0 | 0 | [1, 1, 0, 1, 1, 1, 0, 0] | 0 | 0 |
| [0, 1, 0, 1, 1, 1, 0, 1] | 0 | 0 | [1, 1, 0, 1, 1, 1, 0, 1] | 0 | 0 |
| [0, 1, 0, 1, 1, 1, 1, 0] | 0 | 0 | [1, 1, 0, 1, 1, 1, 1, 0] | 0 | 0 |
| [0, 1, 0, 1, 1, 1, 1, 1] | 0 | 0 | [1, 1, 0, 1, 1, 1, 1, 1] | 0 | 0 |
| [0, 1, 1, 0, 0, 0, 0, 0] | 0 | 0 | [1, 1, 1, 0, 0, 0, 0, 0] | 0 | 0 |
| [0, 1, 1, 0, 0, 0, 0, 1] | 0 | 0 | [1, 1, 1, 0, 0, 0, 0, 1] | 0 | 0 |
| [0, 1, 1, 0, 0, 0, 1, 0] | 0 | 0 | [1, 1, 1, 0, 0, 0, 1, 0] | 0 | 0 |
| [0, 1, 1, 0, 0, 0, 1, 1] | 0 | 0 | [1, 1, 1, 0, 0, 0, 1, 1] | 0 | 0 |
| [0, 1, 1, 0, 0, 1, 0, 0] | 0 | 0 | [1, 1, 1, 0, 0, 1, 0, 0] | 0 | 0 |
| [0, 1, 1, 0, 0, 1, 0, 1] | 0 | 0 | [1, 1, 1, 0, 0, 1, 0, 1] | 0 | 0 |
| [0, 1, 1, 0, 0, 1, 1, 0] | 0 | 0 | [1, 1, 1, 0, 0, 1, 1, 0] | 0 | 0 |
| [0, 1, 1, 0, 0, 1, 1, 1] | 0 | 0 | [1, 1, 1, 0, 0, 1, 1, 1] | 0 | 0 |
| [0, 1, 1, 0, 1, 0, 0, 0] | 0 | 0 | [1, 1, 1, 0, 1, 0, 0, 0] | 0 | 0 |
| [0, 1, 1, 0, 1, 0, 0, 1] | 0 | 0 | [1, 1, 1, 0, 1, 0, 0, 1] | 0 | 0 |
| [0, 1, 1, 0, 1, 0, 1, 0] | 0 | 0 | [1, 1, 1, 0, 1, 0, 1, 0] | 0 | 0 |
| [0, 1, 1, 0, 1, 0, 1, 1] | 0 | 0 | [1, 1, 1, 0, 1, 0, 1, 1] | 0 | 0 |
| [0, 1, 1, 0, 1, 1, 0, 0] | 0 | 0 | [1, 1, 1, 0, 1, 1, 0, 0] | 0 | 0 |

| Spike Pattern | Count | Ratio | Spike Pattern | Count | Ratio |
| --- | --- | --- | --- | --- | --- |
| [0, 1, 1, 0, 1, 1, 0, 1] | 0 | 0 | [1, 1, 1, 0, 1, 1, 0, 1] | 0 | 0 |
| [0, 1, 1, 0, 1, 1, 1, 0] | 0 | 0 | [1, 1, 1, 0, 1, 1, 1, 0] | 0 | 0 |
| [0, 1, 1, 0, 1, 1, 1, 1] | 0 | 0 | [1, 1, 1, 0, 1, 1, 1, 1] | 0 | 0 |
| [0, 1, 1, 1, 0, 0, 0, 0] | 0 | 0 | [1, 1, 1, 1, 0, 0, 0, 0] | 0 | 0 |
| [0, 1, 1, 1, 0, 0, 0, 1] | 0 | 0 | [1, 1, 1, 1, 0, 0, 0, 1] | 0 | 0 |
| [0, 1, 1, 1, 0, 0, 1, 0] | 0 | 0 | [1, 1, 1, 1, 0, 0, 1, 0] | 0 | 0 |
| [0, 1, 1, 1, 0, 0, 1, 1] | 0 | 0 | [1, 1, 1, 1, 0, 0, 1, 1] | 0 | 0 |
| [0, 1, 1, 1, 0, 1, 0, 0] | 0 | 0 | [1, 1, 1, 1, 0, 1, 0, 0] | 0 | 0 |
| [0, 1, 1, 1, 0, 1, 0, 1] | 0 | 0 | [1, 1, 1, 1, 0, 1, 0, 1] | 0 | 0 |
| [0, 1, 1, 1, 0, 1, 1, 0] | 0 | 0 | [1, 1, 1, 1, 0, 1, 1, 0] | 0 | 0 |
| [0, 1, 1, 1, 0, 1, 1, 1] | 0 | 0 | [1, 1, 1, 1, 0, 1, 1, 1] | 0 | 0 |
| [0, 1, 1, 1, 1, 0, 0, 0] | 0 | 0 | [1, 1, 1, 1, 1, 0, 0, 0] | 0 | 0 |
| [0, 1, 1, 1, 1, 0, 0, 1] | 0 | 0 | [1, 1, 1, 1, 1, 0, 0, 1] | 0 | 0 |
| [0, 1, 1, 1, 1, 0, 1, 0] | 0 | 0 | [1, 1, 1, 1, 1, 0, 1, 0] | 0 | 0 |
| [0, 1, 1, 1, 1, 0, 1, 1] | 0 | 0 | [1, 1, 1, 1, 1, 0, 1, 1] | 0 | 0 |
| [0, 1, 1, 1, 1, 1, 0, 0] | 0 | 0 | [1, 1, 1, 1, 1, 1, 0, 0] | 0 | 0 |
| [0, 1, 1, 1, 1, 1, 0, 1] | 0 | 0 | [1, 1, 1, 1, 1, 1, 0, 1] | 0 | 0 |
| [0, 1, 1, 1, 1, 1, 1, 0] | 0 | 0 | [1, 1, 1, 1, 1, 1, 1, 0] | 0 | 0 |
| [0, 1, 1, 1, 1, 1, 1, 1] | 0 | 0 | [1, 1, 1, 1, 1, 1, 1, 1] | 52,486,823 | 0.10679 |
| Spike Entropy | | | 1.6272 | | |
| Firing Rate | | | 0.1868 | | |

Table 17: Distribution of the spike patterns in the encoding layer of Spikformer with GAC coding on CIFAR-10 at $T = 8$.

| Spike Pattern | Count | Ratio | Spike Pattern | Count | Ratio |
| --- | --- | --- | --- | --- | --- |
| [0, 0, 0, 0, 0, 0, 0, 0] | 338,033,724 | 0.68773 | [1, 0, 0, 0, 0, 0, 0, 0] | 0 | 0 |
| [0, 0, 0, 0, 0, 0, 0, 1] | 420,120 | 0.00085 | [1, 0, 0, 0, 0, 0, 0, 1] | 0 | 0 |
| [0, 0, 0, 0, 0, 0, 1, 0] | 791,838 | 0.00161 | [1, 0, 0, 0, 0, 0, 1, 0] | 0 | 0 |
| [0, 0, 0, 0, 0, 0, 1, 1] | 0 | 0 | [1, 0, 0, 0, 0, 0, 1, 1] | 0 | 0 |
| [0, 0, 0, 0, 0, 1, 0, 0] | 1,570,624 | 0.00319 | [1, 0, 0, 0, 0, 1, 0, 0] | 0 | 0 |
| [0, 0, 0, 0, 0, 1, 0, 1] | 0 | 0 | [1, 0, 0, 0, 0, 1, 0, 1] | 0 | 0 |
| [0, 0, 0, 0, 0, 1, 1, 0] | 0 | 0 | [1, 0, 0, 0, 0, 1, 1, 0] | 0 | 0 |
| [0, 0, 0, 0, 0, 1, 1, 1] | 0 | 0 | [1, 0, 0, 0, 0, 1, 1, 1] | 0 | 0 |
| [0, 0, 0, 0, 1, 0, 0, 0] | 3,244,095 | 0.00660 | [1, 0, 0, 0, 1, 0, 0, 0] | 0 | 0 |
| [0, 0, 0, 0, 1, 0, 0, 1] | 0 | 0 | [1, 0, 0, 0, 1, 0, 0, 1] | 0 | 0 |
| [0, 0, 0, 0, 1, 0, 1, 0] | 0 | 0 | [1, 0, 0, 0, 1, 0, 1, 0] | 0 | 0 |
| [0, 0, 0, 0, 1, 0, 1, 1] | 0 | 0 | [1, 0, 0, 0, 1, 0, 1, 1] | 0 | 0 |
| [0, 0, 0, 0, 1, 1, 0, 0] | 0 | 0 | [1, 0, 0, 0, 1, 1, 0, 0] | 0 | 0 |
| [0, 0, 0, 0, 1, 1, 0, 1] | 0 | 0 | [1, 0, 0, 0, 1, 1, 0, 1] | 0 | 0 |
| [0, 0, 0, 0, 1, 1, 1, 0] | 0 | 0 | [1, 0, 0, 0, 1, 1, 1, 0] | 0 | 0 |
| [0, 0, 0, 0, 1, 1, 1, 1] | 0 | 0 | [1, 0, 0, 0, 1, 1, 1, 1] | 0 | 0 |
| [0, 0, 0, 1, 0, 0, 0, 0] | 5,826,008 | 0.01185 | [1, 0, 0, 1, 0, 0, 0, 0] | 0 | 0 |
| [0, 0, 0, 1, 0, 0, 0, 1] | 1,168,235 | 0.00238 | [1, 0, 0, 1, 0, 0, 0, 1] | 0 | 0 |
| [0, 0, 0, 1, 0, 0, 1, 0] | 0 | 0 | [1, 0, 0, 1, 0, 0, 1, 0] | 0 | 0 |
| [0, 0, 0, 1, 0, 0, 1, 1] | 0 | 0 | [1, 0, 0, 1, 0, 0, 1, 1] | 0 | 0 |
| [0, 0, 0, 1, 0, 1, 0, 0] | 0 | 0 | [1, 0, 0, 1, 0, 1, 0, 0] | 0 | 0 |
| [0, 0, 0, 1, 0, 1, 0, 1] | 0 | 0 | [1, 0, 0, 1, 0, 1, 0, 1] | 0 | 0 |
| [0, 0, 0, 1, 0, 1, 1, 0] | 0 | 0 | [1, 0, 0, 1, 0, 1, 1, 0] | 0 | 0 |
| [0, 0, 0, 1, 0, 1, 1, 1] | 0 | 0 | [1, 0, 0, 1, 0, 1, 1, 1] | 0 | 0 |
| [0, 0, 0, 1, 1, 0, 0, 0] | 0 | 0 | [1, 0, 0, 1, 1, 0, 0, 0] | 0 | 0 |
| [0, 0, 0, 1, 1, 0, 0, 1] | 0 | 0 | [1, 0, 0, 1, 1, 0, 0, 1] | 0 | 0 |
| [0, 0, 0, 1, 1, 0, 1, 0] | 0 | 0 | [1, 0, 0, 1, 1, 0, 1, 0] | 0 | 0 |
| [0, 0, 0, 1, 1, 0, 1, 1] | 0 | 0 | [1, 0, 0, 1, 1, 0, 1, 1] | 0 | 0 |
| [0, 0, 0, 1, 1, 1, 0, 0] | 0 | 0 | [1, 0, 0, 1, 1, 1, 0, 0] | 0 | 0 |
| [0, 0, 0, 1, 1, 1, 0, 1] | 0 | 0 | [1, 0, 0, 1, 1, 1, 0, 1] | 0 | 0 |
| [0, 0, 0, 1, 1, 1, 1, 0] | 0 | 0 | [1, 0, 0, 1, 1, 1, 1, 0] | 0 | 0 |
| [0, 0, 0, 1, 1, 1, 1, 1] | 0 | 0 | [1, 0, 0, 1, 1, 1, 1, 1] | 0 | 0 |
| [0, 0, 1, 0, 0, 0, 0, 0] | 0 | 0 | [1, 0, 1, 0, 0, 0, 0, 0] | 0 | 0 |
| [0, 0, 1, 0, 0, 0, 0, 1] | 273,521 | 0.00056 | [1, 0, 1, 0, 0, 0, 0, 1] | 0 | 0 |
| [0, 0, 1, 0, 0, 0, 1, 0] | 11,686,870 | 0.02378 | [1, 0, 1, 0, 0, 0, 1, 0] | 0 | 0 |
| [0, 0, 1, 0, 0, 0, 1, 1] | 0 | 0 | [1, 0, 1, 0, 0, 0, 1, 1] | 0 | 0 |
| [0, 0, 1, 0, 0, 1, 0, 0] | 3,856,410 | 0.00785 | [1, 0, 1, 0, 0, 1, 0, 0] | 0 | 0 |
| [0, 0, 1, 0, 0, 1, 0, 1] | 0 | 0 | [1, 0, 1, 0, 0, 1, 0, 1] | 0 | 0 |
| [0, 0, 1, 0, 0, 1, 1, 0] | 0 | 0 | [1, 0, 1, 0, 0, 1, 1, 0] | 0 | 0 |
| [0, 0, 1, 0, 0, 1, 1, 1] | 0 | 0 | [1, 0, 1, 0, 0, 1, 1, 1] | 0 | 0 |
| [0, 0, 1, 0, 1, 0, 0, 0] | 0 | 0 | [1, 0, 1, 0, 1, 0, 0, 0] | 0 | 0 |
| [0, 0, 1, 0, 1, 0, 0, 1] | 0 | 0 | [1, 0, 1, 0, 1, 0, 0, 1] | 0 | 0 |
| [0, 0, 1, 0, 1, 0, 1, 0] | 0 | 0 | [1, 0, 1, 0, 1, 0, 1, 0] | 35,350,352 | 0.07192 |
| [0, 0, 1, 0, 1, 0, 1, 1] | 0 | 0 | [1, 0, 1, 0, 1, 0, 1, 1] | 0 | 0 |
| [0, 0, 1, 0, 1, 1, 0, 0] | 0 | 0 | [1, 0, 1, 0, 1, 1, 0, 0] | 0 | 0 |
| [0, 0, 1, 0, 1, 1, 0, 1] | 0 | 0 | [1, 0, 1, 0, 1, 1, 0, 1] | 0 | 0 |
| [0, 0, 1, 0, 1, 1, 1, 0] | 0 | 0 | [1, 0, 1, 0, 1, 1, 1, 0] | 0 | 0 |
| [0, 0, 1, 0, 1, 1, 1, 1] | 0 | 0 | [1, 0, 1, 0, 1, 1, 1, 1] | 0 | 0 |
| [0, 0, 1, 1, 0, 0, 0, 0] | 0 | 0 | [1, 0, 1, 1, 0, 0, 0, 0] | 0 | 0 |
| [0, 0, 1, 1, 0, 0, 0, 1] | 0 | 0 | [1, 0, 1, 1, 0, 0, 0, 1] | 0 | 0 |
| [0, 0, 1, 1, 0, 0, 1, 0] | 0 | 0 | [1, 0, 1, 1, 0, 0, 1, 0] | 0 | 0 |
| [0, 0, 1, 1, 0, 0, 1, 1] | 0 | 0 | [1, 0, 1, 1, 0, 0, 1, 1] | 0 | 0 |
| [0, 0, 1, 1, 0, 1, 0, 0] | 0 | 0 | [1, 0, 1, 1, 0, 1, 0, 0] | 0 | 0 |
| [0, 0, 1, 1, 0, 1, 0, 1] | 0 | 0 | [1, 0, 1, 1, 0, 1, 0, 1] | 0 | 0 |

| Spike Pattern | Count | Ratio | Spike Pattern | Count | Ratio |
| --- | --- | --- | --- | --- | --- |
| [0, 0, 1, 1, 0, 1, 1, 0] | 0 | 0 | [1, 0, 1, 1, 0, 1, 1, 0] | 0 | 0 |
| [0, 0, 1, 1, 0, 1, 1, 1] | 0 | 0 | [1, 0, 1, 1, 0, 1, 1, 1] | 0 | 0 |
| [0, 0, 1, 1, 1, 0, 0, 0] | 0 | 0 | [1, 0, 1, 1, 1, 0, 0, 0] | 0 | 0 |
| [0, 0, 1, 1, 1, 0, 0, 1] | 0 | 0 | [1, 0, 1, 1, 1, 0, 0, 1] | 0 | 0 |
| [0, 0, 1, 1, 1, 0, 1, 0] | 0 | 0 | [1, 0, 1, 1, 1, 0, 1, 0] | 0 | 0 |
| [0, 0, 1, 1, 1, 0, 1, 1] | 0 | 0 | [1, 0, 1, 1, 1, 0, 1, 1] | 0 | 0 |
| [0, 0, 1, 1, 1, 1, 0, 0] | 0 | 0 | [1, 0, 1, 1, 1, 1, 0, 0] | 0 | 0 |
| [0, 0, 1, 1, 1, 1, 0, 1] | 0 | 0 | [1, 0, 1, 1, 1, 1, 0, 1] | 0 | 0 |
| [0, 0, 1, 1, 1, 1, 1, 0] | 0 | 0 | [1, 0, 1, 1, 1, 1, 1, 0] | 0 | 0 |
| [0, 0, 1, 1, 1, 1, 1, 1] | 0 | 0 | [1, 0, 1, 1, 1, 1, 1, 1] | 0 | 0 |
| [0, 1, 0, 0, 0, 0, 0, 0] | 0 | 0 | [1, 1, 0, 0, 0, 0, 0, 0] | 0 | 0 |
| [0, 1, 0, 0, 0, 0, 0, 1] | 0 | 0 | [1, 1, 0, 0, 0, 0, 0, 1] | 0 | 0 |
| [0, 1, 0, 0, 0, 0, 1, 0] | 0 | 0 | [1, 1, 0, 0, 0, 0, 1, 0] | 0 | 0 |
| [0, 1, 0, 0, 0, 0, 1, 1] | 0 | 0 | [1, 1, 0, 0, 0, 0, 1, 1] | 0 | 0 |
| [0, 1, 0, 0, 0, 1, 0, 0] | 103,065 | 0.00021 | [1, 1, 0, 0, 0, 1, 0, 0] | 0 | 0 |
| [0, 1, 0, 0, 0, 1, 0, 1] | 0 | 0 | [1, 1, 0, 0, 0, 1, 0, 1] | 0 | 0 |
| [0, 1, 0, 0, 0, 1, 1, 0] | 0 | 0 | [1, 1, 0, 0, 0, 1, 1, 0] | 0 | 0 |
| [0, 1, 0, 0, 0, 1, 1, 1] | 0 | 0 | [1, 1, 0, 0, 0, 1, 1, 1] | 0 | 0 |
| [0, 1, 0, 0, 1, 0, 0, 0] | 0 | 0 | [1, 1, 0, 0, 1, 0, 0, 0] | 0 | 0 |
| [0, 1, 0, 0, 1, 0, 0, 1] | 22,682,460 | 0.04615 | [1, 1, 0, 0, 1, 0, 0, 1] | 0 | 0 |
| [0, 1, 0, 0, 1, 0, 1, 0] | 0 | 0 | [1, 1, 0, 0, 1, 0, 1, 0] | 0 | 0 |
| [0, 1, 0, 0, 1, 0, 1, 1] | 0 | 0 | [1, 1, 0, 0, 1, 0, 1, 1] | 0 | 0 |
| [0, 1, 0, 0, 1, 1, 0, 0] | 0 | 0 | [1, 1, 0, 0, 1, 1, 0, 0] | 0 | 0 |
| [0, 1, 0, 0, 1, 1, 0, 1] | 0 | 0 | [1, 1, 0, 0, 1, 1, 0, 1] | 0 | 0 |
| [0, 1, 0, 0, 1, 1, 1, 0] | 0 | 0 | [1, 1, 0, 0, 1, 1, 1, 0] | 0 | 0 |
| [0, 1, 0, 0, 1, 1, 1, 1] | 0 | 0 | [1, 1, 0, 0, 1, 1, 1, 1] | 0 | 0 |
| [0, 1, 0, 1, 0, 0, 0, 0] | 0 | 0 | [1, 1, 0, 1, 0, 0, 0, 0] | 0 | 0 |
| [0, 1, 0, 1, 0, 0, 0, 1] | 0 | 0 | [1, 1, 0, 1, 0, 0, 0, 1] | 0 | 0 |
| [0, 1, 0, 1, 0, 0, 1, 0] | 0 | 0 | [1, 1, 0, 1, 0, 0, 1, 0] | 0 | 0 |
| [0, 1, 0, 1, 0, 0, 1, 1] | 0 | 0 | [1, 1, 0, 1, 0, 0, 1, 1] | 0 | 0 |
| [0, 1, 0, 1, 0, 1, 0, 0] | 0 | 0 | [1, 1, 0, 1, 0, 1, 0, 0] | 0 | 0 |
| [0, 1, 0, 1, 0, 1, 0, 1] | 17,535,536 | 0.03568 | [1, 1, 0, 1, 0, 1, 0, 1] | 0 | 0 |
| [0, 1, 0, 1, 0, 1, 1, 0] | 0 | 0 | [1, 1, 0, 1, 0, 1, 1, 0] | 0 | 0 |
| [0, 1, 0, 1, 0, 1, 1, 1] | 0 | 0 | [1, 1, 0, 1, 0, 1, 1, 1] | 0 | 0 |
| [0, 1, 0, 1, 1, 0, 0, 0] | 0 | 0 | [1, 1, 0, 1, 1, 0, 0, 0] | 0 | 0 |
| [0, 1, 0, 1, 1, 0, 0, 1] | 0 | 0 | [1, 1, 0, 1, 1, 0, 0, 1] | 0 | 0 |
| [0, 1, 0, 1, 1, 0, 1, 0] | 0 | 0 | [1, 1, 0, 1, 1, 0, 1, 0] | 0 | 0 |
| [0, 1, 0, 1, 1, 0, 1, 1] | 0 | 0 | [1, 1, 0, 1, 1, 0, 1, 1] | 0 | 0 |
| [0, 1, 0, 1, 1, 1, 0, 0] | 0 | 0 | [1, 1, 0, 1, 1, 1, 0, 0] | 0 | 0 |
| [0, 1, 0, 1, 1, 1, 0, 1] | 0 | 0 | [1, 1, 0, 1, 1, 1, 0, 1] | 0 | 0 |
| [0, 1, 0, 1, 1, 1, 1, 0] | 0 | 0 | [1, 1, 0, 1, 1, 1, 1, 0] | 0 | 0 |
| [0, 1, 0, 1, 1, 1, 1, 1] | 0 | 0 | [1, 1, 0, 1, 1, 1, 1, 1] | 0 | 0 |
| [0, 1, 1, 0, 0, 0, 0, 0] | 0 | 0 | [1, 1, 1, 0, 0, 0, 0, 0] | 0 | 0 |
| [0, 1, 1, 0, 0, 0, 0, 1] | 0 | 0 | [1, 1, 1, 0, 0, 0, 0, 1] | 0 | 0 |
| [0, 1, 1, 0, 0, 0, 1, 0] | 0 | 0 | [1, 1, 1, 0, 0, 0, 1, 0] | 0 | 0 |
| [0, 1, 1, 0, 0, 0, 1, 1] | 0 | 0 | [1, 1, 1, 0, 0, 0, 1, 1] | 0 | 0 |
| [0, 1, 1, 0, 0, 1, 0, 0] | 0 | 0 | [1, 1, 1, 0, 0, 1, 0, 0] | 0 | 0 |
| [0, 1, 1, 0, 0, 1, 0, 1] | 0 | 0 | [1, 1, 1, 0, 0, 1, 0, 1] | 0 | 0 |
| [0, 1, 1, 0, 0, 1, 1, 0] | 0 | 0 | [1, 1, 1, 0, 0, 1, 1, 0] | 0 | 0 |
| [0, 1, 1, 0, 0, 1, 1, 1] | 0 | 0 | [1, 1, 1, 0, 0, 1, 1, 1] | 0 | 0 |
| [0, 1, 1, 0, 1, 0, 0, 0] | 0 | 0 | [1, 1, 1, 0, 1, 0, 0, 0] | 0 | 0 |
| [0, 1, 1, 0, 1, 0, 0, 1] | 0 | 0 | [1, 1, 1, 0, 1, 0, 0, 1] | 0 | 0 |
| [0, 1, 1, 0, 1, 0, 1, 0] | 0 | 0 | [1, 1, 1, 0, 1, 0, 1, 0] | 0 | 0 |
| [0, 1, 1, 0, 1, 0, 1, 1] | 0 | 0 | [1, 1, 1, 0, 1, 0, 1, 1] | 0 | 0 |
| [0, 1, 1, 0, 1, 1, 0, 0] | 0 | 0 | [1, 1, 1, 0, 1, 1, 0, 0] | 0 | 0 |

| Spike Pattern | Count | Ratio | Spike Pattern | Count | Ratio |
| --- | --- | --- | --- | --- | --- |
| [0, 1, 1, 0, 1, 1, 0, 1] | 0 | 0 | [1, 1, 1, 0, 1, 1, 0, 1] | 0 | 0 |
| [0, 1, 1, 0, 1, 1, 1, 0] | 0 | 0 | [1, 1, 1, 0, 1, 1, 1, 0] | 0 | 0 |
| [0, 1, 1, 0, 1, 1, 1, 1] | 0 | 0 | [1, 1, 1, 0, 1, 1, 1, 1] | 0 | 0 |
| [0, 1, 1, 1, 0, 0, 0, 0] | 0 | 0 | [1, 1, 1, 1, 0, 0, 0, 0] | 0 | 0 |
| [0, 1, 1, 1, 0, 0, 0, 1] | 0 | 0 | [1, 1, 1, 1, 0, 0, 0, 1] | 0 | 0 |
| [0, 1, 1, 1, 0, 0, 1, 0] | 0 | 0 | [1, 1, 1, 1, 0, 0, 1, 0] | 0 | 0 |
| [0, 1, 1, 1, 0, 0, 1, 1] | 0 | 0 | [1, 1, 1, 1, 0, 0, 1, 1] | 0 | 0 |
| [0, 1, 1, 1, 0, 1, 0, 0] | 0 | 0 | [1, 1, 1, 1, 0, 1, 0, 0] | 0 | 0 |
| [0, 1, 1, 1, 0, 1, 0, 1] | 0 | 0 | [1, 1, 1, 1, 0, 1, 0, 1] | 0 | 0 |
| [0, 1, 1, 1, 0, 1, 1, 0] | 0 | 0 | [1, 1, 1, 1, 0, 1, 1, 0] | 0 | 0 |
| [0, 1, 1, 1, 0, 1, 1, 1] | 0 | 0 | [1, 1, 1, 1, 0, 1, 1, 1] | 0 | 0 |
| [0, 1, 1, 1, 1, 0, 0, 0] | 0 | 0 | [1, 1, 1, 1, 1, 0, 0, 0] | 0 | 0 |
| [0, 1, 1, 1, 1, 0, 0, 1] | 0 | 0 | [1, 1, 1, 1, 1, 0, 0, 1] | 0 | 0 |
| [0, 1, 1, 1, 1, 0, 1, 0] | 0 | 0 | [1, 1, 1, 1, 1, 0, 1, 0] | 0 | 0 |
| [0, 1, 1, 1, 1, 0, 1, 1] | 0 | 0 | [1, 1, 1, 1, 1, 0, 1, 1] | 0 | 0 |
| [0, 1, 1, 1, 1, 1, 0, 0] | 0 | 0 | [1, 1, 1, 1, 1, 1, 0, 0] | 0 | 0 |
| [0, 1, 1, 1, 1, 1, 0, 1] | 0 | 0 | [1, 1, 1, 1, 1, 1, 0, 1] | 0 | 0 |
| [0, 1, 1, 1, 1, 1, 1, 0] | 0 | 0 | [1, 1, 1, 1, 1, 1, 1, 0] | 0 | 0 |
| [0, 1, 1, 1, 1, 1, 1, 1] | 0 | 0 | [1, 1, 1, 1, 1, 1, 1, 1] | 48,977,142 | 0.09964 |
| Spike Entropy | | | 1.7386 | | |
| Firing Rate | | | 0.1825 | | |

Table 18: Distribution of the spike patterns in the encoding layer of Spikformer with IMP coding on CIFAR-10 at $T = 8$.

| Spike Pattern | Count | Ratio | Spike Pattern | Count | Ratio |
|---|---|---|---|---|---|
| [0, 0, 0, 0, 0, 0, 0, 0] | 166,978,088 | 0.33972 | [1, 0, 0, 0, 0, 0, 0, 0] | 9,513,446 | 0.01936 |
| [0, 0, 0, 0, 0, 0, 0, 1] | 8,830,253 | 0.01796 | [1, 0, 0, 0, 0, 0, 0, 1] | 392,394 | 0.00080 |
| [0, 0, 0, 0, 0, 0, 1, 0] | 8,802,154 | 0.01791 | [1, 0, 0, 0, 0, 0, 1, 0] | 2,181,415 | 0.00444 |
| [0, 0, 0, 0, 0, 0, 1, 1] | 924,376 | 0.00188 | [1, 0, 0, 0, 0, 0, 1, 1] | 135,224 | 0.00028 |
| [0, 0, 0, 0, 0, 1, 0, 0] | 10,439,976 | 0.02124 | [1, 0, 0, 0, 0, 1, 0, 0] | 418,115 | 0.00085 |
| [0, 0, 0, 0, 0, 1, 0, 1] | 6,276,028 | 0.01277 | [1, 0, 0, 0, 0, 1, 0, 1] | 230,261 | 0.00047 |
| [0, 0, 0, 0, 0, 1, 1, 0] | 707,642 | 0.00144 | [1, 0, 0, 0, 0, 1, 1, 0] | 109,016 | 0.00022 |
| [0, 0, 0, 0, 0, 1, 1, 1] | 654,040 | 0.00133 | [1, 0, 0, 0, 0, 1, 1, 1] | 80,885 | 0.00016 |
| [0, 0, 0, 0, 1, 0, 0, 0] | 10,414,246 | 0.02119 | [1, 0, 0, 0, 1, 0, 0, 0] | 1,152,106 | 0.00234 |
| [0, 0, 0, 0, 1, 0, 0, 1] | 709,126 | 0.00144 | [1, 0, 0, 0, 1, 0, 0, 1] | 60,478 | 0.00012 |
| [0, 0, 0, 0, 1, 0, 1, 0] | 6,254,257 | 0.01272 | [1, 0, 0, 0, 1, 0, 1, 0] | 1,728,587 | 0.00352 |
| [0, 0, 0, 0, 1, 0, 1, 1] | 653,259 | 0.00133 | [1, 0, 0, 0, 1, 0, 1, 1] | 133,119 | 0.00027 |
| [0, 0, 0, 0, 1, 1, 0, 0] | 1,055,761 | 0.00215 | [1, 0, 0, 0, 1, 1, 0, 0] | 74,482 | 0.00015 |
| [0, 0, 0, 0, 1, 1, 0, 1] | 611,561 | 0.00124 | [1, 0, 0, 0, 1, 1, 0, 1] | 42,115 | 0.00009 |
| [0, 0, 0, 0, 1, 1, 1, 0] | 608,929 | 0.00124 | [1, 0, 0, 0, 1, 1, 1, 0] | 112,564 | 0.00023 |
| [0, 0, 0, 0, 1, 1, 1, 1] | 569,225 | 0.00116 | [1, 0, 0, 0, 1, 1, 1, 1] | 96,101 | 0.00020 |
| [0, 0, 0, 1, 0, 0, 0, 0] | 13,875,066 | 0.02823 | [1, 0, 0, 1, 0, 0, 0, 0] | 801,618 | 0.00163 |
| [0, 0, 0, 1, 0, 0, 0, 1] | 2,203,497 | 0.00448 | [1, 0, 0, 1, 0, 0, 0, 1] | 96,118 | 0.00020 |
| [0, 0, 0, 1, 0, 0, 1, 0] | 586,408 | 0.00119 | [1, 0, 0, 1, 0, 0, 1, 0] | 133,181 | 0.00027 |
| [0, 0, 0, 1, 0, 0, 1, 1] | 155,802 | 0.00032 | [1, 0, 0, 1, 0, 0, 1, 1] | 26,500 | 0.00005 |
| [0, 0, 0, 1, 0, 1, 0, 0] | 8,315,241 | 0.01692 | [1, 0, 0, 1, 0, 1, 0, 0] | 357,330 | 0.00073 |
| [0, 0, 0, 1, 0, 1, 0, 1] | 11,461,260 | 0.02332 | [1, 0, 0, 1, 0, 1, 0, 1] | 378,856 | 0.00077 |
| [0, 0, 0, 1, 0, 1, 1, 0] | 433,513 | 0.00088 | [1, 0, 0, 1, 0, 1, 1, 0] | 71,315 | 0.00015 |
| [0, 0, 0, 1, 0, 1, 1, 1] | 860,104 | 0.00175 | [1, 0, 0, 1, 0, 1, 1, 1] | 105,466 | 0.00021 |
| [0, 0, 0, 1, 1, 0, 0, 0] | 883,826 | 0.00180 | [1, 0, 0, 1, 1, 0, 0, 0] | 87,782 | 0.00018 |
| [0, 0, 0, 1, 1, 0, 0, 1] | 167,854 | 0.00034 | [1, 0, 0, 1, 1, 0, 0, 1] | 13,783 | 0.00003 |
| [0, 0, 0, 1, 1, 0, 1, 0] | 405,627 | 0.00082 | [1, 0, 0, 1, 1, 0, 1, 0] | 99,569 | 0.00020 |
| [0, 0, 0, 1, 1, 0, 1, 1] | 108,798 | 0.00022 | [1, 0, 0, 1, 1, 0, 1, 1] | 22,741 | 0.00005 |
| [0, 0, 0, 1, 1, 1, 0, 0] | 755,319 | 0.00154 | [1, 0, 0, 1, 1, 1, 0, 0] | 55,661 | 0.00011 |
| [0, 0, 0, 1, 1, 1, 0, 1] | 1,068,954 | 0.00217 | [1, 0, 0, 1, 1, 1, 0, 1] | 67,964 | 0.00014 |
| [0, 0, 0, 1, 1, 1, 1, 0] | 359,435 | 0.00073 | [1, 0, 0, 1, 1, 1, 1, 0] | 69,454 | 0.00014 |
| [0, 0, 0, 1, 1, 1, 1, 1] | 744,798 | 0.00152 | [1, 0, 0, 1, 1, 1, 1, 1] | 115,930 | 0.00024 |
| [0, 0, 1, 0, 0, 0, 0, 0] | 13,875,540 | 0.02823 | [1, 0, 1, 0, 0, 0, 0, 0] | 5,314,878 | 0.01081 |
| [0, 0, 1, 0, 0, 0, 0, 1] | 586,328 | 0.00119 | [1, 0, 1, 0, 0, 0, 0, 1] | 165,056 | 0.00034 |
| [0, 0, 1, 0, 0, 0, 1, 0] | 2,202,679 | 0.00448 | [1, 0, 1, 0, 0, 0, 1, 0] | 1,868,819 | 0.00380 |
| [0, 0, 1, 0, 0, 0, 1, 1] | 155,493 | 0.00032 | [1, 0, 1, 0, 0, 0, 1, 1] | 86,177 | 0.00017 |
| [0, 0, 1, 0, 0, 1, 0, 0] | 885,163 | 0.00180 | [1, 0, 1, 0, 0, 1, 0, 0] | 230,771 | 0.00047 |
| [0, 0, 1, 0, 0, 1, 0, 1] | 404,652 | 0.00082 | [1, 0, 1, 0, 0, 1, 0, 1] | 100,306 | 0.00020 |
| [0, 0, 1, 0, 0, 1, 1, 0] | 167,824 | 0.00034 | [1, 0, 1, 0, 0, 1, 1, 0] | 90,448 | 0.00018 |
| [0, 0, 1, 0, 0, 1, 1, 1] | 108,576 | 0.00022 | [1, 0, 1, 0, 0, 1, 1, 1] | 50,985 | 0.00010 |
| [0, 0, 1, 0, 1, 0, 0, 0] | 8,283,191 | 0.01685 | [1, 0, 1, 0, 1, 0, 0, 0] | 3,759,113 | 0.00765 |
| [0, 0, 1, 0, 1, 0, 0, 1] | 434,009 | 0.00088 | [1, 0, 1, 0, 1, 0, 0, 1] | 150,195 | 0.00031 |
| [0, 0, 1, 0, 1, 0, 1, 0] | 11,461,617 | 0.02332 | [1, 0, 1, 0, 1, 0, 1, 0] | 25,557,536 | 0.05200 |
| [0, 0, 1, 0, 1, 0, 1, 1] | 863,192 | 0.00176 | [1, 0, 1, 0, 1, 0, 1, 1] | 1,423,576 | 0.00290 |
| [0, 0, 1, 0, 1, 1, 0, 0] | 754,439 | 0.00153 | [1, 0, 1, 0, 1, 1, 0, 0] | 236,823 | 0.00048 |
| [0, 0, 1, 0, 1, 1, 0, 1] | 358,953 | 0.00073 | [1, 0, 1, 0, 1, 1, 0, 1] | 105,855 | 0.00021 |
| [0, 0, 1, 0, 1, 1, 1, 0] | 1,070,523 | 0.00218 | [1, 0, 1, 0, 1, 1, 1, 0] | 1,705,014 | 0.00347 |
| [0, 0, 1, 0, 1, 1, 1, 1] | 749,485 | 0.00153 | [1, 0, 1, 0, 1, 1, 1, 1] | 1,076,246 | 0.00219 |
| [0, 0, 1, 1, 0, 0, 0, 0] | 2,069,223 | 0.00421 | [1, 0, 1, 1, 0, 0, 0, 0] | 619,136 | 0.00126 |
| [0, 0, 1, 1, 0, 0, 0, 1] | 246,054 | 0.00050 | [1, 0, 1, 1, 0, 0, 0, 1] | 59,275 | 0.00012 |
| [0, 0, 1, 1, 0, 0, 1, 0] | 246,872 | 0.00050 | [1, 0, 1, 1, 0, 0, 1, 0] | 174,810 | 0.00036 |
| [0, 0, 1, 1, 0, 0, 1, 1] | 51,835 | 0.00011 | [1, 0, 1, 1, 0, 0, 1, 1] | 26,124 | 0.00005 |
| [0, 0, 1, 1, 0, 1, 0, 0] | 1,162,307 | 0.00236 | [1, 0, 1, 1, 0, 1, 0, 0] | 264,551 | 0.00054 |
| [0, 0, 1, 1, 0, 1, 0, 1] | 1,138,064 | 0.00231 | [1, 0, 1, 1, 0, 1, 0, 1] | 228,051 | 0.00046 |

| Spike Pattern | Count | Ratio | Spike Pattern | Count | Ratio |
| --- | --- | --- | --- | --- | --- |
| [0, 0, 1, 1, 0, 1, 1, 0] | 173,135 | 0.00035 | [1, 0, 1, 1, 0, 1, 1, 0] | 86,136 | 0.00017 |
| [0, 0, 1, 1, 0, 1, 1, 1] | 258,765 | 0.00053 | [1, 0, 1, 1, 0, 1, 1, 1] | 99,880 | 0.00020 |
| [0, 0, 1, 1, 1, 0, 0, 0] | 1,156,988 | 0.00235 | [1, 0, 1, 1, 1, 0, 0, 0] | 403,385 | 0.00082 |
| [0, 0, 1, 1, 1, 0, 0, 1] | 172,341 | 0.00035 | [1, 0, 1, 1, 1, 0, 0, 1] | 50,620 | 0.00010 |
| [0, 0, 1, 1, 1, 0, 1, 0] | 1,139,849 | 0.00232 | [1, 0, 1, 1, 1, 0, 1, 0] | 2,158,541 | 0.00439 |
| [0, 0, 1, 1, 1, 0, 1, 1] | 260,298 | 0.00053 | [1, 0, 1, 1, 1, 0, 1, 1] | 413,268 | 0.00084 |
| [0, 0, 1, 1, 1, 1, 0, 0] | 938,898 | 0.00191 | [1, 0, 1, 1, 1, 1, 0, 0] | 256,378 | 0.00052 |
| [0, 0, 1, 1, 1, 1, 0, 1] | 1,051,950 | 0.00214 | [1, 0, 1, 1, 1, 1, 0, 1] | 249,442 | 0.00051 |
| [0, 0, 1, 1, 1, 1, 1, 0] | 1,053,394 | 0.00214 | [1, 0, 1, 1, 1, 1, 1, 0] | 1,597,605 | 0.00325 |
| [0, 0, 1, 1, 1, 1, 1, 1] | 1,832,734 | 0.00373 | [1, 0, 1, 1, 1, 1, 1, 1] | 2,463,570 | 0.00501 |
| [0, 1, 0, 0, 0, 0, 0, 0] | 9,590,581 | 0.01951 | [1, 1, 0, 0, 0, 0, 0, 0] | 1,647,164 | 0.00335 |
| [0, 1, 0, 0, 0, 0, 0, 1] | 2,226,551 | 0.00453 | [1, 1, 0, 0, 0, 0, 0, 1] | 340,639 | 0.00069 |
| [0, 1, 0, 0, 0, 0, 1, 0] | 397,399 | 0.00081 | [1, 1, 0, 0, 0, 0, 1, 0] | 342,632 | 0.00070 |
| [0, 1, 0, 0, 0, 0, 1, 1] | 137,468 | 0.00028 | [1, 1, 0, 0, 0, 0, 1, 1] | 101,762 | 0.00021 |
| [0, 1, 0, 0, 0, 1, 0, 0] | 1,172,184 | 0.00238 | [1, 1, 0, 0, 0, 1, 0, 0] | 157,183 | 0.00032 |
| [0, 1, 0, 0, 0, 1, 0, 1] | 1,748,051 | 0.00356 | [1, 1, 0, 0, 0, 1, 0, 1] | 196,883 | 0.00040 |
| [0, 1, 0, 0, 0, 1, 1, 0] | 61,588 | 0.00013 | [1, 1, 0, 0, 0, 1, 1, 0] | 36,292 | 0.00007 |
| [0, 1, 0, 0, 0, 1, 1, 1] | 133,859 | 0.00027 | [1, 1, 0, 0, 0, 1, 1, 1] | 60,848 | 0.00012 |
| [0, 1, 0, 0, 1, 0, 0, 0] | 425,154 | 0.00086 | [1, 1, 0, 0, 1, 0, 0, 0] | 156,747 | 0.00032 |
| [0, 1, 0, 0, 1, 0, 0, 1] | 112,041 | 0.00023 | [1, 1, 0, 0, 1, 0, 0, 1] | 36,216 | 0.00007 |
| [0, 1, 0, 0, 1, 0, 1, 0] | 233,155 | 0.00047 | [1, 1, 0, 0, 1, 0, 1, 0] | 196,911 | 0.00040 |
| [0, 1, 0, 0, 1, 0, 1, 1] | 81,906 | 0.00017 | [1, 1, 0, 0, 1, 0, 1, 1] | 61,451 | 0.00013 |
| [0, 1, 0, 0, 1, 1, 0, 0] | 75,695 | 0.00015 | [1, 1, 0, 0, 1, 1, 0, 0] | 22,643 | 0.00005 |
| [0, 1, 0, 0, 1, 1, 0, 1] | 114,457 | 0.00023 | [1, 1, 0, 0, 1, 1, 0, 1] | 28,090 | 0.00006 |
| [0, 1, 0, 0, 1, 1, 1, 0] | 43,172 | 0.00009 | [1, 1, 0, 0, 1, 1, 1, 0] | 27,892 | 0.00006 |
| [0, 1, 0, 0, 1, 1, 1, 1] | 96,608 | 0.00020 | [1, 1, 0, 0, 1, 1, 1, 1] | 51,863 | 0.00011 |
| [0, 1, 0, 1, 0, 0, 0, 0] | 5,357,109 | 0.01090 | [1, 1, 0, 1, 0, 0, 0, 0] | 736,347 | 0.00150 |
| [0, 1, 0, 1, 0, 0, 0, 1] | 1,897,437 | 0.00386 | [1, 1, 0, 1, 0, 0, 0, 1] | 247,622 | 0.00050 |
| [0, 1, 0, 1, 0, 0, 1, 0] | 168,450 | 0.00034 | [1, 1, 0, 1, 0, 0, 1, 0] | 113,743 | 0.00023 |
| [0, 1, 0, 1, 0, 0, 1, 1] | 86,691 | 0.00018 | [1, 1, 0, 1, 0, 0, 1, 1] | 55,728 | 0.00011 |
| [0, 1, 0, 1, 0, 1, 0, 0] | 3,782,721 | 0.00770 | [1, 1, 0, 1, 0, 1, 0, 0] | 399,293 | 0.00081 |
| [0, 1, 0, 1, 0, 1, 0, 1] | 25,384,806 | 0.05165 | [1, 1, 0, 1, 0, 1, 0, 1] | 1,931,294 | 0.00393 |
| [0, 1, 0, 1, 0, 1, 1, 0] | 151,732 | 0.00031 | [1, 1, 0, 1, 0, 1, 1, 0] | 72,079 | 0.00015 |
| [0, 1, 0, 1, 0, 1, 1, 1] | 1,418,875 | 0.00289 | [1, 1, 0, 1, 0, 1, 1, 1] | 460,687 | 0.00094 |
| [0, 1, 0, 1, 1, 0, 0, 0] | 235,228 | 0.00048 | [1, 1, 0, 1, 1, 0, 0, 0] | 68,648 | 0.00014 |
| [0, 1, 0, 1, 1, 0, 0, 1] | 92,351 | 0.00019 | [1, 1, 0, 1, 1, 0, 0, 1] | 25,366 | 0.00005 |
| [0, 1, 0, 1, 1, 0, 1, 0] | 102,671 | 0.00021 | [1, 1, 0, 1, 1, 0, 1, 0] | 67,715 | 0.00014 |
| [0, 1, 0, 1, 1, 0, 1, 1] | 51,490 | 0.00011 | [1, 1, 0, 1, 1, 0, 1, 1] | 33,684 | 0.00007 |
| [0, 1, 0, 1, 1, 1, 0, 0] | 239,042 | 0.00049 | [1, 1, 0, 1, 1, 1, 0, 0] | 53,849 | 0.00011 |
| [0, 1, 0, 1, 1, 1, 0, 1] | 1,699,280 | 0.00346 | [1, 1, 0, 1, 1, 1, 0, 1] | 284,509 | 0.00058 |
| [0, 1, 0, 1, 1, 1, 1, 0] | 107,626 | 0.00022 | [1, 1, 0, 1, 1, 1, 1, 0] | 54,865 | 0.00011 |
| [0, 1, 0, 1, 1, 1, 1, 1] | 1,071,170 | 0.00218 | [1, 1, 0, 1, 1, 1, 1, 1] | 405,608 | 0.00082 |
| [0, 1, 1, 0, 0, 0, 0, 0] | 805,300 | 0.00164 | [1, 1, 1, 0, 0, 0, 0, 0] | 735,791 | 0.00150 |
| [0, 1, 1, 0, 0, 0, 0, 1] | 137,191 | 0.00028 | [1, 1, 1, 0, 0, 0, 0, 1] | 114,502 | 0.00023 |
| [0, 1, 1, 0, 0, 0, 1, 0] | 98,351 | 0.00020 | [1, 1, 1, 0, 0, 0, 1, 0] | 245,772 | 0.00050 |
| [0, 1, 1, 0, 0, 0, 1, 1] | 27,141 | 0.00006 | [1, 1, 1, 0, 0, 0, 1, 1] | 55,104 | 0.00011 |
| [0, 1, 1, 0, 0, 1, 0, 0] | 89,126 | 0.00018 | [1, 1, 1, 0, 0, 1, 0, 0] | 68,869 | 0.00014 |
| [0, 1, 1, 0, 0, 1, 0, 1] | 100,748 | 0.00020 | [1, 1, 1, 0, 0, 1, 0, 1] | 67,376 | 0.00014 |
| [0, 1, 1, 0, 0, 1, 1, 0] | 14,038 | 0.00003 | [1, 1, 1, 0, 0, 1, 1, 0] | 25,216 | 0.00005 |
| [0, 1, 1, 0, 0, 1, 1, 1] | 22,889 | 0.00005 | [1, 1, 1, 0, 0, 1, 1, 1] | 33,126 | 0.00007 |
| [0, 1, 1, 0, 1, 0, 0, 0] | 359,617 | 0.00073 | [1, 1, 1, 0, 1, 0, 0, 0] | 396,169 | 0.00081 |
| [0, 1, 1, 0, 1, 0, 0, 1] | 72,902 | 0.00015 | [1, 1, 1, 0, 1, 0, 0, 1] | 71,954 | 0.00015 |
| [0, 1, 1, 0, 1, 0, 1, 0] | 383,809 | 0.00078 | [1, 1, 1, 0, 1, 0, 1, 0] | 1,940,450 | 0.00395 |
| [0, 1, 1, 0, 1, 0, 1, 1] | 107,570 | 0.00022 | [1, 1, 1, 0, 1, 0, 1, 1] | 463,704 | 0.00094 |
| [0, 1, 1, 0, 1, 1, 0, 0] | 56,599 | 0.00012 | [1, 1, 1, 0, 1, 1, 0, 0] | 53,706 | 0.00011 |

| Spike Pattern | Count | Ratio | Spike Pattern | Count | Ratio |
|---|---|---|---|---|---|
| [0, 1, 1, 0, 1, 1, 0, 1] | 69,347 | 0.00014 | [1, 1, 1, 0, 1, 1, 0, 1] | 55,115 | 0.00011 |
| [0, 1, 1, 0, 1, 1, 1, 0] | 68,691 | 0.00014 | [1, 1, 1, 0, 1, 1, 1, 0] | 288,965 | 0.00059 |
| [0, 1, 1, 0, 1, 1, 1, 1] | 117,212 | 0.00024 | [1, 1, 1, 0, 1, 1, 1, 1] | 410,063 | 0.00083 |
| [0, 1, 1, 1, 0, 0, 0, 0] | 626,320 | 0.00127 | [1, 1, 1, 1, 0, 0, 0, 0] | 524,667 | 0.00107 |
| [0, 1, 1, 1, 0, 0, 0, 1] | 176,202 | 0.00036 | [1, 1, 1, 1, 0, 0, 0, 1] | 141,241 | 0.00029 |
| [0, 1, 1, 1, 0, 0, 1, 0] | 60,238 | 0.00012 | [1, 1, 1, 1, 0, 0, 1, 0] | 140,653 | 0.00029 |
| [0, 1, 1, 1, 0, 0, 1, 1] | 26,423 | 0.00005 | [1, 1, 1, 1, 0, 0, 1, 1] | 53,288 | 0.00011 |
| [0, 1, 1, 1, 0, 1, 0, 0] | 407,982 | 0.00083 | [1, 1, 1, 1, 0, 1, 0, 0] | 282,270 | 0.00057 |
| [0, 1, 1, 1, 0, 1, 0, 1] | 2,151,062 | 0.00438 | [1, 1, 1, 1, 0, 1, 0, 1] | 1,174,816 | 0.00239 |
| [0, 1, 1, 1, 0, 1, 1, 0] | 51,179 | 0.00010 | [1, 1, 1, 1, 0, 1, 1, 0] | 83,582 | 0.00017 |
| [0, 1, 1, 1, 0, 1, 1, 1] | 411,145 | 0.00084 | [1, 1, 1, 1, 0, 1, 1, 1] | 456,045 | 0.00093 |
| [0, 1, 1, 1, 1, 0, 0, 0] | 270,206 | 0.00055 | [1, 1, 1, 1, 1, 0, 0, 0] | 282,787 | 0.00057 |
| [0, 1, 1, 1, 1, 0, 0, 1] | 87,441 | 0.00018 | [1, 1, 1, 1, 1, 0, 0, 1] | 83,519 | 0.00017 |
| [0, 1, 1, 1, 1, 0, 1, 0] | 231,147 | 0.00047 | [1, 1, 1, 1, 1, 0, 1, 0] | 1,183,180 | 0.00241 |
| [0, 1, 1, 1, 1, 0, 1, 1] | 101,422 | 0.00021 | [1, 1, 1, 1, 1, 0, 1, 1] | 458,876 | 0.00093 |
| [0, 1, 1, 1, 1, 1, 0, 0] | 258,769 | 0.00053 | [1, 1, 1, 1, 1, 1, 0, 0] | 228,059 | 0.00046 |
| [0, 1, 1, 1, 1, 1, 0, 1] | 1,586,548 | 0.00323 | [1, 1, 1, 1, 1, 1, 0, 1] | 1,057,587 | 0.00215 |
| [0, 1, 1, 1, 1, 1, 1, 0] | 252,294 | 0.00051 | [1, 1, 1, 1, 1, 1, 1, 0] | 1,065,182 | 0.00217 |
| [0, 1, 1, 1, 1, 1, 1, 1] | 2,453,351 | 0.00499 | [1, 1, 1, 1, 1, 1, 1, 1] | 8,398,365 | 0.01709 |
| Spike Entropy | | | | 4.9319 | |
| Firing Rate | | | | 0.2464 | |

Table 19: Distribution of the spike patterns in the encoding layer of Spikformer with STF on CIFAR-10 at $T = 8$.



\end{document}